\documentclass[10pt,twocolumn,letterpaper]{article}

\usepackage{iccv}
\usepackage{times}
\usepackage{epsfig}
\usepackage{graphicx}
\usepackage{amsmath}
\usepackage{amssymb}
\usepackage{float}
\usepackage[norelsize, linesnumbered, ruled, lined, boxed, commentsnumbered]{algorithm2e}
\usepackage{subcaption}
\usepackage{placeins}
\usepackage{verbatim}
\usepackage{booktabs}
\usepackage{multirow}
\usepackage{nicefrac}

\usepackage[pagebackref=true,breaklinks=true,letterpaper=true,colorlinks,bookmarks=false]{hyperref}

\iccvfinalcopy 

\def\iccvPaperID{3080} 

\ificcvfinal\fi
\begin{document}

\title{Image2StyleGAN: How to Embed Images Into the StyleGAN Latent Space?}

\author{Rameen Abdal\\ KAUST \\{\tt\small rameen.abdal@kaust.edu.sa} \and Yipeng Qin \\ KAUST \\ {\tt\small yipeng.qin@kaust.edu.sa} \and Peter Wonka\\ KAUST \\{\tt\small pwonka@gmail.com}}


\maketitle

\begin{abstract}

We propose an efficient algorithm to embed a given image into the latent space of StyleGAN. This embedding enables semantic image editing operations that can be applied to existing photographs. Taking the StyleGAN trained on the FFHQ dataset as an example, we show results for image morphing, style transfer, and expression transfer.
Studying the results of the embedding algorithm provides valuable insights into the structure of the StyleGAN latent space. We propose a set of experiments to test what class of images can be embedded, how they are embedded, what latent space is suitable for embedding, and if the embedding is semantically meaningful.

\end{abstract}

\section{Introduction}

Generative Adverserial Networks (GANs) are very successfully applied in various computer vision applications, \eg texture synthesis \cite{TextureSynthesis2016,Xian_2018_CVPR,slossberg2018high}, video generation \cite{VideoGeneration2016,Tulyakov_2018_CVPR}, image-to-image translation \cite{pix2pix2017,CycleGAN2017,alharbi2018latent,park2019SPADE} and object detection \cite{ObjectDetection2017}.

In the few past years, the quality of images synthesized by GANs has increased rapidly.
Compared to the seminal DCGAN framework~\cite{DCGAN2015} in 2015, the current state-of-the-art GANs \cite{ProgressiveGAN2018,BigGAN2019,StyleGAN2018,CycleGAN2017,NIPS2017_6650} can synthesize at a much higher resolution and produce significantly more realistic images.
Among them, StyleGAN \cite{StyleGAN2018} makes use of an intermediate $W$ latent space that holds the promise of enabling some controlled image modifications.
We believe that image modifications are a lot more exciting when it becomes possible to modify a given image rather than a randomly GAN generated one. This leads to the natural question if it is possible to embed a given photograph into the GAN latent space.

To tackle this question, we build an embedding algorithm that can map a given image $I$ in the latent space of StyleGAN pre-trained on the FFHQ dataset. One of our important insights is that the generalization ability of the pre-trained StyleGAN is significantly enhanced when using an extended latent space $W^+$ (See Sec.~\ref{subsec:whichlatentspace}). As a consequence, somewhat surprisingly, our embedding algorithm is not only able to embed human face images, but also successfully embeds non-face images from different classes. Therefore, we continue our investigation by analyzing the quality of the embedding to see if the embedding is semantically meaningful.
To this end, we propose to use three basic operations on vectors in the latent space: linear interpolation, crossover, and adding a vector and a scaled difference vector. These operations correspond to three semantic image processing applications:  morphing, style transfer, and expression transfer. As a result, we gain more insight into the structure of the latent space and can solve the mystery why even instances of non-face images such as cars can be embedded.

Our contributions include:
\begin{itemize}
    \item An efficient embedding algorithm which can map a given image into the extended latent space $W^+$ of a pre-trained StyleGAN.
    \item We study multiple questions providing insight into the structure of the StyleGAN latent space, e.g.: What type of images can be embedded? What type of faces can be embedded? What latent space can be used for the embedding?
    \item We propose to use three basic operations on vectors to study the quality of the embedding. As a result, we can better understand the latent space and how different classes of images are embedded. As a byproduct, we obtain excellent results on multiple face image editing applications including morphing, style transfer, and expression transfer.
\end{itemize}


\begin{figure*}[t]
\includegraphics[width=\textwidth]{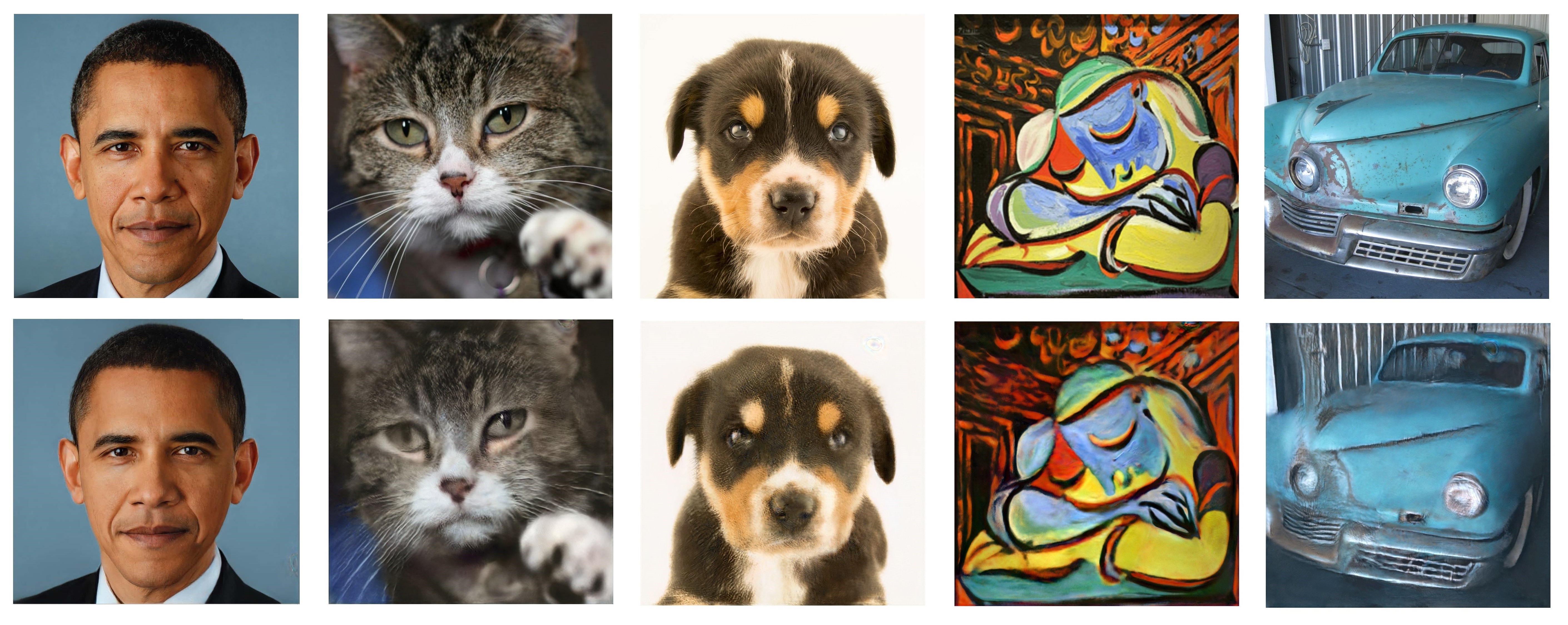}
 \caption{Top row: input images. Bottom row: results of embedding the images into the StyleGAN latent space.}
    \label{fig:embedding_result}
\end{figure*}

\section{Related Work}

\paragraph{High-quality GANs}
Starting from the groundbreaking work by Goodfellow \etal \cite{goodfellow2014generative} in 2014, the entire computer vision community has witnessed the fast-paced improvements on GANs in the past years.
For image generation tasks, DCGAN \cite{DCGAN2015} is the first milestone that lays down the foundation of GAN architectures as fully-convolutional neural networks.
Since then, various efforts have been made to improve the performance of GANs from different aspects, \eg the loss function \cite{LSGAN2017,WGAN2017}, the regularization or normalization \cite{WGANGP2017,SpectralNormalization2018}, and the architecture \cite{WGANGP2017}.
However, due to the limitation of computational power and the shortage of high-quality training data, these works are only tested with low resolution and poor quality datasets collected for classification / recognition tasks.
Addressing this issue, Karras \etal collected the first high-quality human face dataset CelebA-HQ and proposed a progressive strategy to train GANs for high resolution image generation tasks \cite{ProgressiveGAN2018}. 
Their ProGAN is the first GAN that can generate realistic human faces at a high resolution of $1024 \times 1024$.
However, the generation of high-quality images from complex datasets (\eg ImageNet) remains a challenge.
To this end, Brock \etal proposed BigGAN and argued that the training of GANs benefit dramatically from large batch sizes \cite{BigGAN2019}.
Their BigGAN can generate realistic samples and smooth interpolations spanning different classes.
Recently, Karras \etal collected a more diverse and higher quality human face dataset FFHQ and proposed a new generator architecture inspired by the idea of neural style transfer \cite{Adain2017}, which further improves the performance of GANs on human face generation tasks \cite{StyleGAN2018}.
However, the lack of control over image modification ascribed to the interpretability of neural networks, is still an open problem.
In this paper, we tackle the interpretability problem by embedding user-specified images back to the GAN latent space, which leads to a variety of potential applications.

\paragraph{Latent Space Embedding}
In general, there are two existing approaches to embed instances from the image space to the latent space: i) learn an encoder that maps a given image to the latent space (\eg the Variational Auto-Encoder \cite{VAE2013}); ii) select a random initial latent code and optimize it using gradient descent \cite{Zhu_2016,GANEmbedding2018}.
Between them, the first approach provides a fast solution of image embedding by performing a forward pass through the encoder neural network.
However, it usually has problems generalizing beyond the training dataset.
In this paper, we decided to build on the second approach as the more general and stable solution. As a concurrently developed work, the Github repository stylegan-encoder \cite{git2} also demonstrated that the optimization-based approach leads to embeddings of very high visual quality.

\paragraph{Perceptual Loss and Style Transfer}
Traditionally, the low-level similarity between two images is measured in the pixel space with $L1/L2$ loss functions.
While in the past years, inspired by the success of complex image classification \cite{Alexnet2012,VGG2015}, Gatys \etal \cite{GatysTexture2015,GatysStyle2015} observed that the learned filters of the VGG image classification model \cite{VGG2015} are excellent general-purpose feature extractors and proposed to use the covariance statistics of the extracted features to measure the high-level similarity between images \textit{perceptually}, which is then formalized as the \textit{perceptual loss} \cite{PerceptualLoss2016,dosovitskiy2016generating}.
To demonstrate the power of their method, they showed promising results on style transfer \cite{GatysStyle2015}.

Specifically, they argued that different layers of the VGG neural network extract the image features at different scales and can be separated into \textit{content} and \textit{style}.

To accelerate the initial algorithm, Johnson \etal \cite{PerceptualLoss2016} proposed to train a neural network to solve the optimization problem of \cite{GatysStyle2015}, which can transfer the style of a given image to any other image in real-time.
The only limitation of their method is that they need to train separate neural networks for different style images.
Finally, this issue is resolved by Huang and Belongie \cite{Adain2017} with adaptive instance normalization.
As a result, they can transfer arbitrary style in real-time.

\begin{figure*}[t]
\includegraphics[width=\textwidth]{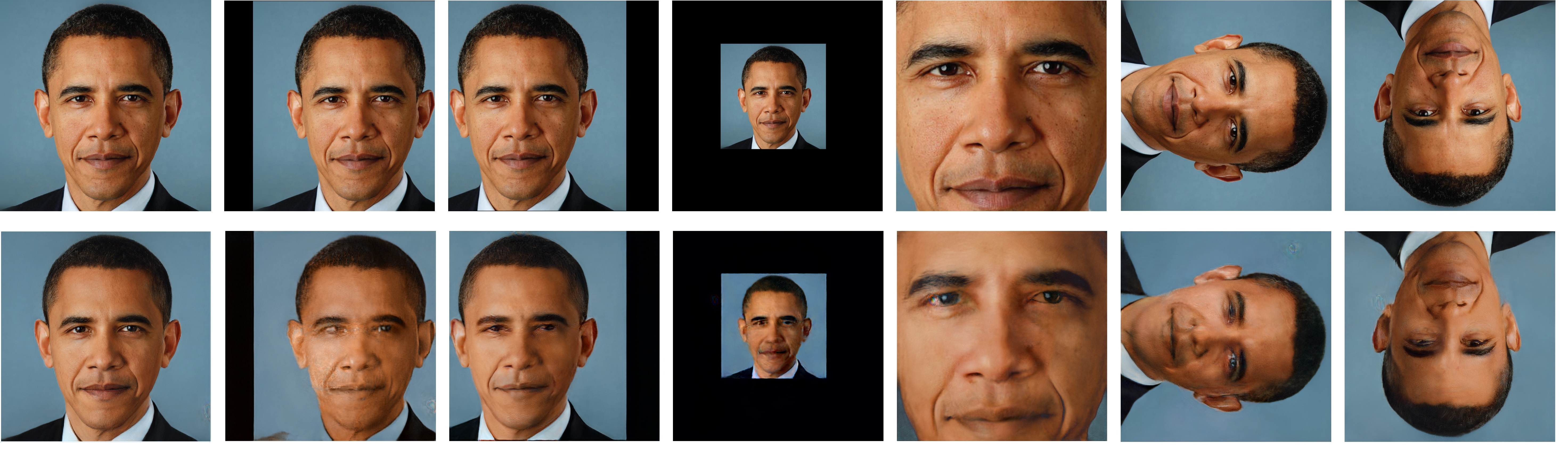}
\hspace*{0.9cm}(a)\hspace{2.1cm}(b)\hspace{2.1cm}(c)\hspace{2.1cm}(d)\hspace{2.1cm}(e)\hspace{2.1cm}(f)\hspace{2.1cm}(g)	
 \caption{Top row: the input images. Bottom row: the embedded results. (a) Standard embedding results. (b) Translation $140$ pixels to the right. (c) Translation $160$ pixels to the left. (d) Zoom out by $2X$. (e) Zoom in by $2X$. (f) 90$^\circ$ rotation. (g) 180$^\circ$ rotation.}
\label{fig:affine_transformation}
\end{figure*}
\section{What images can be embedded into the StyleGAN latent space?}

We set out to study the question if it is even possible to embed images into the StyleGAN latent space. This question is not trivial, because our initial embedding experiments with faces and with other GANs resulted in faces that were no longer recognizable as the same person.
Due to the improved variability of the FFHQ dataset and the superior quality of the StyleGAN architecture, there is a renewed hope that embedding existing images in the latent space is possible.

\subsection{Embedding Results for Various Image Classes}
To test our method, we collect a small-scale dataset of $25$ diverse images spanning $5$ categories (\ie faces, cats, dogs, cars, and paintings). Details of the dataset are shown in the supplementary material. We use the code provided by StyleGAN~\cite{StyleGAN2018} to preprocess the face images. This preprocess includes registration to a canonical face position.

To better understand the structure and attributes of the latent space, it is beneficial to study the embedding of a larger variety of image classes. We choose faces of cats, dogs, and paintings as they share the overall structure with human faces, but are depicted in a very different style. Cars are selected as they have no structural similarity to faces.

Figure \ref{fig:embedding_result} shows the embedding results consist of one example for each image class in the collected test dataset. It can be observed that the embedded Obama face is of very high perceptual quality and faithfully reproduces the input. However, it is noted that the embedded face is slightly smoothed and minor details are absent.

Going beyond faces, interestingly, we find that although the StyleGAN generator is trained on a human face dataset, the embedding algorithm is capable to go far beyond human faces.
As Figure \ref{fig:embedding_result} shows, although slightly worse than those of human faces, we can obtain reasonable and relatively high-quality embeddings of cats, dogs and even paintings and cars. This reveals the effective embedding capability of the algorithm and the generality of the learned filters of the generator.

Another interesting question is how the quality of the pre-trained latent space affects the embedding. To conduct these tests we also used StyleGANs trained on cars, cats, ... The quality of these results is significantly lower, as shown in supplementary materials.

\subsection{How Robust is the Embedding of Face Images?}

\paragraph{Affine Transformation}

\begin{table}[t]
    \centering
    \begin{tabular}{ l r r }
    \toprule
        Transformation & $L$($\times 10^5$) & $\| w^* - \bar{\mathbf{w}} \|$ \\ \hline
        Translation (Right 140 pixels) &    0.782 & 48.56 \\
        Translation (Left 160 pixels) &   0.406 & 44.12 \\ 
        Zoom out (2X) &    0.225 & 38.04 \\
        Zoom in (2X) &   0.718 & 40.55 \\ 
         90$^\circ$ Rotation &    0.622 & 47.21 \\
        180$^\circ$ Rotation &   0.599 & 42.93 \\ 

    \bottomrule
    \end{tabular}
    \caption{Embedding results of the transformed images. $L$ is the loss (Eq.\ref{eq:loss_function}) after optimization. $\| w^* - \bar{\mathbf{w}} \|$ is the distance between the latent codes $w^*$ and $\bar{\mathbf{w}}$ (Section   \ref{sec:initial_latent_code}) of the average face \cite{StyleGAN2018}.}
    \label{tb:affine_xans}
\end{table}

As Figure \ref{fig:affine_transformation} and Table \ref{tb:affine_xans} show, the performance of StyleGAN embedding is very sensitive to affine transformations (translation, resizing and rotation).
Among them, the translation seems to have the worst performance as it can fail to produce a valid face embedding.
For resizing and rotation, the results are valid faces.
However, they are blurry and lose many details, which are still worse than the normal embedding.
From these observations, we argue that the generalization ability of GANs is sensitive to affine transformation, which implies that the learned representations are still scale and position dependent to some extent.

\begin{figure}[h]
\includegraphics[width=0.49\textwidth]{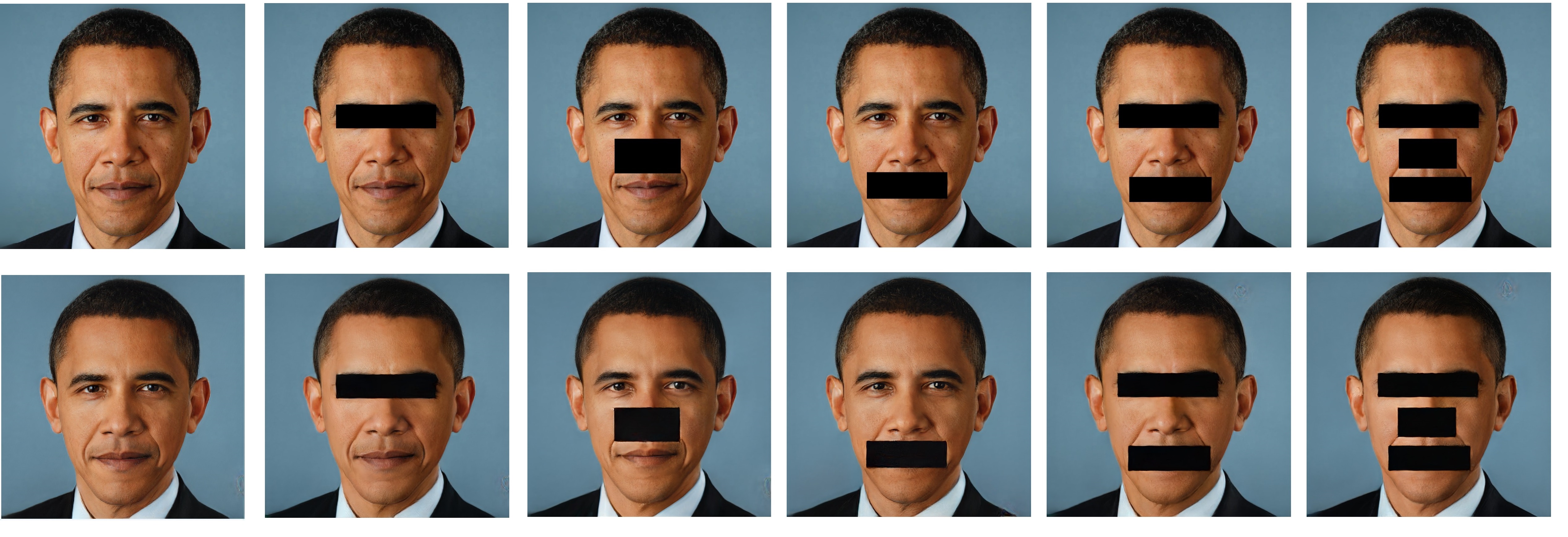}

\caption{Stress test results on defective image embedding. Top row: the input images. Bottom row: the embedded results.}

\label{fig:defective_images}
\end{figure}
\paragraph{Embedding Defective Images}

As Figure \ref{fig:defective_images} shows, the performance of StyleGAN embedding is quite robust to defects in images.
It can be observed that the embeddings of different facial features are independent of each other.
For example, removing the nose does not have an obvious influence on the embedding of the eyes and the mouth.
On the one hand, this phenomenon is good for general image editing applications.
On the other hand, it shows that the latent space does not force the embedded image to be a complete face, i.e. it does not inpaint the missing information.
\begin{figure}[t]
\includegraphics[width=0.99\linewidth]{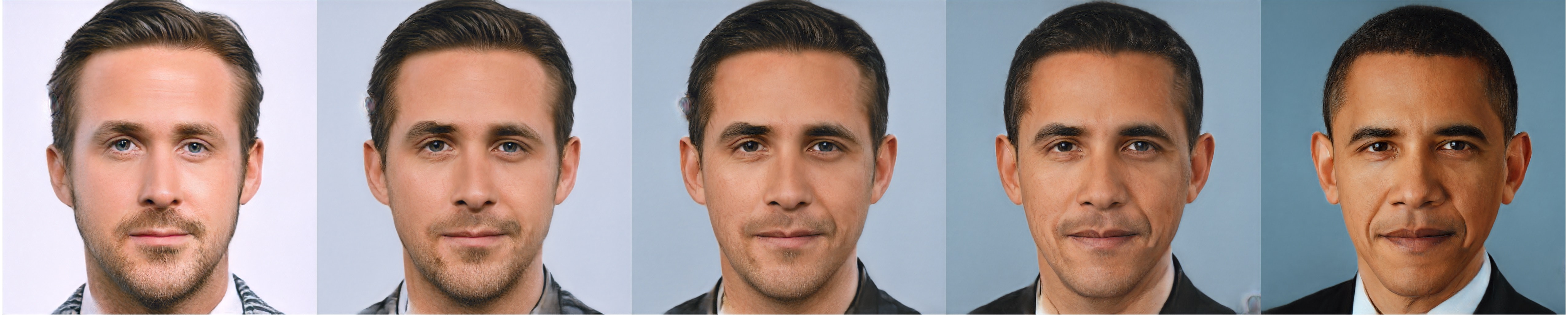}
\includegraphics[width=0.99\linewidth]{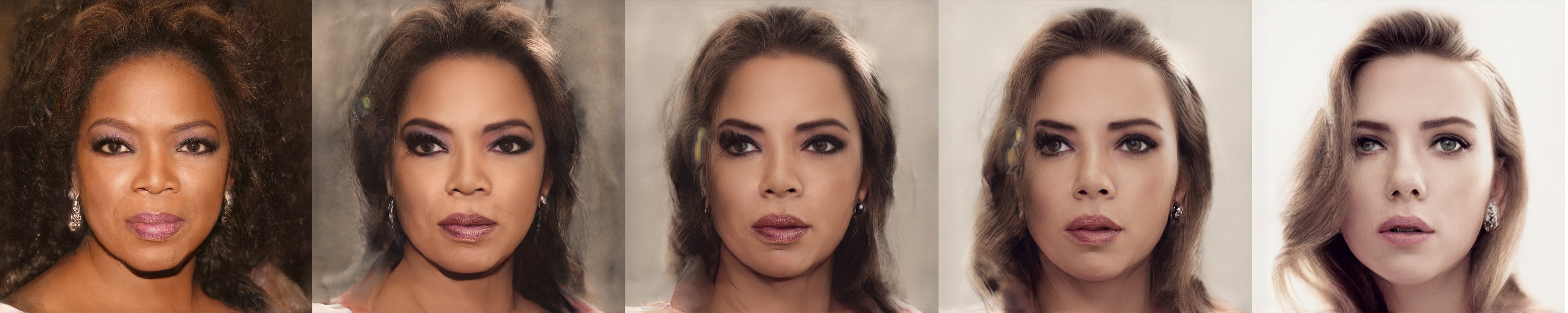}
\includegraphics[width=0.99\linewidth]{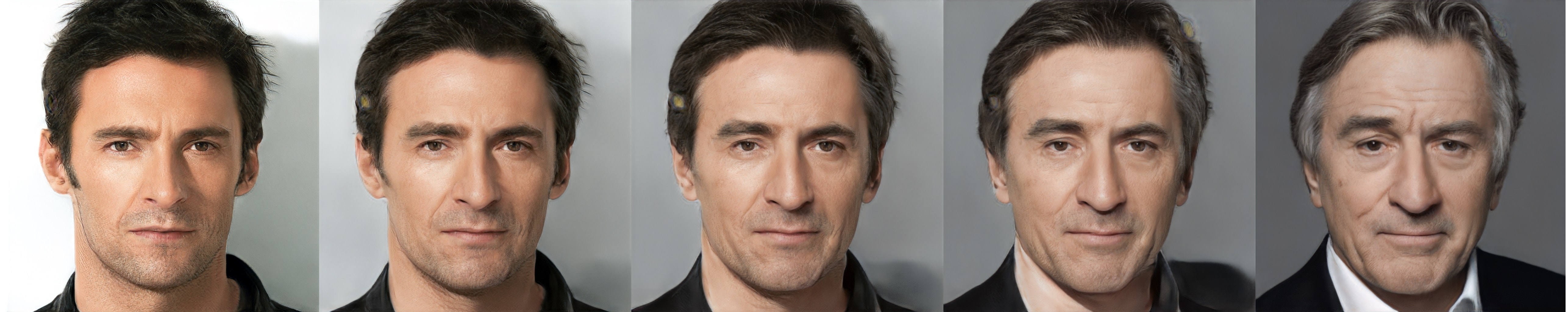}
\includegraphics[width=0.99\linewidth]{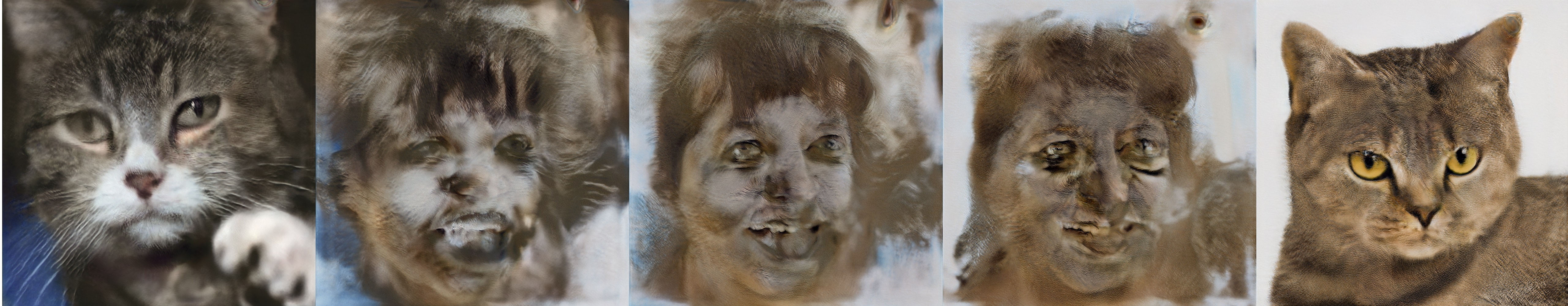}
\includegraphics[width=0.99\linewidth]{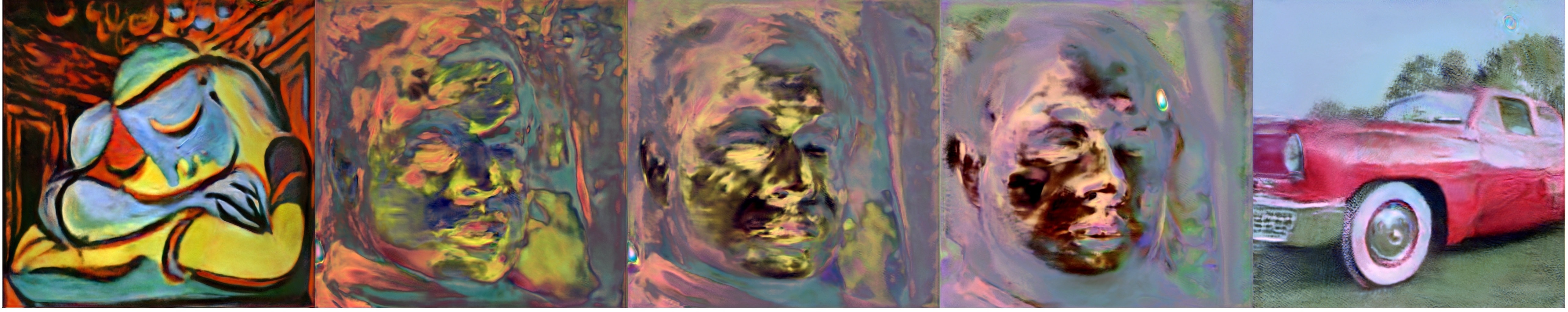}
\caption{Morphing between two embedded images (the left-most and right-most ones).}
    \label{fig:morph}
\end{figure}
\subsection{Which Latent Space to Choose?}
\label{subsec:whichlatentspace}
\begin{figure*}[t]
\includegraphics[width=\textwidth]{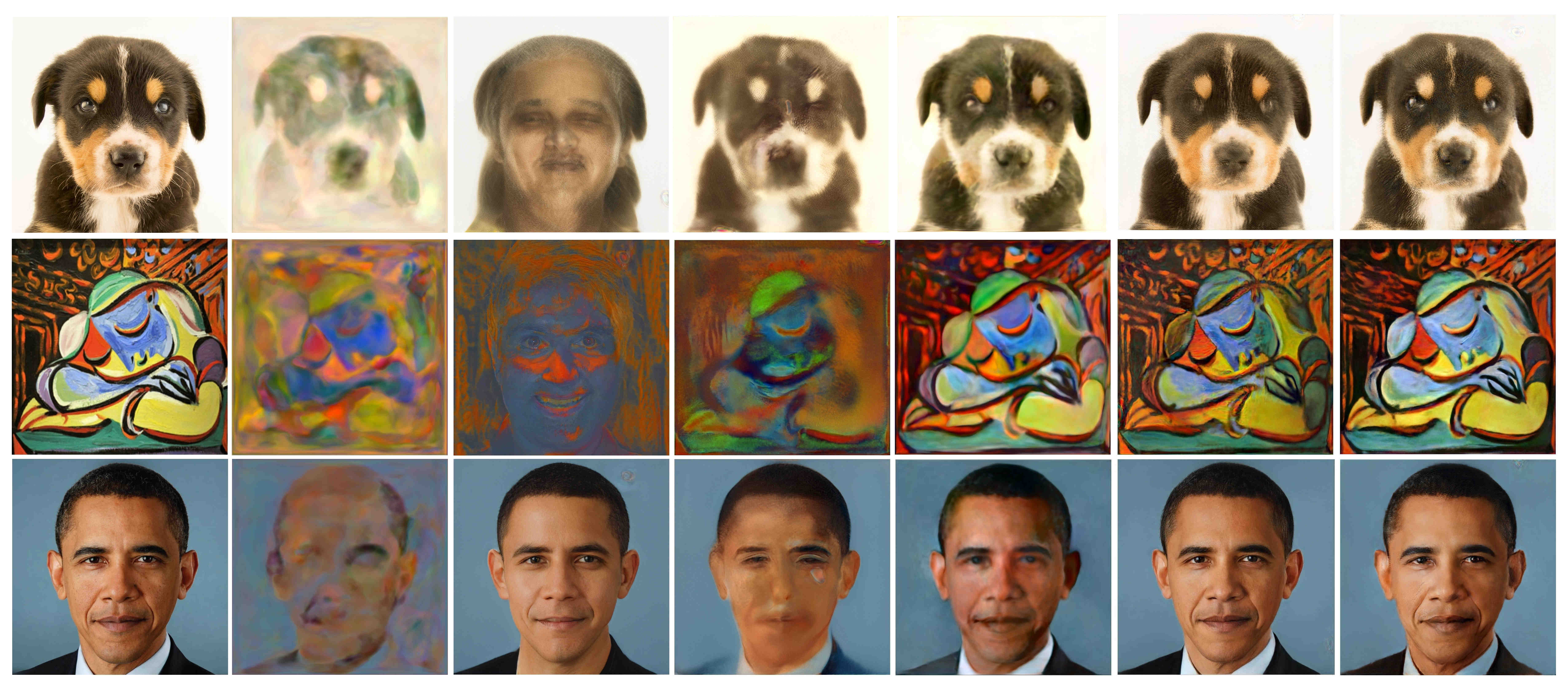}
\hspace*{1.1cm}(a)\hspace{2.1cm}(b)\hspace{2.1cm}(c)\hspace{2.1cm}(d)\hspace{2.1cm}(e)\hspace{2.1cm}(f)\hspace{2.1cm}(g)
\caption{(a) Original images. Embedding results into the original space $W$: (b) using random weights in the network layers; (c) with $\bar{\mathbf{w}}$ initialization; (d) with random initialization. Embedding results into the $W+$ space: (e) using random weights in the network layers; (f) with $\bar{\mathbf{w}}$ initialization; (g) with random initialization.}
\label{fig:w_space}
\end{figure*}

There are multiple latent spaces in StyleGAN~\cite{StyleGAN2018} that could be used for an embedding. 
Two obvious candidates are the initial latent space $Z$ and the intermediate latent space $W$. The $512$-dimensional vectors $w \in W$ are obtained from the $512$-dimensional vectors $z \in Z$ by passing them through a fully connected neural network. An important insight of our work is that it is not easily possible to embed into $W$ or $Z$ directly. Therefore, we propose to embed into an extended latent space $W^{+}$. $W^{+}$ is a concatenation of $18$ different $512$-dimensional $w$  vectors, one for each layer of the StyleGAN architecture that can receive input via AdaIn. 
As shown in Figure \ref{fig:w_space} (c)(d), embedding into $W$ directly does not give reasonable results.
Another interesting question is how important the learned network weights are for the result. We answer this question in Figure \ref{fig:w_space} (b)(e) by showing an embedding into a network that is simply initialized with random weights.

\section{How Meaningful is the Embedding?}

We propose three tests to evaluate if an embedding is semantically meaningful. Each of these tests can be conducted by simple latent code manipulations of vectors $w_i$ and these tests correspond to semantic image editing applications in computer vision and computer graphics: morphing, expression transfer, and style transfer. We consider a test successful if the resulting manipulation results in high quality images.

\subsection{Morphing}
\label{sec:morphing}

Image morphing is a longstanding research topic in computer graphics and computer vision, e.g.~\cite{MorphingSurvey1998,MorphAttack2019,seibold2017detection,steyvers1999morphing,yang2012face,korshunov2013using}).
Given two embedded images with their respective latent vectors $w_1$ and $w_2$, morphing is computed by a linear interpolation, $w = \lambda w_1 + (1-\lambda) w_2, \lambda \in (0,1)$, and subsequent image generation using the new code $w$. 
As Figure \ref{fig:morph} shows, our method generates high-quality morphing between face images (row 1,2,3) but fails on non-face images in both in-class (row 4) and inter-class (row 5) morphing.
Interestingly, it can be observed that there are contours of human faces in the intermediate images of the inter-class morphing, which shows that the latent space structure of this StyleGAN is dedicated to human faces. We therefore conjecture that non-face images are actually embedded the following way. The initial layers create a face like structure but the later layers paint over this structure so that it is no longer recognizable. While an extensive study of morphing itself is beyond the scope of this paper, we believe that the face morphing results are excellent and might be superior to the current state of the art. We leave this investigation to future work.
\begin{figure}[h]
\includegraphics[width=0.49\textwidth]{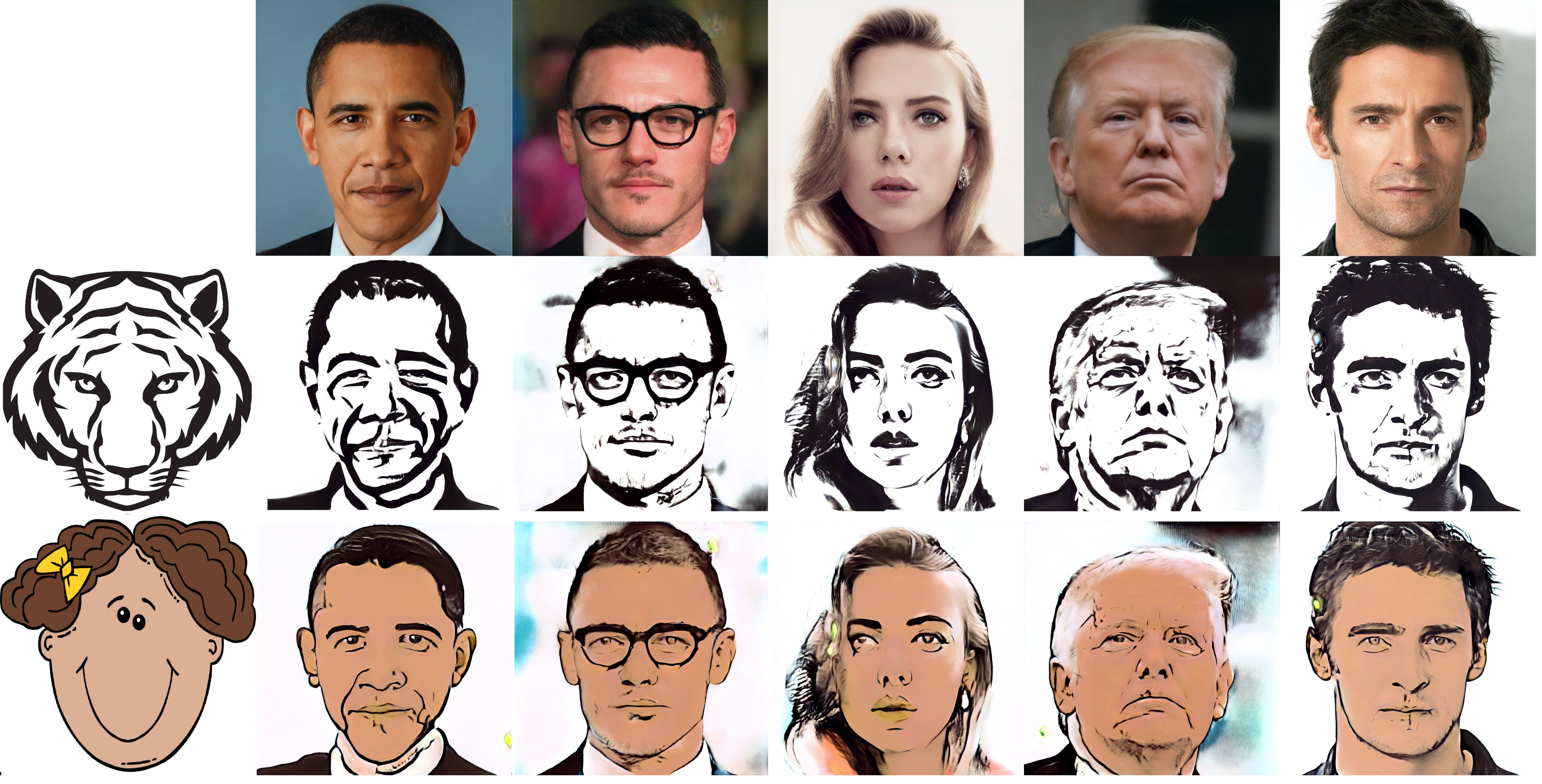}

 \caption{First column: style image; Second column: embedded stylized image using style loss from $conv4\_2$ layer of VGG-16; Third to Sixth column: style transfer by replacing latent code of last 9 layers of base image with the embedded style image.}
    \label{fig:style2}
\end{figure}
\begin{figure*}[t]
\centering
\begin{subfigure}{0.49\textwidth}
    \includegraphics[width=0.99\linewidth]{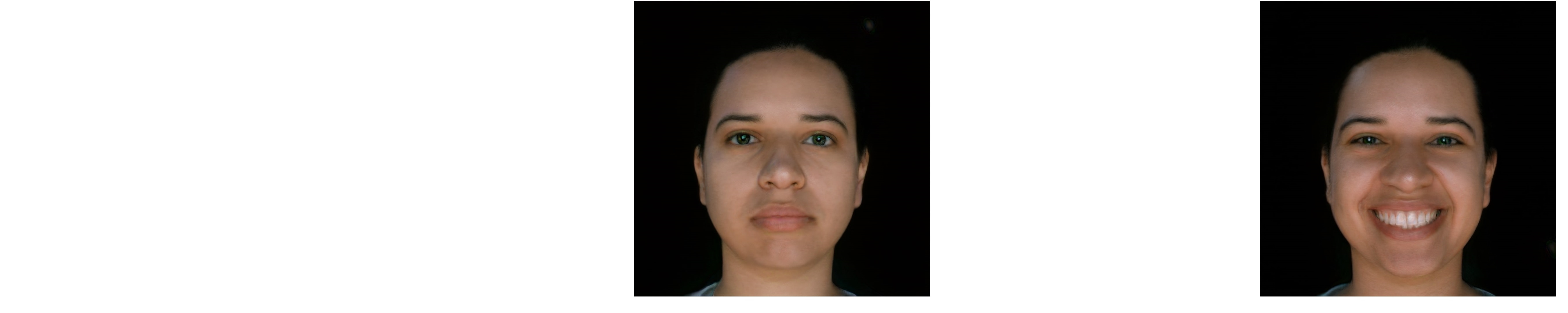}
    \includegraphics[width=0.99\linewidth]{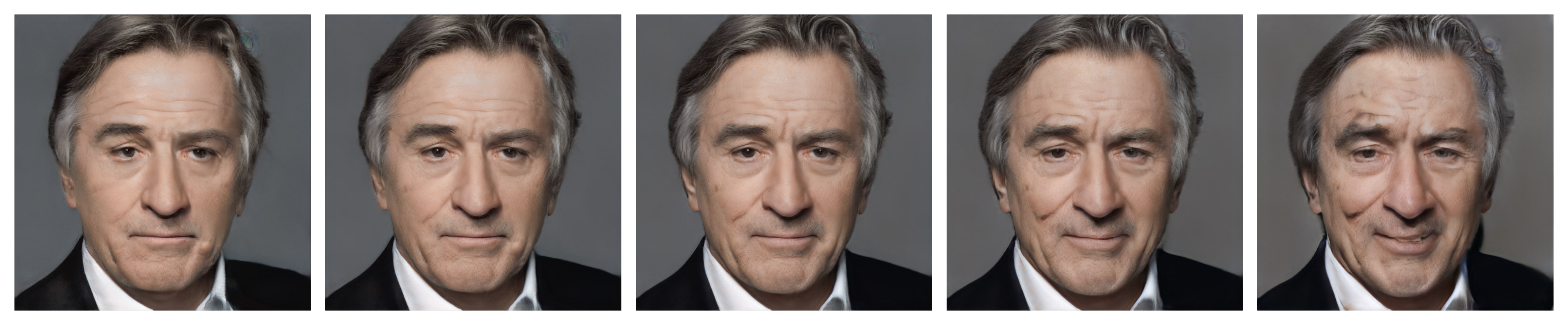}
    \includegraphics[width=0.99\linewidth]{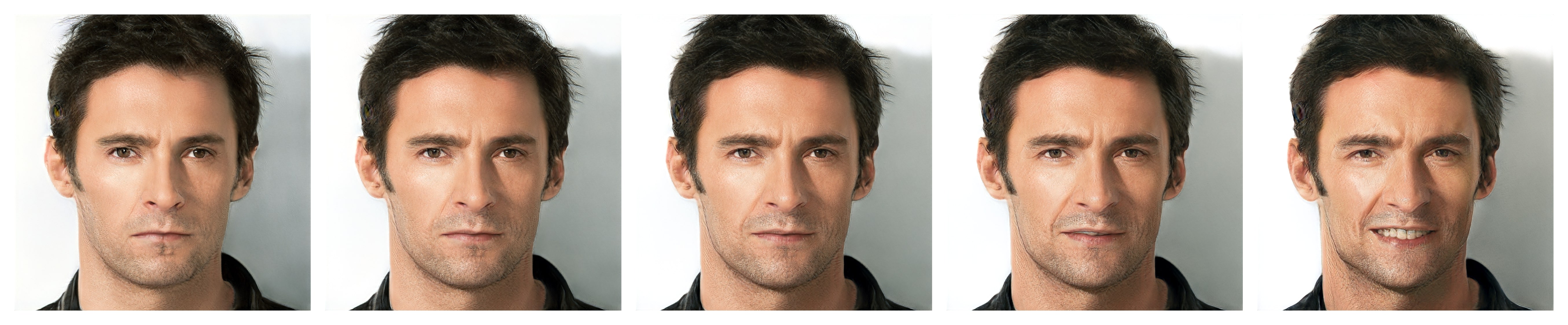}
    \includegraphics[width=0.99\linewidth]{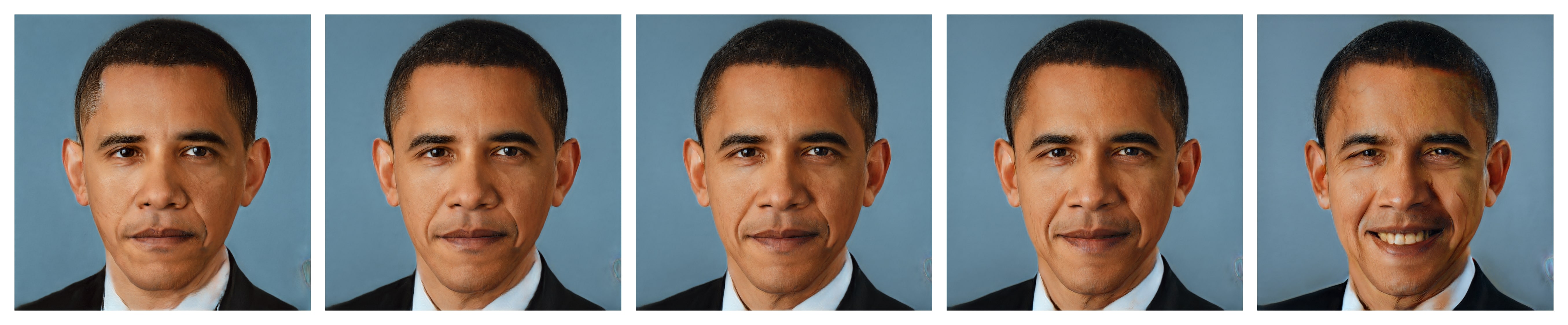}
    \includegraphics[width=0.99\linewidth]{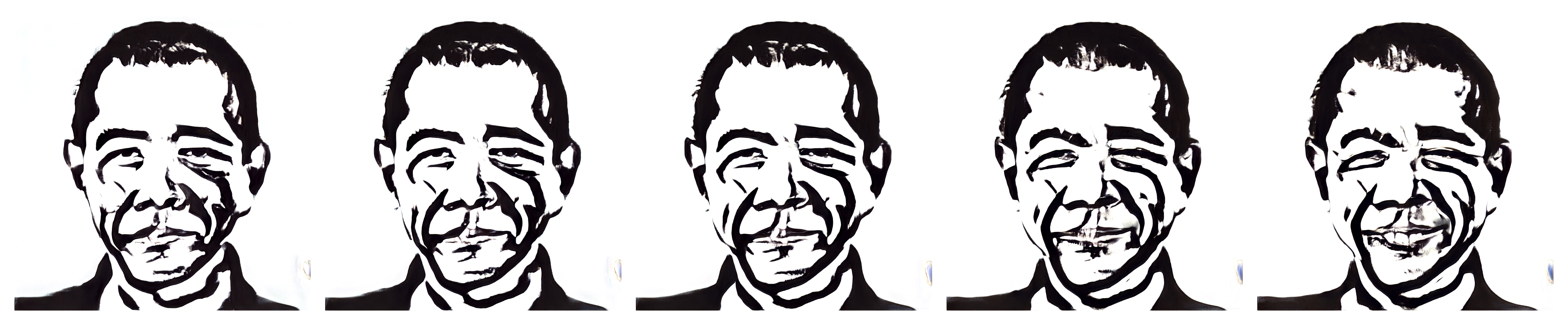}
\end{subfigure}
\begin{subfigure}{0.49\textwidth}
    \includegraphics[width=0.99\linewidth]{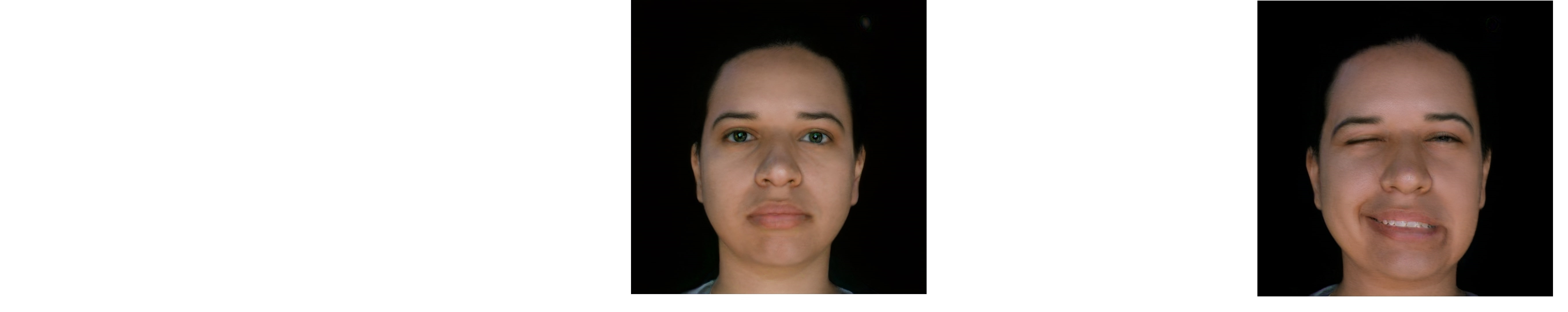}
    \includegraphics[width=0.99\linewidth]{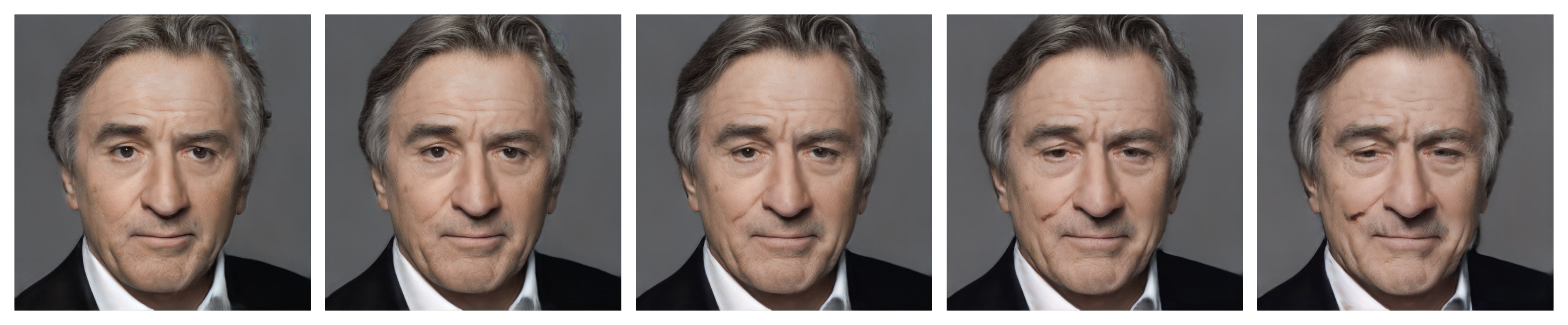}
    \includegraphics[width=0.99\linewidth]{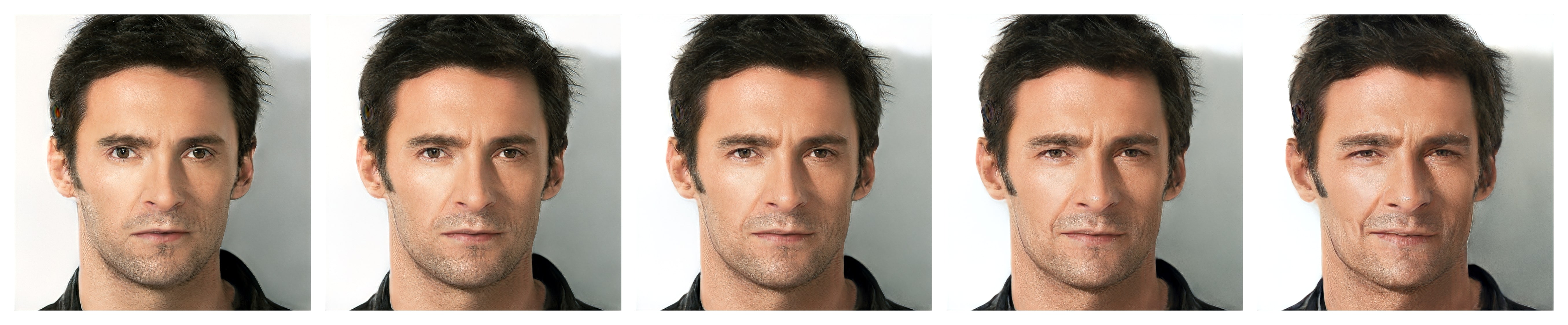}
    \includegraphics[width=0.99\linewidth]{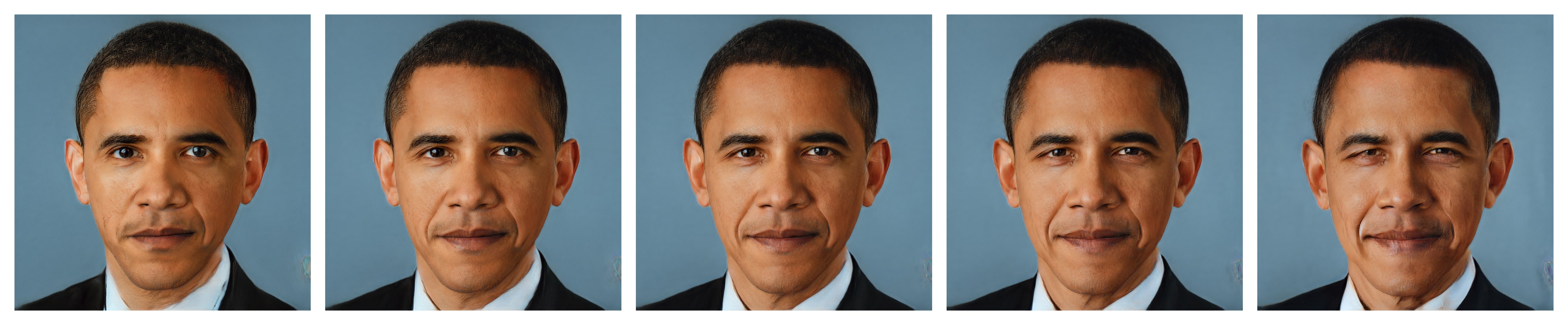}
    \includegraphics[width=0.99\linewidth]{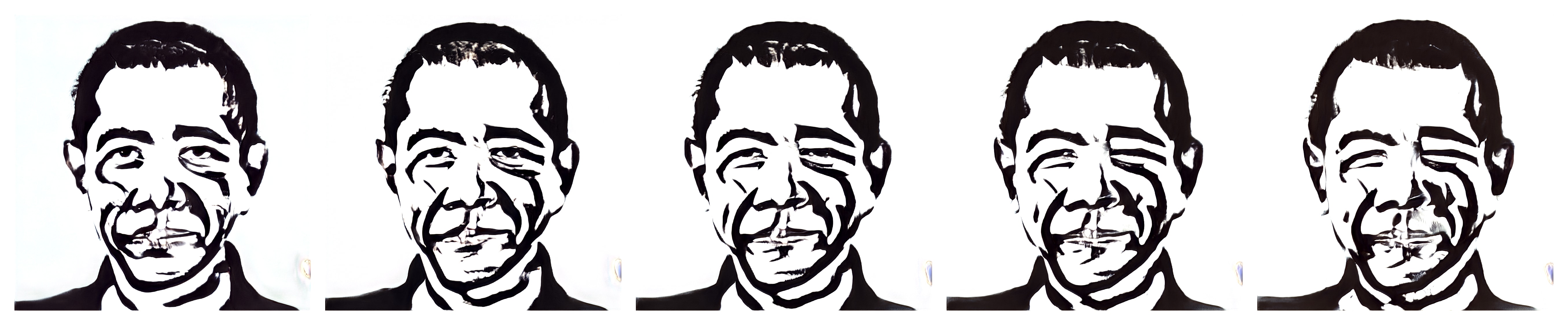}
\end{subfigure}

\caption{Results on expression transfer.
The first row shows the reference images from IMPA-FACES3D \cite{IMPA_FACES3D2011} dataset. In the following rows, the middle image in each of the examples is the embedded image, whose expression is gradually transferred to the reference expression (on the right) and the opposite direction (on the left) respectively. More results are included in the supplementary material.}
\label{fig:expression_transfer_1}
\end{figure*}

\begin{figure}[h]
    \centering
    \includegraphics[width=0.90\linewidth]{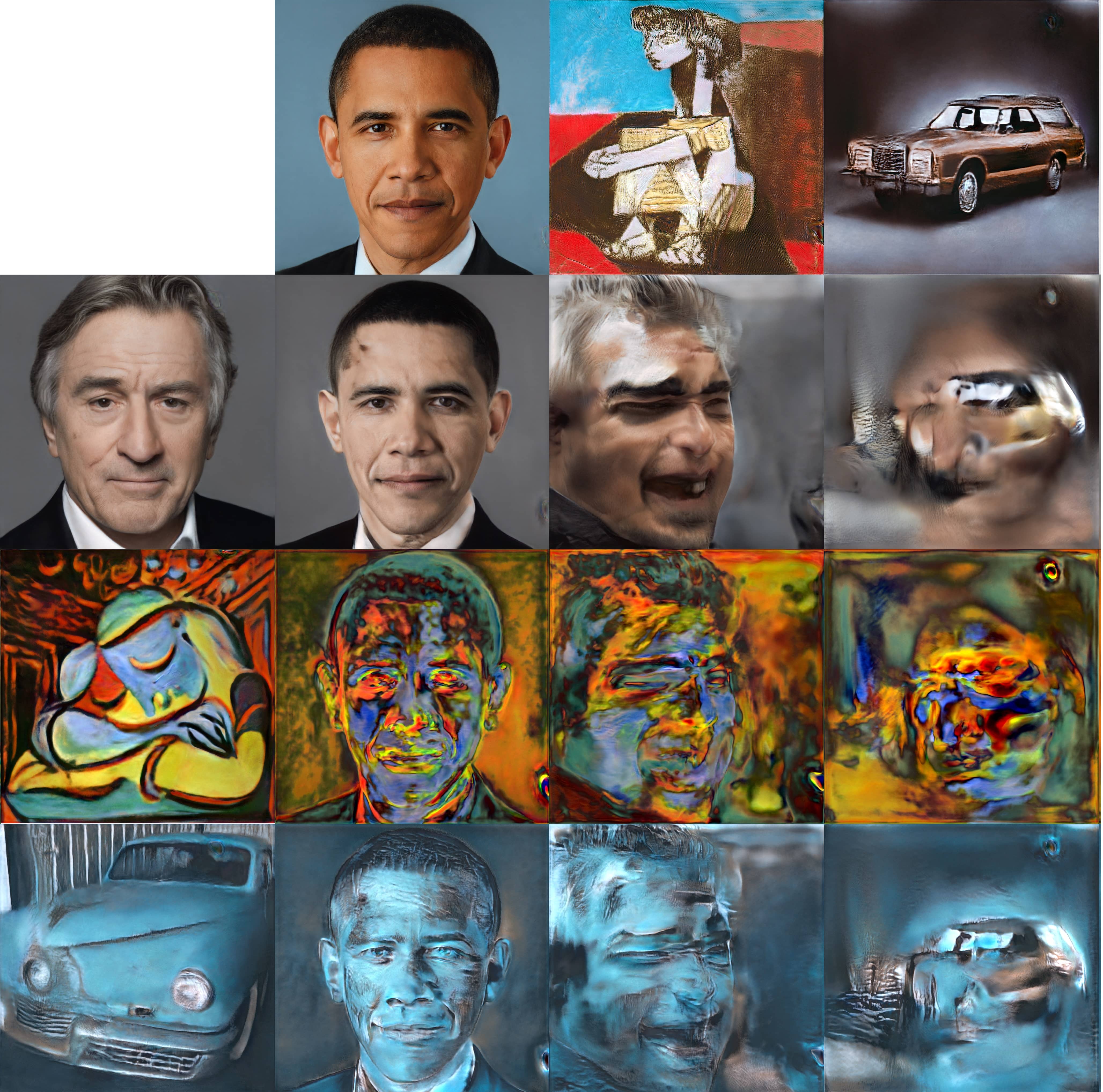}
    \caption{Style transfer between the embedded style image (first column) and the embedded content images (first row).}
    \label{fig:style4}
\end{figure}

\subsection{Style Transfer}
Given two latent codes $w_1$ and $w_2$, style transfer is computed by a crossover operation~\cite{StyleGAN2018}.
We show the style transfer results between an embedded stylized image and other face images (Figure \ref{fig:style2}) and between embedded images from different classes (Figure \ref{fig:style4}).

More specifically in Figure \ref{fig:style4}, we retain the latent codes of the embedded content image for the first $9$ layers (corresponding to spatial resolution $4^2 - 64^2$) and override the latent codes with the ones of the style image for the last $9$ layers (corresponding to spatial resolution $64^2 - 1024^2$).
Our method is able to transfer the low level features (\eg colors and textures) but fails to faithfully maintain the content structure of non-face images (second column Figure \ref{fig:style4}), especially the painting.
This phenomenon reveals that the generalization and expressing power of StyleGAN is more likely to reside in the style layers corresponding to higher spatial resolutions.

\subsection{Expression Transfer and Face Reenactment}

Given three input vectors $w_1, w_2, w_3$, expression transfer is computed as $w = w_1 + \lambda(w_3-w_2)$, where $w_1$ is the latent code of the target image, $w_2$ corresponds to a neutral expression of the source image, and $w_3$ corresponds to a more distinct expression.
For example, $w_3$ could correspond to a smiling face and $w_2$ to an expressionless face of the same person. To eliminate the noise (\eg background noise), we heuristically set a lower bound threshold on the $L2-norm$ of the channels of difference latent code, below which, the channel is replaced by a zero vector. For the above experiment, the selected value of the threshold is 1. We normalize the resultant vectors to control the intensity of an expression in a particular direction. Such code is relatively independent of the source faces and can be used to transfer expressions (Figure \ref{fig:expression_transfer_1}).
We believe that these expression transfer results are also of very high quality. Additional results are available in supplementary materials and the accompanying video.

\section{Embedding Algorithm}

Our method follows a straightforward optimization framework \cite{GANEmbedding2018} to embed a given image onto the manifold of the pre-trained generator. Starting from a suitable initialization $w$, we search for an optimized vector $w^*$ that minimizes the loss function that measures the similarity between the given image and the image generated from $w^*$.
Algorithm \ref{alg:latent_space_embedding} shows the pseudo-code of our method.
An interesting aspect of this work is that not all design choices lead to good results and that experimenting with the design choices provides further insights into the embedding.

\begin{algorithm}[h]
\SetAlgoLined
 \KwIn{An image $I \in \mathbb{R}^{n \times m \times 3}$ to embed; a pre-trained generator $G(\cdot)$.}
 \KwOut{The embedded latent code $w^*$ and the embedded image $G(w^*)$ optimzed via $F'$.}
 Initialize latent code $w^*$ = $w$\;
 \While{not converged}{
  $L \leftarrow L_{percept}(G(w^*), I) + \frac{\lambda}{N}\|G(w^*) - I\|_2^2$ \;
  $w^* \leftarrow w^* - \eta F'( \nabla_{w^*} L$ )\;
 }
 \caption{Latent Space Embedding for GANs}
 \label{alg:latent_space_embedding}
\end{algorithm}

\subsection{Initialization}
\label{sec:initial_latent_code}

We investigate two design choices for the initialization. The first choice is random initialization. In this case, each variable is sampled independently from a uniform distribution $\mathcal{U}[-1,1]$.
The second choice is motivated by the observation that the distance to the mean latent vector $\bar{\mathbf{w}}$ can be used to identify low quality faces~\cite{StyleGAN2018}.
Therefore, we propose to use $\bar{\mathbf{w}}$ as initialization and expect the optimization to converge to a vector $w^*$ that is closer to $\bar{\mathbf{w}}$.

\begin{table}[t]
    \centering
    \begin{tabular}{ l l r r }
    \toprule
        Data class & $w$ Init. & $L$($\times 10^5$) & $\| w^* - \bar{\mathbf{w}} \|$ \\ \hline
        \multirow{2}{*}{Face} & $w = \bar{\mathbf{w}}$ &   \textbf{0.309} & \textbf{30.67} \\ 
                             & Random & 0.351 & 35.60 \\ \hline
        \multirow{2}{*}{Cat} & $w = \bar{\mathbf{w}}$ & 0.752 & 70.86 \\ 
                             & Random & \textbf{0.740} & 70.97 \\
        \multirow{2}{*}{Dog} & $w = \bar{\mathbf{w}}$ & 0.922 & 74.78 \\ 
                             & Random & \textbf{0.845} & 75.14 \\
        \multirow{2}{*}{Painting} & $w = \bar{\mathbf{w}}$ &  3.530 & 103.61 \\ 
                             & Random & \textbf{3.451} & 105.29 \\
        \multirow{2}{*}{Car} & $w = \bar{\mathbf{w}}$ & 1.390 & 82.53 \\ 
                             & Random & \textbf{1.269} & 82.60 \\ 
    \bottomrule
    \end{tabular}
    \caption{Algorithmic choice justification on the latent code initialization. $w$ Init. is the initialization method for the latent code $w$. $L$ is the mean of the loss (Eq.\ref{eq:loss_function}) after optimization. $\| w^* - \bar{\mathbf{w}} \|$ is the distance between the latent codes $w^*$ and $\bar{\mathbf{w}}$ of the average face \cite{StyleGAN2018}.}
    \label{tb:latent_code_initialization_quantatitive_results}
\end{table}

To evaluate these two design choices, we compared the loss values and the distance $\| w^* - \bar{\mathbf{w}} \|$ between the optimized latent code $w^*$ and $\bar{\mathbf{w}}$ after optimization.
As Table \ref{tb:latent_code_initialization_quantatitive_results} shows, initializing $w = \bar{\mathbf{w}}$ for face image embeddings not only makes the optimized $w^*$ closer to $\bar{\mathbf{w}}$, but also achieves a much lower loss value.
However, for images in other classes (\eg dog), random initialization proves to be the better option.
Intuitively, the phenomenon suggests that the distribution has only one cluster of faces, the other instances (\eg dogs, cats) are scattered points surrounding the cluster without obvious patterns.
Qualitative results are shown in Figure \ref{fig:w_space} (f)(g).

\subsection{Loss Function}
\label{sec:loss_function}

To measure the similarity between the input image and the embedded image during optimization, we employ a loss function that is a weighted combination of the VGG-16 perceptual loss \cite{PerceptualLoss2016} and the pixel-wise MSE loss:
\begin{equation} 
    w^* = \min_w L_{percept}(G(w), I) + \frac{\lambda_{mse}}{N} \|G(w) - I\|_2^2
    \label{eq:loss_function}
\end{equation}
where $I \in \mathbb{R}^{n \times n \times 3}$ is the input image, $G(\cdot)$ is the pre-trained generator, $N$ is the number of scalars in the image (\ie $N = n \times n \times 3$), $w$ is the latent code to optimize, $\lambda_{mse}=1$ is empirically obtained for good performance.
For the perceptual loss term $L_{percept}(\cdot)$ in Eq.\ref{eq:loss_function}, we use:
\begin{equation} 
    L_{percept} (I_1, I_2) = \sum_{j=1}^{4} \frac{\lambda_j}{N_j} \| F_j (I_1)-F_j (I_2) \|_2^2
    \label{eq:perceptual_loss}
\end{equation}
where $I_1, I_2 \in \mathbb{R}^{n \times n \times 3}$ are the input images, $F_j$ is the feature output of VGG-16 layers $‘conv1\_1’$, $‘conv1\_2’$, $‘conv3\_2’$ and $‘conv4\_2’$ respectively, $N_j$ is the number of scalars in the $jth$ layer output, $\lambda_j=1$ for all $j$s are empirically obtained for good performance.

Our choice of the perceptual loss together with the pixel-wise MSE loss comes from the fact that the pixel-wise MSE loss alone cannot find a high quality embedding. The perceptual loss therefore acts as some sort of regularizer to guide the optimization into the right region of the latent space.

\begin{figure}[t]
\centering
\includegraphics[width=0.49\textwidth]{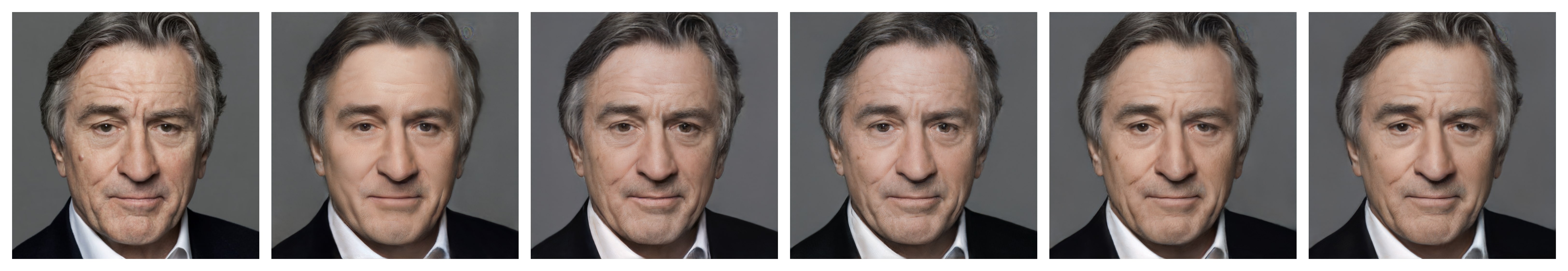}
\includegraphics[width=0.49\textwidth]{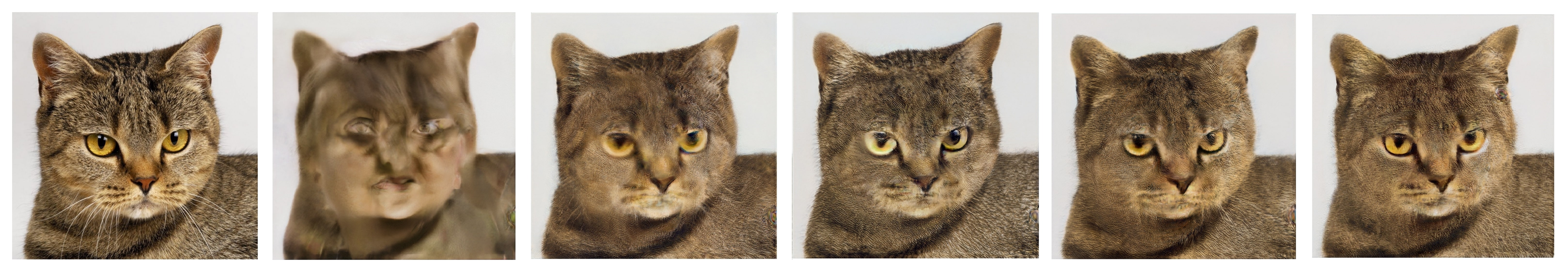}
\includegraphics[width=0.49\textwidth]{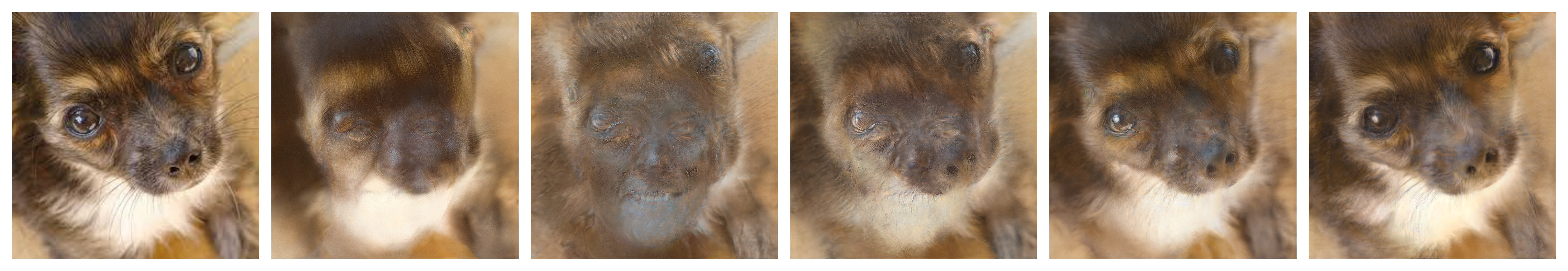}
\includegraphics[width=0.49\textwidth]{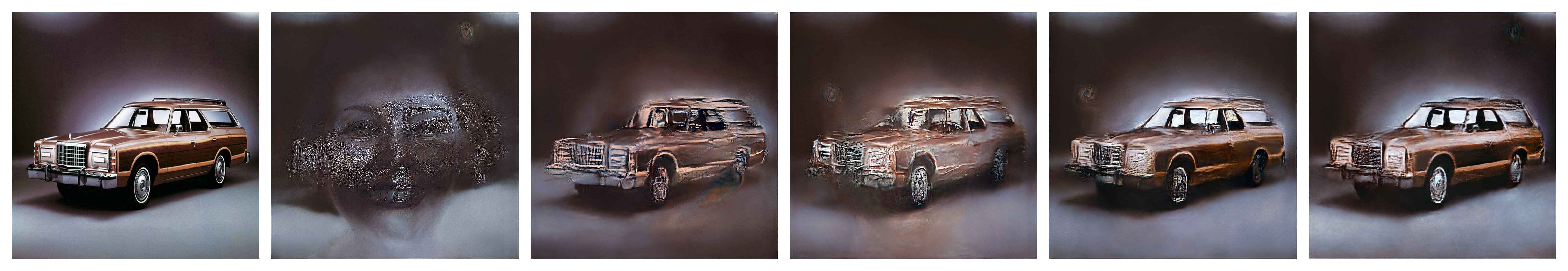}
\includegraphics[width=0.49\textwidth]{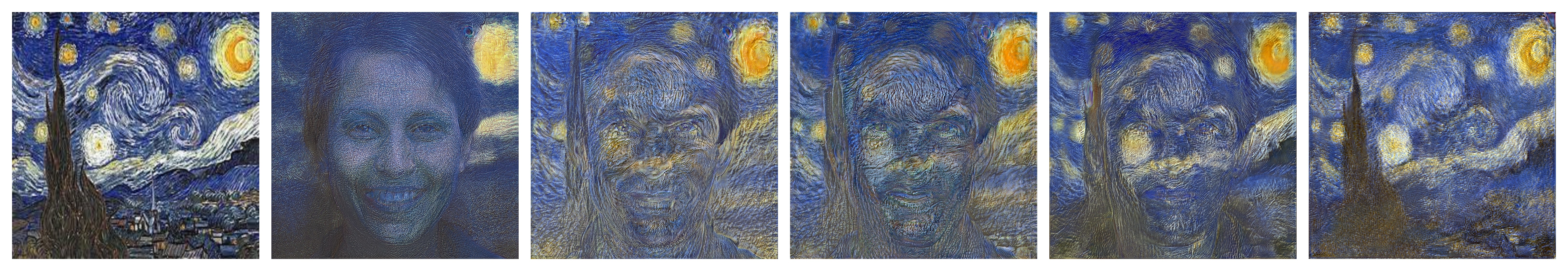}
 \caption{Algorithmic choice justification on the loss function. Each row shows the results of an image from the five different classes in our test dataset respectively.
 From left to right, each column shows: (1) the original image; (2) pixel-wise MSE loss only; (3) perceptual loss on VGG-16 $conv3\_2$ layer only; (4) pixel-wise MSE loss and VGG-16 $conv3\_2$; (5) perceptual loss (Eq.\ref{eq:perceptual_loss}) only; (6) our loss function (Eq.\ref{eq:loss_function}). More results are included in the supplementary material.}
    \label{fig:algorithmic_choice_loss_function_1}
\end{figure}

We perform an ablation study to justify our choice of loss function in Eq.\ref{eq:loss_function}.
As Figure \ref{fig:algorithmic_choice_loss_function_1} shows, using the pixel-wise MSE loss term alone (column 2) embeds the general colors well but fails to catch the features of non-face images. 
In addition, it has a smoothing effect that does not preserve the details even for the human faces. 
Interestingly, due to the pixel-wise MSE loss working in the pixel space and ignoring the differences in feature space, its embedding results on non-face images (\eg the car and the painting) have a tendency towards the average face of the pre-trained StyleGAN \cite{StyleGAN2018}.
This problem is addressed by the perceptual losses (column 3, 5) that measures image similarity in the feature space. Since our embedding task requires the embedded image to be close to the input at all scales, we found that matching the features at multiple layers of the VGG-16 network (column 5) works better than using only a single layer (column 3).
This further motivates us to combine the pixel-wise MSE loss with the perceptual loss (column 4, 6) from that the pixel-wise MSE loss can be viewed as the lowest level perceptual loss at pixel scale.
Column 6 of Figure \ref{fig:algorithmic_choice_loss_function_1} shows the embedding results of our final choice (pixel-wise MSE + multi-layer perceptual loss), which achieves the best performance among different algorithmic choices.

\subsection{Other Parameters} 

We use the Adam optimizer with a learning rate of $0.01$, $\beta_1=0.9$, $\beta_2=0.999$, and $\epsilon=1e^{-8}$ in all our experiments.
We use 5000 gradient descent steps for the optimization, taking less than $7$ minutes per image on a 32GB Nvidia TITAN V100 GPU.

\begin{figure}[t]
\includegraphics[width=0.99\linewidth]{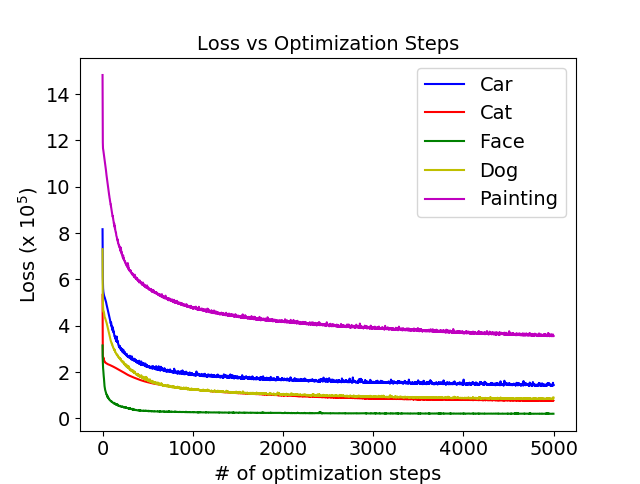}
 \caption{Loss values \vs the number of optimization steps.}
    \label{fig:convergence}
\end{figure}

To justify our choice of $5000$ optimization steps, we investigated the change in the loss function as a function of the number of iterations.
As Figure \ref{fig:convergence} shows, the loss value of the human face image drops the quickest and converges at around $1000$ optimization steps; those of the cat, the dog and the car images converge slower at around $3000$ optimization steps; while the painting curve is the slowest and converges around $5000$ optimization steps.
We choose to optimize the loss function for $5000$ steps in all our experiments.

\paragraph{Iterative Embedding}

\begin{figure}[h]
\includegraphics[width=0.49\textwidth]{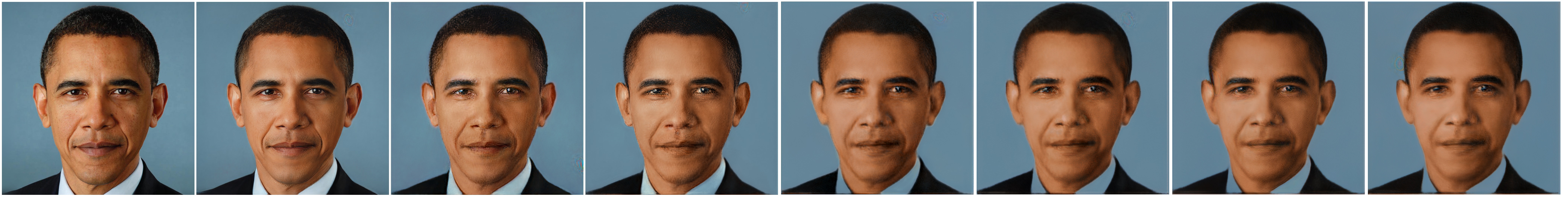}
\includegraphics[width=0.49\textwidth]{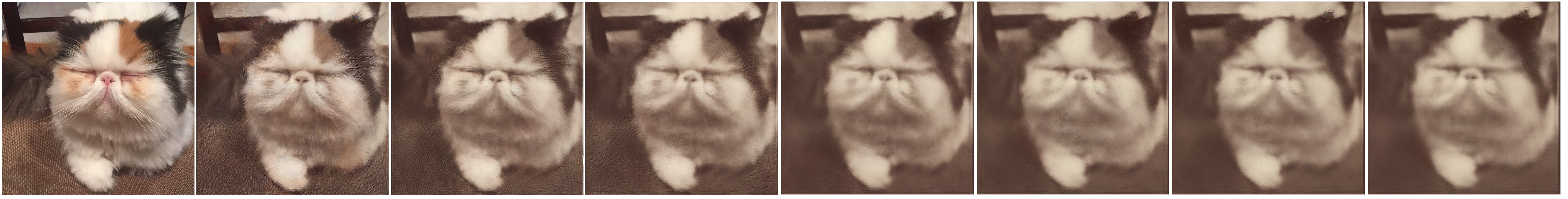}
 \caption{Stress test results on iterative embedding. The left most column shows the original images and the subsequent columns are the results of iterative embedding.}
    \label{fig:iterative_embedding}
\end{figure}

We tested the robustness of the proposed method on iterative embedding, \ie we iteratively take the embedding results as new input images and do the embedding again. This process is repeated seven times. 
As Figure \ref{fig:iterative_embedding} shows, although it is guaranteed that the input image exists in the model distribution after the first embedding, the performance of the proposed method slowly degenerates (more details are lost) with the number of iterative embedding.
The reason for this observation may be that the employed optimization approach suffers from slow convergence around local optimum.
For the embeddings other than human faces, the stochastic initial latent codes may also be a factor for the degradation.
In summary, these observations show that our embedding approach can reach reasonably ``good'' embeddings on the model distribution easily, although ``perfect'' embeddings are hard to reach.

\section{Conclusion}

We proposed an efficient algorithm to embed a given image into the latent space of StyleGAN. This algorithm enables semantic image editing operations, such as image morphing, style transfer, and expression transfer.
We also used the algorithm to study multiple aspects of the StyleGAN latent space. We proposed experiments to analyze what type of images can be embedded, how they are embedded, and how meaningful the embedding is. Important conclusions of our work are that embedding works best into the extended latent space $W+$ and that any type of image can be embedded. However, only the embedding of faces is semantically meaningful.

Our framework still has several limitations. First, we inherit image artifacts present in pre-trained StyleGAN that we illustrate in supplementary materials. Second, the optimization takes several minutes and an embedding algorithm that can work in under a second would be more appealing for interactive editing.

In future work, we hope to extend our framework to process videos in addition to static images. Further, we would like to explore embeddings into GANs trained on three-dimensional data, such as point clouds or meshes.

\noindent \textbf{Acknowledgement} This work was supported by the KAUST Office of Sponsored Research (OSR) under Award No. OSR-CRG2017-3426.

{\small
\bibliographystyle{ieee}
\bibliography{egbib}

\begin{thebibliography}{10}\itemsep=-1pt

\bibitem{alharbi2018latent}
Yazeed Alharbi, Neil Smith, and Peter Wonka.
\newblock Latent filter scaling for multimodal unsupervised image-to-image
  translation.
\newblock {\em arXiv preprint arXiv:1812.09877}, 2018.

\bibitem{WGAN2017}
Martin Arjovsky, Soumith Chintala, and L{\'e}on Bottou.
\newblock {W}asserstein generative adversarial networks.
\newblock In {\em Proceedings of the 34th International Conference on Machine
  Learning}, volume~70, pages 214--223, 2017.

\bibitem{BigGAN2019}
Andrew Brock, Jeff Donahue, and Karen Simonyan.
\newblock Large scale {GAN} training for high fidelity natural image synthesis.
\newblock In {\em International Conference on Learning Representations}, 2019.

\bibitem{GANEmbedding2018}
A. {Creswell} and A.~A. {Bharath}.
\newblock Inverting the generator of a generative adversarial network.
\newblock {\em IEEE Transactions on Neural Networks and Learning Systems},
  2018.

\bibitem{dosovitskiy2016generating}
Alexey Dosovitskiy and Thomas Brox.
\newblock Generating images with perceptual similarity metrics based on deep
  networks.
\newblock In {\em Advances in neural information processing systems}, pages
  658--666, 2016.

\bibitem{GatysStyle2015}
L.~A. Gatys, A.~S. Ecker, and M. Bethge.
\newblock A neural algorithm of artistic style.
\newblock {\em arXiv}, Aug 2015.

\bibitem{GatysTexture2015}
Leon~A. Gatys, Alexander~S. Ecker, and Matthias Bethge.
\newblock Texture synthesis using convolutional neural networks.
\newblock In {\em Proceedings of the 28th International Conference on Neural
  Information Processing Systems - Volume 1}, NIPS'15, 2015.

\bibitem{goodfellow2014generative}
Ian Goodfellow, Jean Pouget-Abadie, Mehdi Mirza, Bing Xu, David Warde-Farley,
  Sherjil Ozair, Aaron Courville, and Yoshua Bengio.
\newblock Generative adversarial nets.
\newblock In {\em Advances in neural information processing systems}, 2014.

\bibitem{WGANGP2017}
Ishaan Gulrajani, Faruk Ahmed, Martin Arjovsky, Vincent Dumoulin, and Aaron~C
  Courville.
\newblock Improved training of wasserstein gans.
\newblock In {\em Advances in Neural Information Processing Systems}, pages
  5767--5777, 2017.

\bibitem{Adain2017}
Xun Huang and Serge Belongie.
\newblock Arbitrary style transfer in real-time with adaptive instance
  normalization.
\newblock In {\em ICCV}, 2017.

\bibitem{pix2pix2017}
Phillip Isola, Jun-Yan Zhu, Tinghui Zhou, and Alexei~A Efros.
\newblock Image-to-image translation with conditional adversarial networks.
\newblock {\em CVPR}, 2017.

\bibitem{PerceptualLoss2016}
Justin Johnson, Alexandre Alahi, and Li Fei-Fei.
\newblock Perceptual losses for real-time style transfer and super-resolution.
\newblock In {\em European conference on computer vision}, 2016.

\bibitem{johnson2016perceptual}
Justin Johnson, Alexandre Alahi, and Li Fei-Fei.
\newblock Perceptual losses for real-time style transfer and super-resolution.
\newblock In {\em European conference on computer vision}, pages 694--711.
  Springer, 2016.

\bibitem{ProgressiveGAN2018}
Tero Karras, Timo Aila, Samuli Laine, and Jaakko Lehtinen.
\newblock Progressive growing of {GAN}s for improved quality, stability, and
  variation.
\newblock In {\em International Conference on Learning Representations}, 2018.

\bibitem{StyleGAN2018}
Tero Karras, Samuli Laine, and Timo Aila.
\newblock A style-based generator architecture for generative adversarial
  networks.
\newblock {\em arXiv preprint arXiv:1812.04948}, 2018.

\bibitem{VAE2013}
Diederik~P Kingma and Max Welling.
\newblock Auto-encoding variational bayes.
\newblock {\em arXiv preprint arXiv:1312.6114}, 2013.

\bibitem{korshunov2013using}
Pavel Korshunov and Touradj Ebrahimi.
\newblock Using face morphing to protect privacy.
\newblock In {\em 2013 10th IEEE International Conference on Advanced Video and
  Signal Based Surveillance}, pages 208--213. IEEE, 2013.

\bibitem{Alexnet2012}
Alex Krizhevsky, Ilya Sutskever, and Geoffrey~E Hinton.
\newblock Imagenet classification with deep convolutional neural networks.
\newblock In {\em Advances in Neural Information Processing Systems 25}. 2012.

\bibitem{laine2018feature-based}
Samuli Laine.
\newblock Feature-based metrics for exploring the latent space of generative
  models, 2018.

\bibitem{TextureSynthesis2016}
Chuan Li and Michael Wand.
\newblock Precomputed real-time texture synthesis with markovian generative
  adversarial networks.
\newblock In {\em Computer Vision - {ECCV} 2016 - 14th European Conference,
  Amsterdam, The Netherlands, October 11-14, 2016, Proceedings, Part {III}},
  2016.

\bibitem{ObjectDetection2017}
Jianan Li, Xiaodan Liang, Yunchao Wei, Tingfa Xu, Jiashi Feng, and Shuicheng
  Yan.
\newblock Perceptual generative adversarial networks for small object
  detection.
\newblock In {\em The IEEE Conference on Computer Vision and Pattern
  Recognition (CVPR)}, July 2017.

\bibitem{VGG2015}
S. {Liu} and W. {Deng}.
\newblock Very deep convolutional neural network based image classification
  using small training sample size.
\newblock In {\em 2015 3rd IAPR Asian Conference on Pattern Recognition
  (ACPR)}, Nov 2015.

\bibitem{LSGAN2017}
Xudong Mao, Qing Li, Haoran Xie, Raymond~Y.K. Lau, Zhen Wang, and Stephen
  Paul~Smolley.
\newblock Least squares generative adversarial networks.
\newblock In {\em The IEEE International Conference on Computer Vision (ICCV)},
  Oct 2017.

\bibitem{IMPA_FACES3D2011}
Jes{\'u}s~P Mena-Chalco, Luiz Velho, and RM~Cesar Junior.
\newblock 3d human face reconstruction using principal components spaces.
\newblock In {\em Proceedings of WTD SIBGRAPI Conference on Graphics, Patterns
  and Images}, 2011.

\bibitem{SpectralNormalization2018}
Takeru Miyato, Toshiki Kataoka, Masanori Koyama, and Yuichi Yoshida.
\newblock Spectral normalization for generative adversarial networks.
\newblock In {\em International Conference on Learning Representations}, 2018.

\bibitem{git2}
Dmitry Nikitko.
\newblock stylegan-encoder.
\newblock \url{https://github.com/Puzer/stylegan-encoder}, 2019.

\bibitem{park2019SPADE}
Taesung Park, Ming-Yu Liu, Ting-Chun Wang, and Jun-Yan Zhu.
\newblock Semantic image synthesis with spatially-adaptive normalization.
\newblock In {\em Proceedings of the IEEE Conference on Computer Vision and
  Pattern Recognition}, 2019.

\bibitem{DCGAN2015}
Alec Radford, Luke Metz, and Soumith Chintala.
\newblock Unsupervised representation learning with deep convolutional
  generative adversarial networks.
\newblock {\em arXiv preprint arXiv:1511.06434}, 2015.

\bibitem{MorphAttack2019}
Ulrich Scherhag, Christian Rathgeb, Johannes Merkle, Ralph Breithaupt, and
  Christoph Busch.
\newblock Face recognition systems under morphing attacks: A survey.
\newblock {\em IEEE Access}, 7, 2019.

\bibitem{seibold2017detection}
Clemens Seibold, Wojciech Samek, Anna Hilsmann, and Peter Eisert.
\newblock Detection of face morphing attacks by deep learning.
\newblock In {\em International Workshop on Digital Watermarking}, pages
  107--120. Springer, 2017.

\bibitem{slossberg2018high}
Ron Slossberg, Gil Shamai, and Ron Kimmel.
\newblock High quality facial surface and texture synthesis via generative
  adversarial networks.
\newblock In {\em European Conference on Computer Vision}, pages 498--513.
  Springer, 2018.

\bibitem{steyvers1999morphing}
Mark Steyvers.
\newblock Morphing techniques for manipulating face images.
\newblock {\em Behavior Research Methods, Instruments, \& Computers},
  31(2):359--369, 1999.

\bibitem{git}
Timo~Aila Tero~Karras, Samuli~Laine.
\newblock Stylegan - official tensorflow implementation.
\newblock \url{https://github.com/NVlabs/stylegan}, 2018.

\bibitem{Tulyakov_2018_CVPR}
Sergey Tulyakov, Ming-Yu Liu, Xiaodong Yang, and Jan Kautz.
\newblock Mocogan: Decomposing motion and content for video generation.
\newblock In {\em The IEEE Conference on Computer Vision and Pattern
  Recognition (CVPR)}, June 2018.

\bibitem{VideoGeneration2016}
Carl Vondrick, Hamed Pirsiavash, and Antonio Torralba.
\newblock Generating videos with scene dynamics.
\newblock In {\em Advances in Neural Information Processing Systems 29}. 2016.

\bibitem{MorphingSurvey1998}
George Wolberg.
\newblock Image morphing: a survey.
\newblock {\em The Visual Computer}, 14(8), 1998.

\bibitem{Xian_2018_CVPR}
Wenqi Xian, Patsorn Sangkloy, Varun Agrawal, Amit Raj, Jingwan Lu, Chen Fang,
  Fisher Yu, and James Hays.
\newblock Texturegan: Controlling deep image synthesis with texture patches.
\newblock In {\em The IEEE Conference on Computer Vision and Pattern
  Recognition (CVPR)}, June 2018.

\bibitem{yang2012face}
Fei Yang, Eli Shechtman, Jue Wang, Lubomir Bourdev, and Dimitris Metaxas.
\newblock Face morphing using 3d-aware appearance optimization.
\newblock In {\em Proceedings of Graphics Interface 2012}, pages 93--99.
  Canadian Information Processing Society, 2012.

\bibitem{Zhu_2016}
Jun-Yan Zhu, Philipp Krähenbühl, Eli Shechtman, and Alexei~A. Efros.
\newblock Generative visual manipulation on the natural image manifold.
\newblock {\em Lecture Notes in Computer Science}, page 597–613, 2016.

\bibitem{CycleGAN2017}
Jun-Yan Zhu, Taesung Park, Phillip Isola, and Alexei~A Efros.
\newblock Unpaired image-to-image translation using cycle-consistent
  adversarial networkss.
\newblock In {\em Computer Vision (ICCV), 2017 IEEE International Conference
  on}, 2017.

\bibitem{NIPS2017_6650}
Jun-Yan Zhu, Richard Zhang, Deepak Pathak, Trevor Darrell, Alexei~A Efros,
  Oliver Wang, and Eli Shechtman.
\newblock Toward multimodal image-to-image translation.
\newblock In I. Guyon, U.~V. Luxburg, S. Bengio, H. Wallach, R. Fergus, S.
  Vishwanathan, and R. Garnett, editors, {\em Advances in Neural Information
  Processing Systems 30}, pages 465--476. Curran Associates, Inc., 2017.

\end{thebibliography}
}

\newpage
\clearpage
\newpage

\section{Additional Materials on Embedding}
\paragraph{Dataset}
In order to test our embedding algorithm, we collect a small dataset of 25 images in five different categories: human faces, cats, dogs, cars and paintings (Figure \ref{fig:whole}).

\paragraph{Additional Embedding Results}
To further support our findings about the initial latent code in the main paper, we show more results in Figure \ref{fig:comp22}.
It can be observed that: for face images, initializing the optimization with the mean face latent code works better; while for non-face images, using the latent codes randomly sampled from a multivariate uniform distribution is a better option.

\paragraph{Quantitative Results on Defective Image Embedding}
Table \ref{tb:defect} shows the corresponding quantitative results on defective image embedding (Figure 3 in the main paper).
The results show that compared to non-defective faces, the embedded images of defective faces are farther from the mean face.
This reaffirms that the valid faces form a cluster around the mean face.

\begin{figure}[t]
    \centering
    \includegraphics[width=\linewidth]{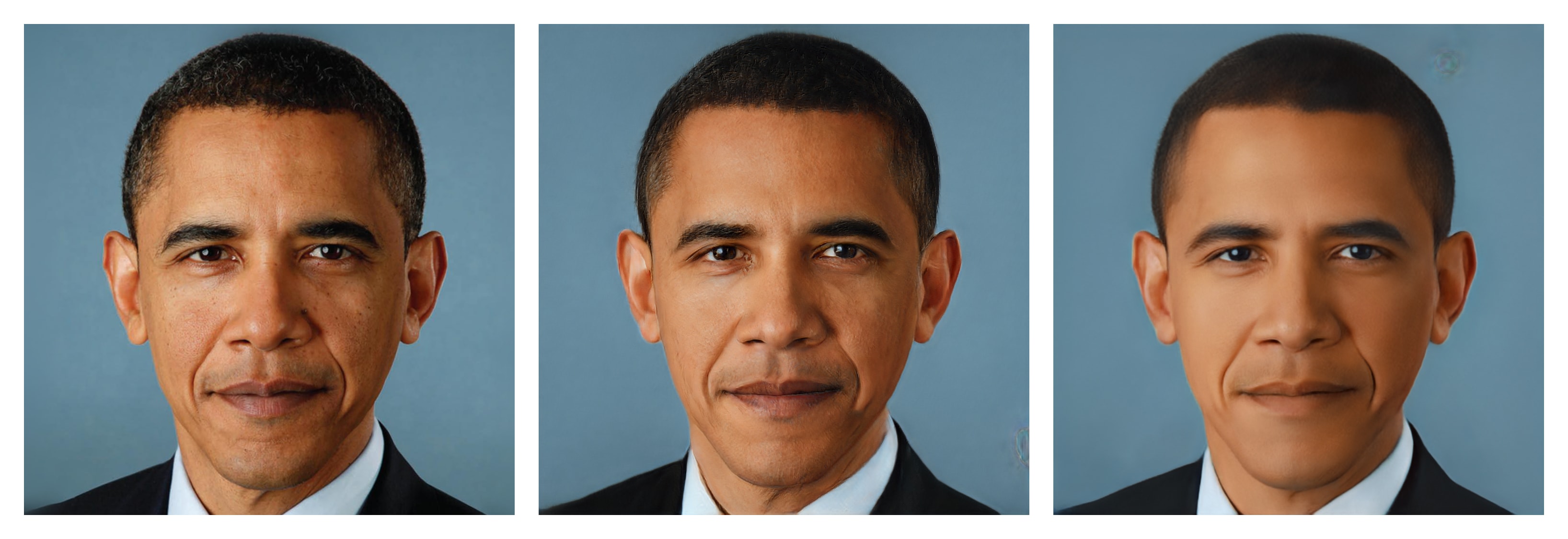}
    \includegraphics[width=\linewidth]{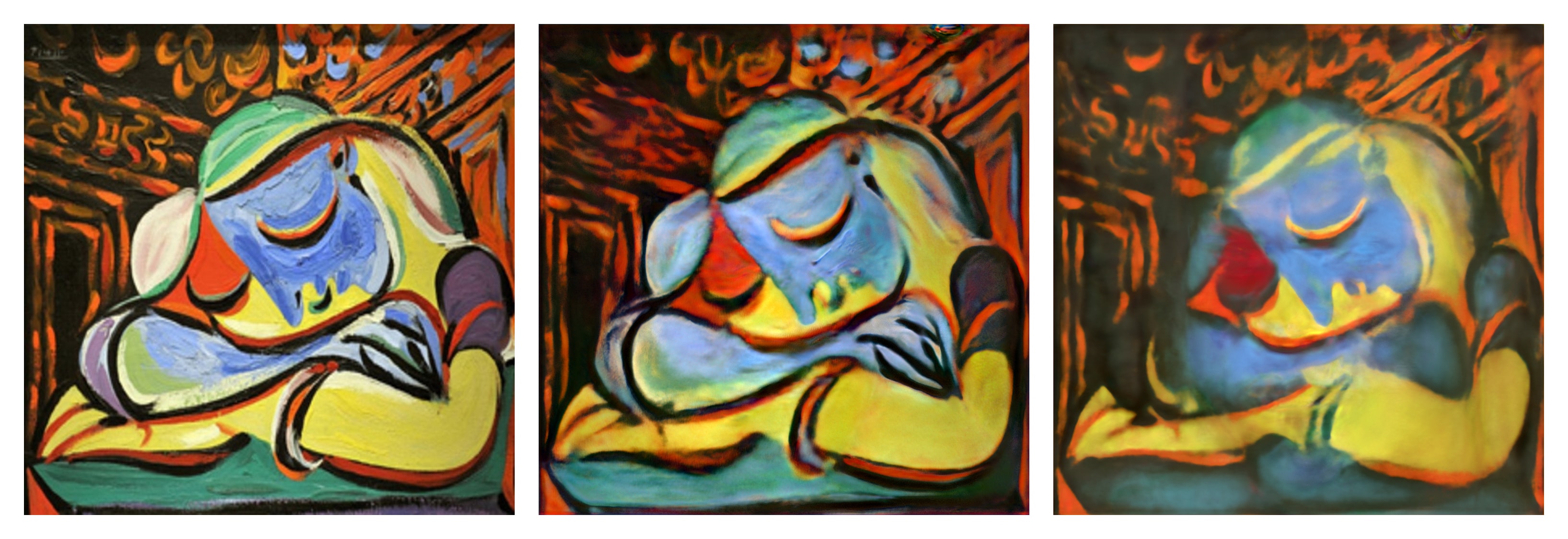}
    \includegraphics[width=\linewidth]{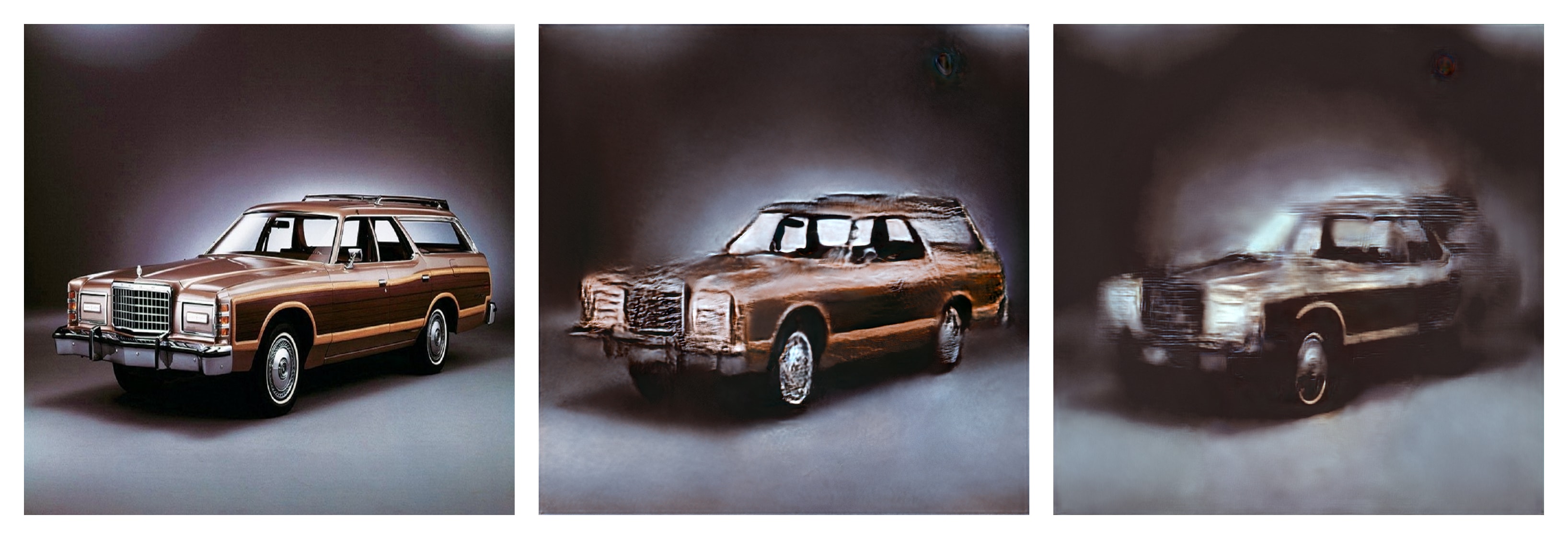}
    \caption{First column: original image ($1024 \times 1024$). Second column: embedded image with the perceptual loss applied to resized images of $256 \times 256$ resolution. Third column: embedded image with the perceptual loss applied to the images at the original $1024 \times 1024$ resolution.}
    \label{fig:comp}
\end{figure}

\paragraph{Inherent Circular Artifacts of StyleGAN}
Interestingly, we observed that the StyleGAN model trained on the FFHQ dataset (officially released \cite{StyleGAN2018,git}) inherently creates circular artifacts in the generated images, which are also observable in our embedding results (Figure \ref{fig:eye}).
These artifacts are thus independent of our embedding algorithm and may be resolved by employing better pretrained models in the future.

\paragraph{Limitation of the ImageNet-based Perceptual loss}
All existing perceptual losses utilize the classifiers trained on the ImageNet dataset (\eg VGG-16, VGG-19), which are restricted to the resolution of $224 \times 224$.
While in our paper, we aim to embed images of high resolution ($1024 \times 1024$) that are much larger than that of ImageNet images.
Such inconsistency in the resolution may disable the learned image filters as they are scale-dependent.
To this end, we follow the common practice \cite{johnson2016perceptual,laine2018feature-based} and use a simple resizing trick to compute the perceptual loss on resized images of $256 \times 256$ resolution.
As Figure \ref{fig:comp} shows, the embedding results with the resizing trick outperform the ones at the original resolution.
However, small details are lost during the resizing, which can slightly smoothen the embedding results.
We expect to get better results with future perceptual losses that work on higher resolutions.

\begin{figure*}
    \centering
\begin{subfigure}{0.35\textwidth}
   \centering
    \includegraphics[width=\linewidth]{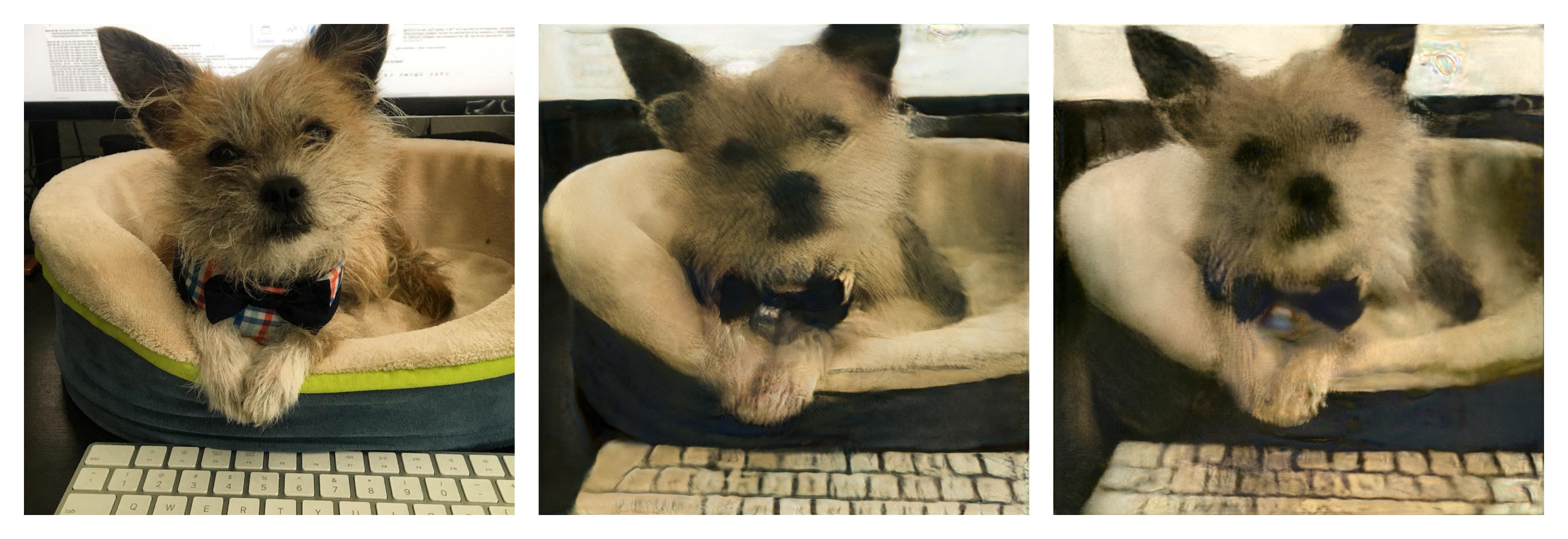}
    \includegraphics[width=\linewidth]{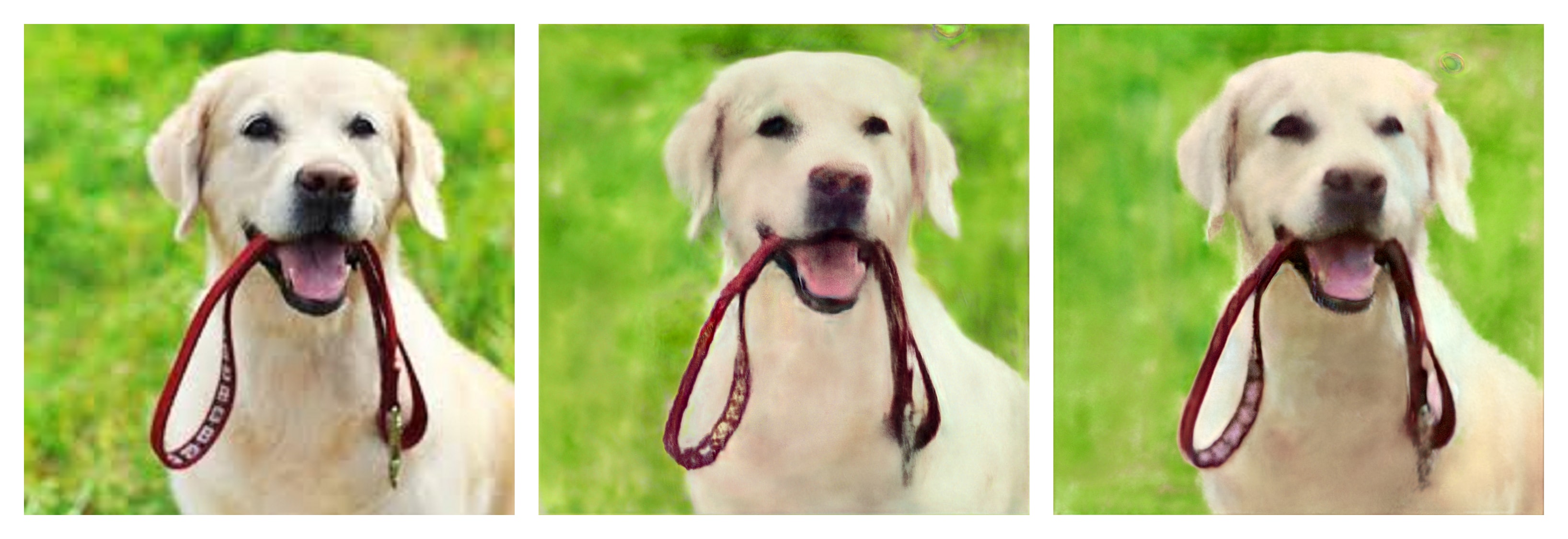}

    \includegraphics[width=\linewidth]{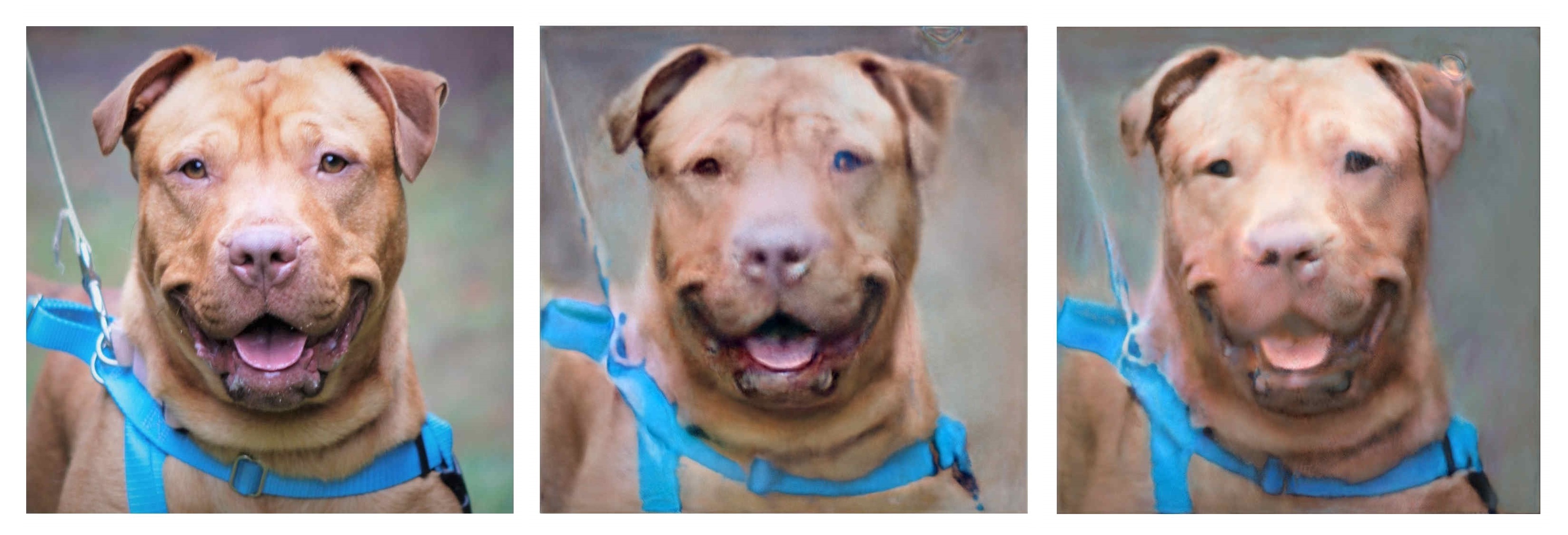}
    \includegraphics[width=\linewidth]{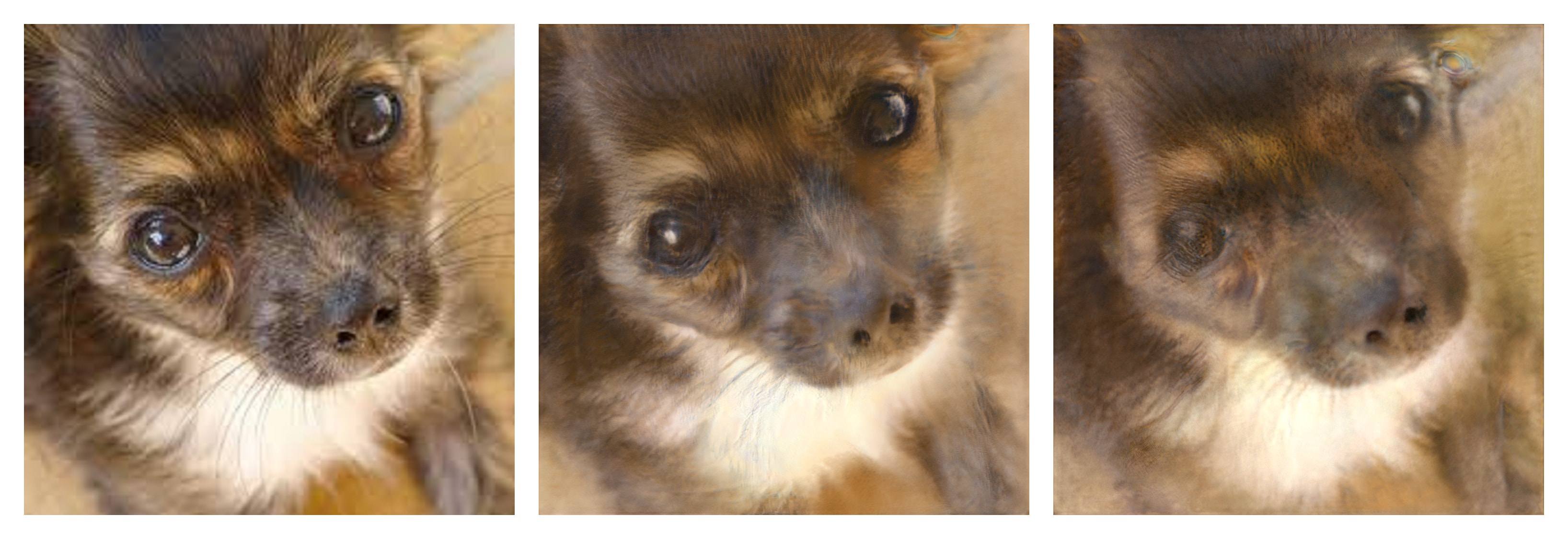}
     \includegraphics[width=\linewidth]{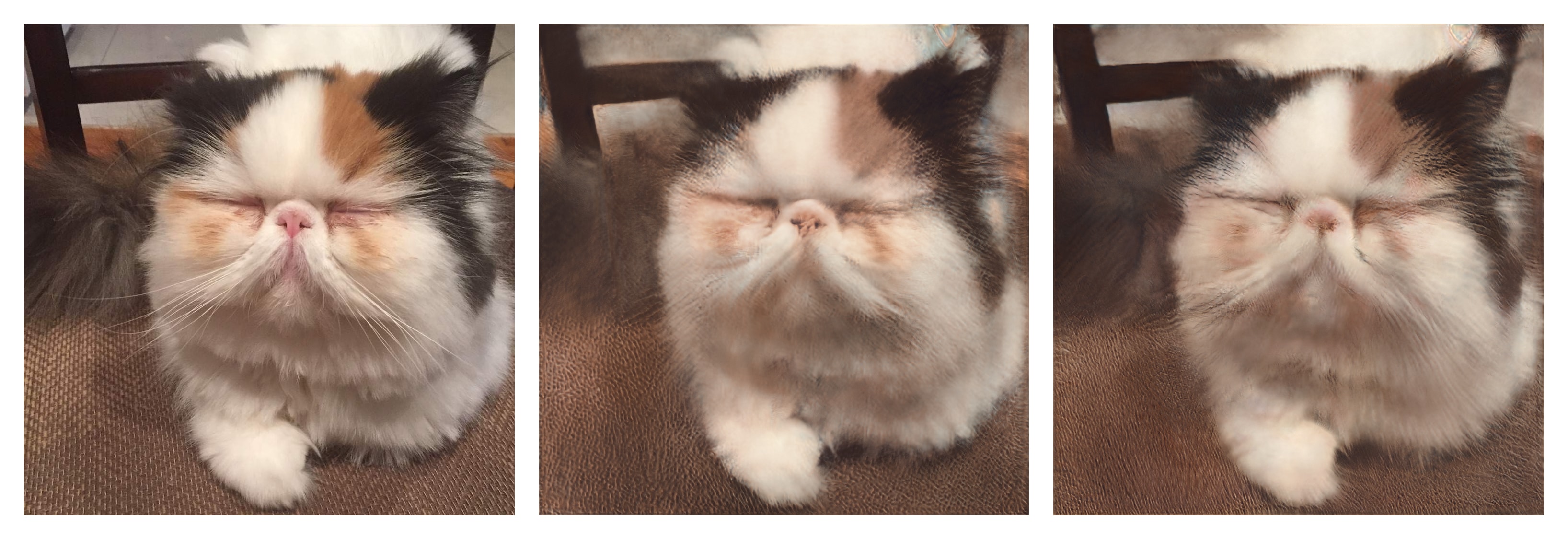}

    \includegraphics[width=\linewidth]{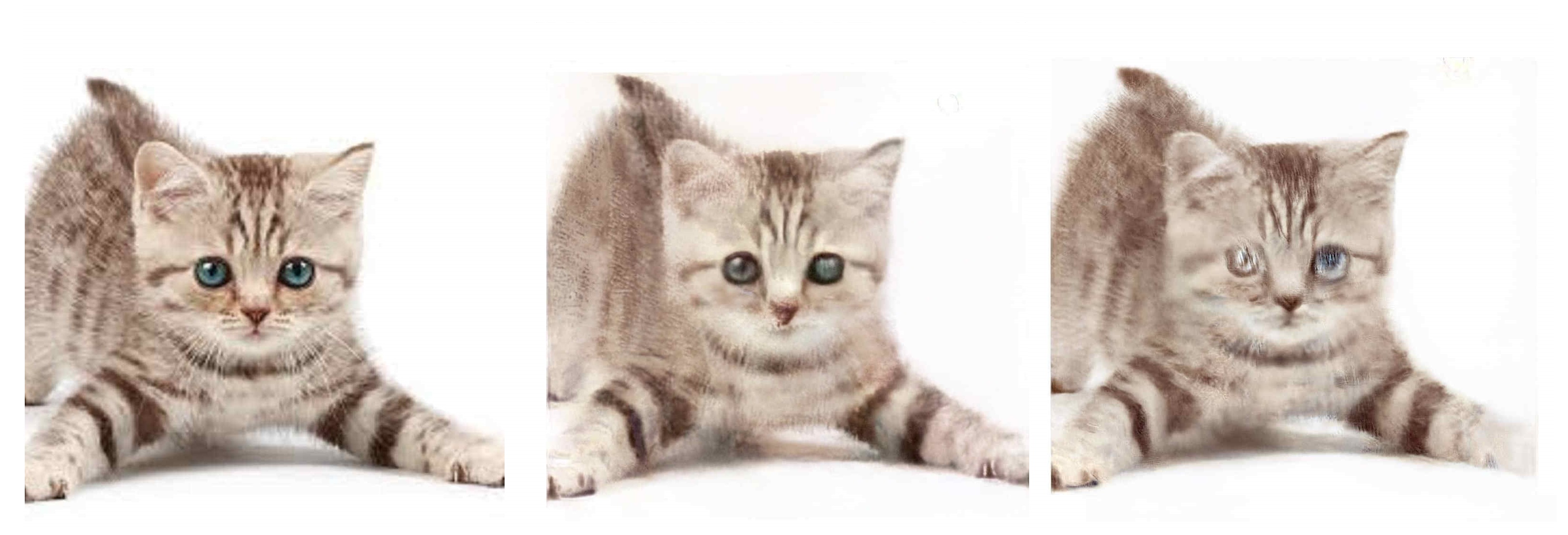}

     \includegraphics[width=\linewidth]{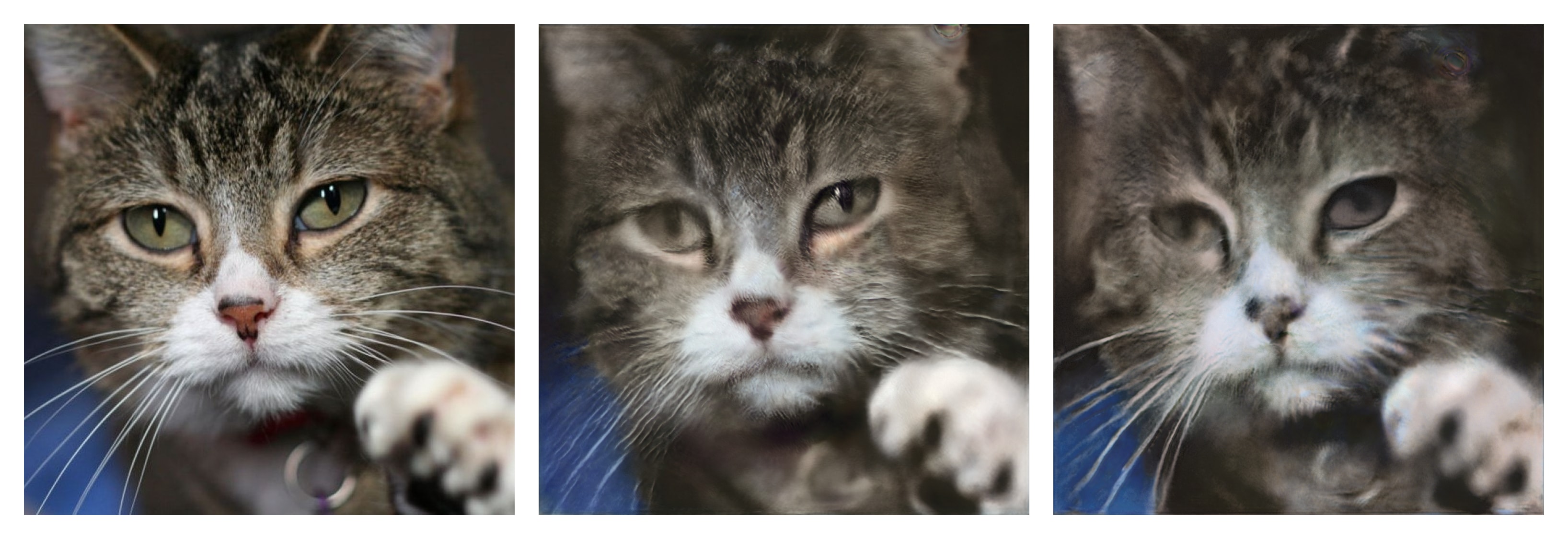}
     \includegraphics[width=\linewidth]{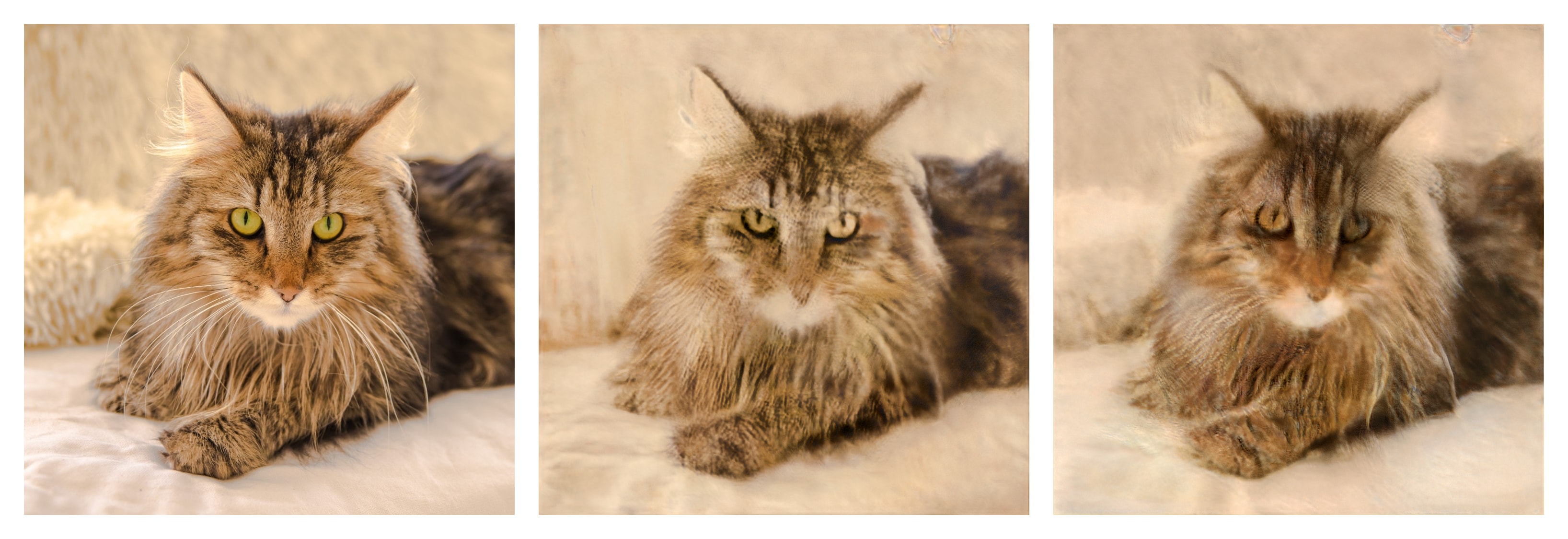}

   \includegraphics[width=\linewidth]{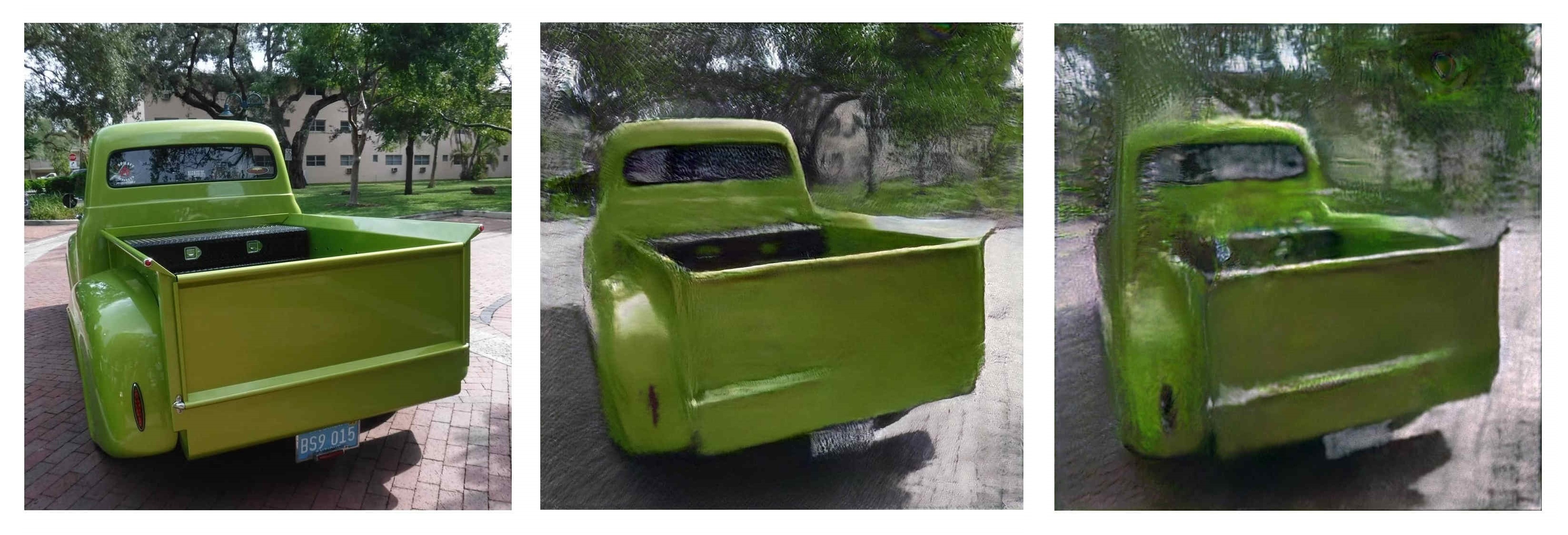}
     \includegraphics[width=\linewidth]{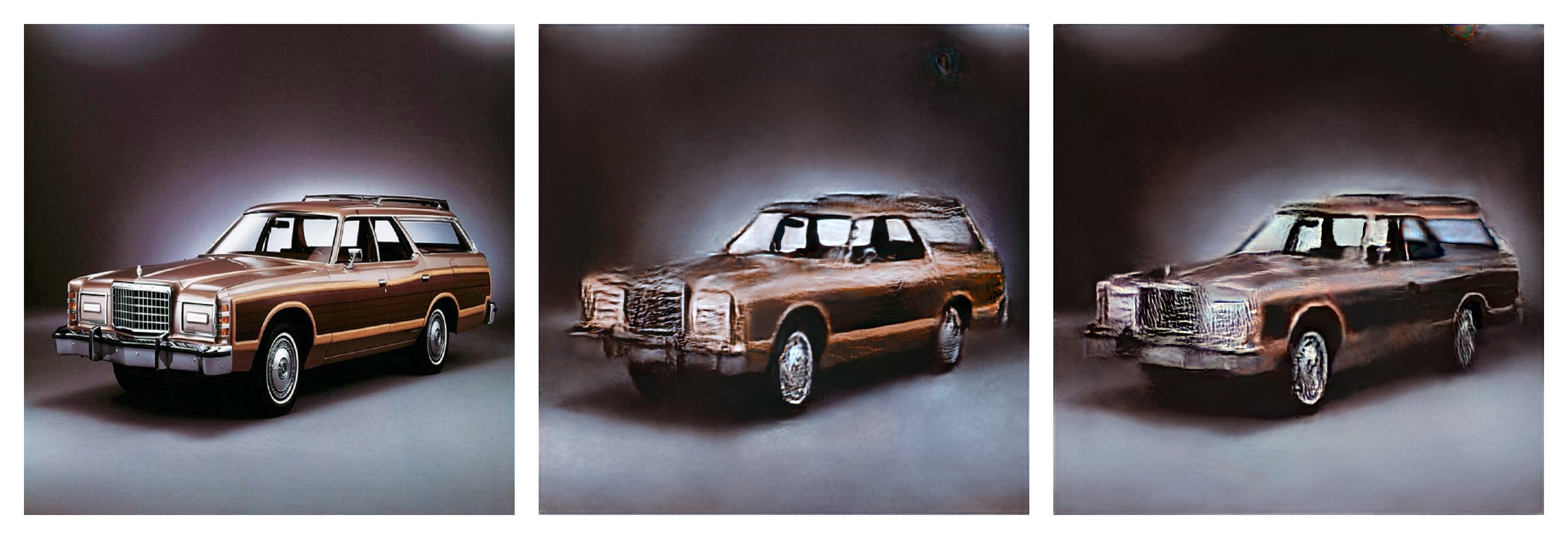}
     
\end{subfigure}
\hspace{1cm}
\begin{subfigure}{0.35\textwidth}  
\centering     
     \includegraphics[width=\linewidth]{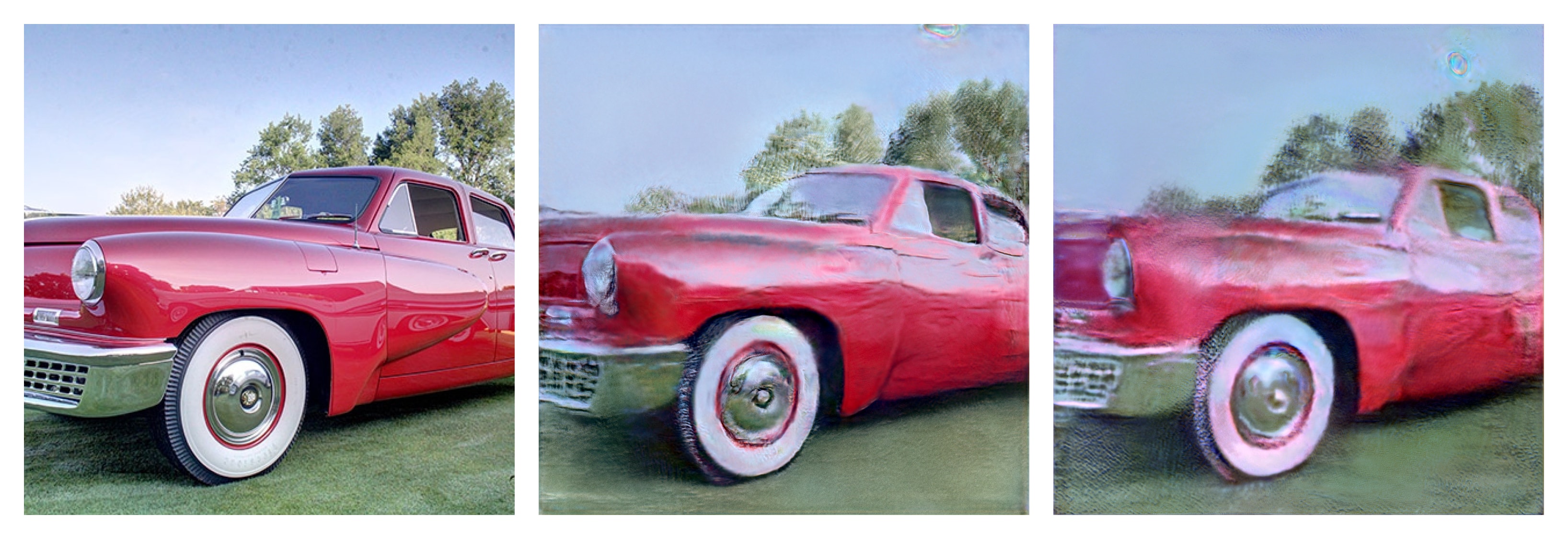}
     \includegraphics[width=\linewidth]{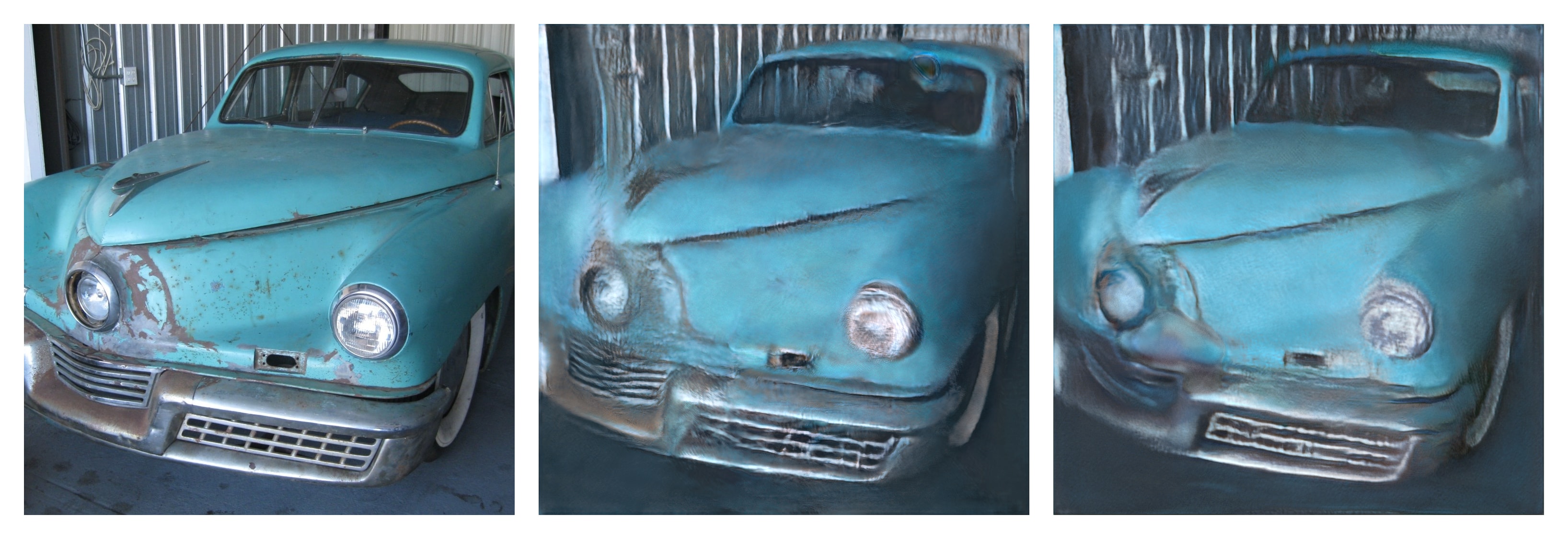}

    \includegraphics[width=\linewidth]{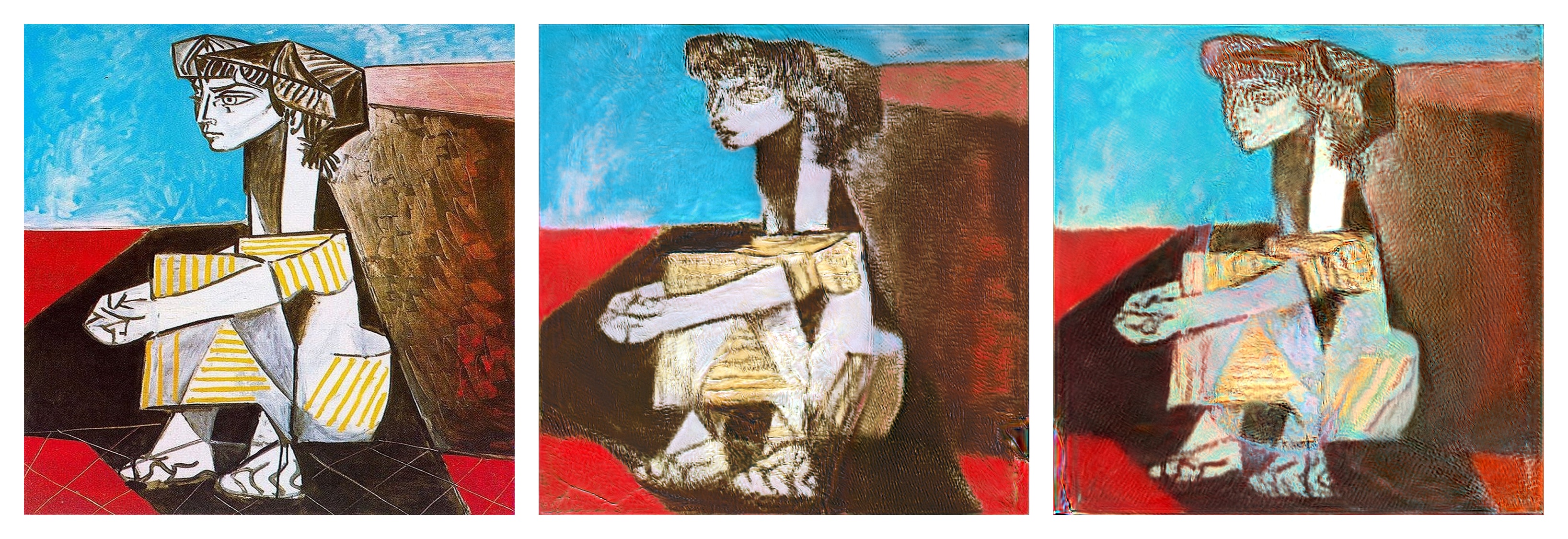}
    \includegraphics[width=\linewidth]{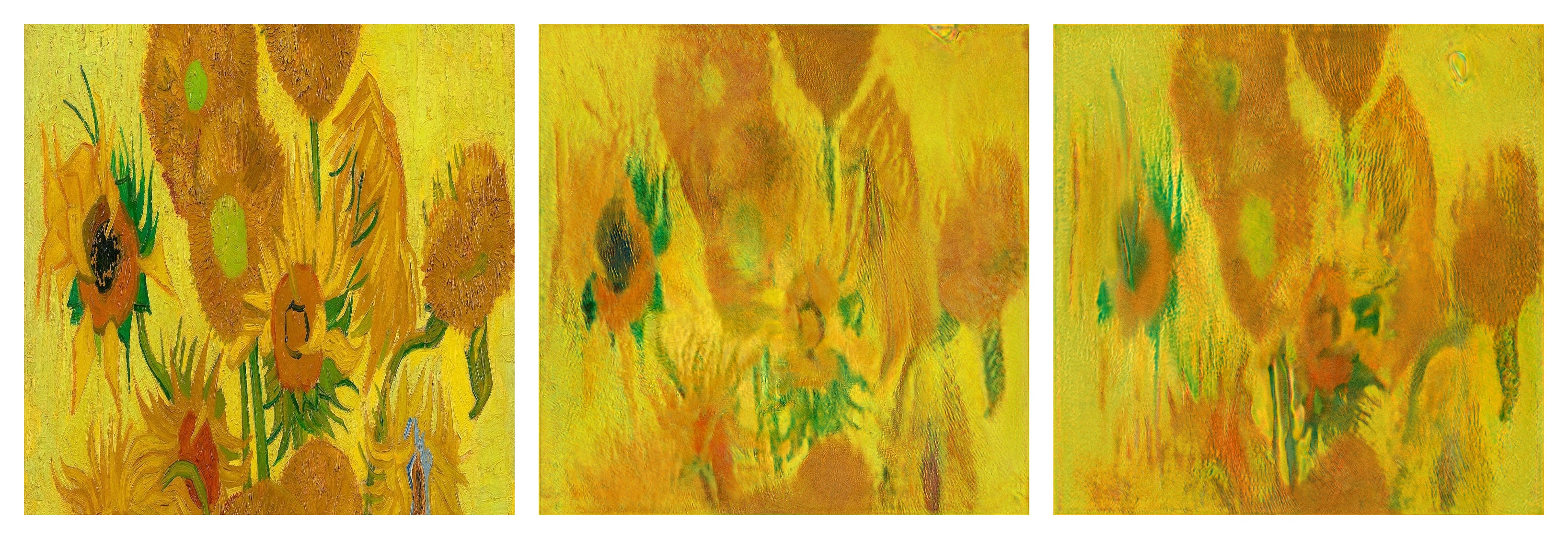}
    \includegraphics[width=\linewidth]{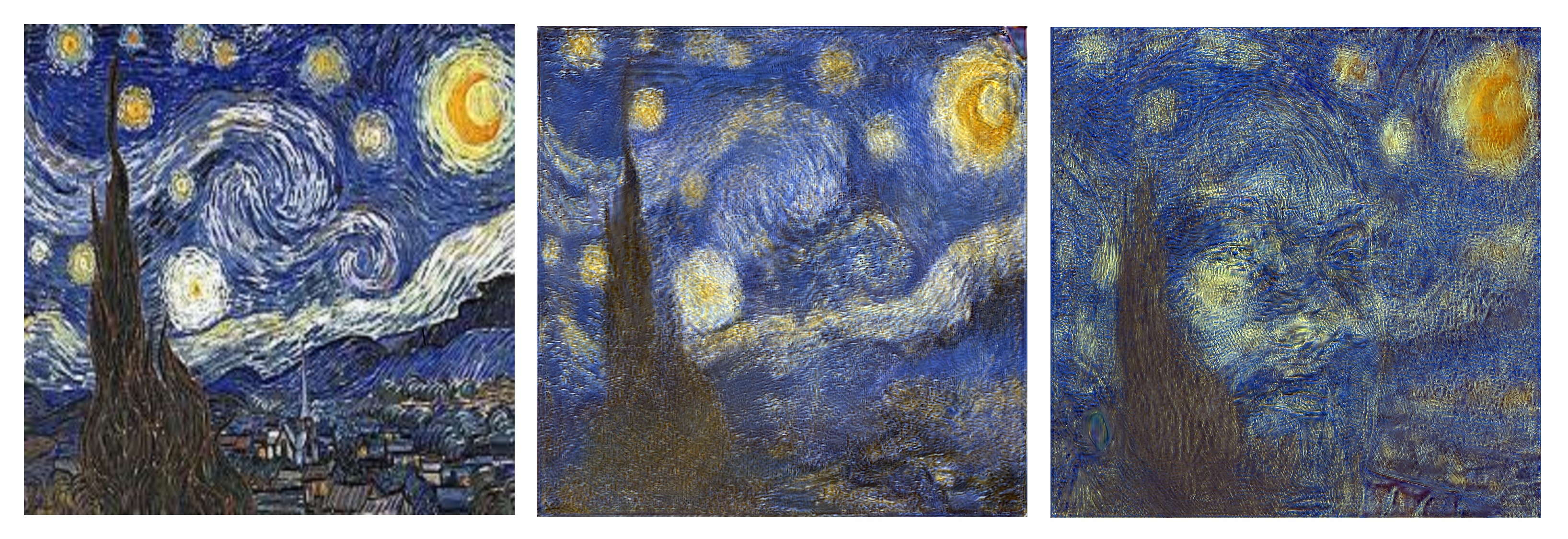}
    \includegraphics[width=\linewidth]{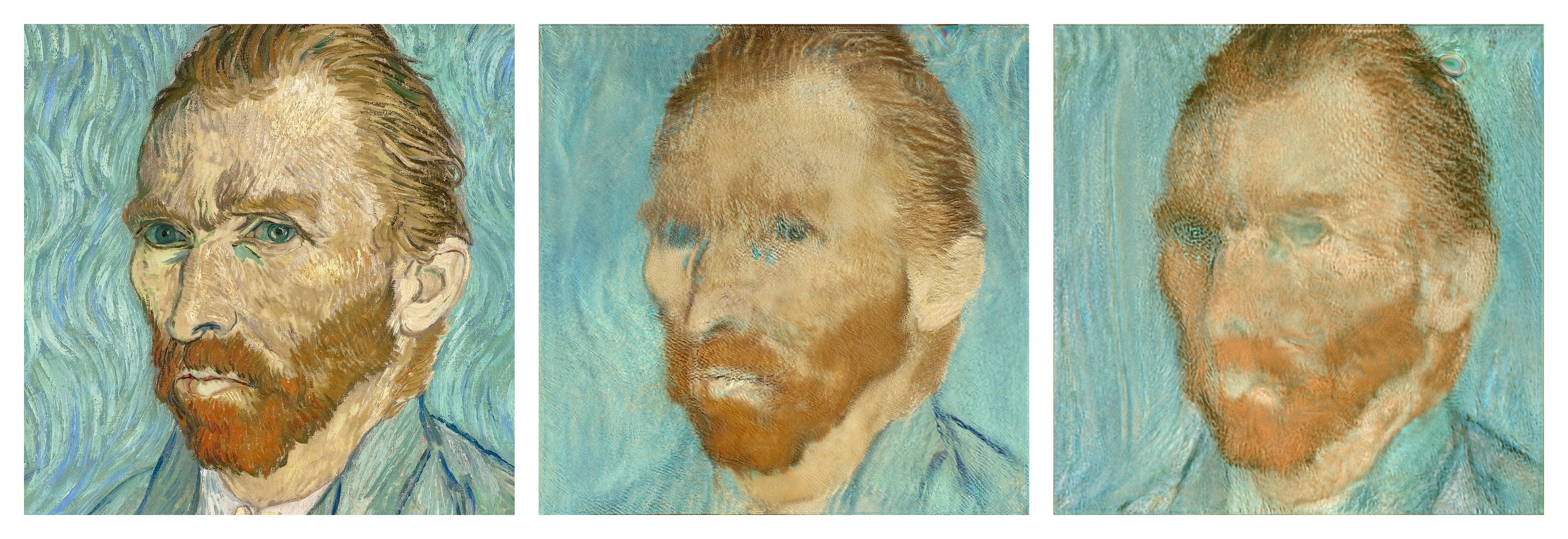}

     \includegraphics[width=\linewidth]{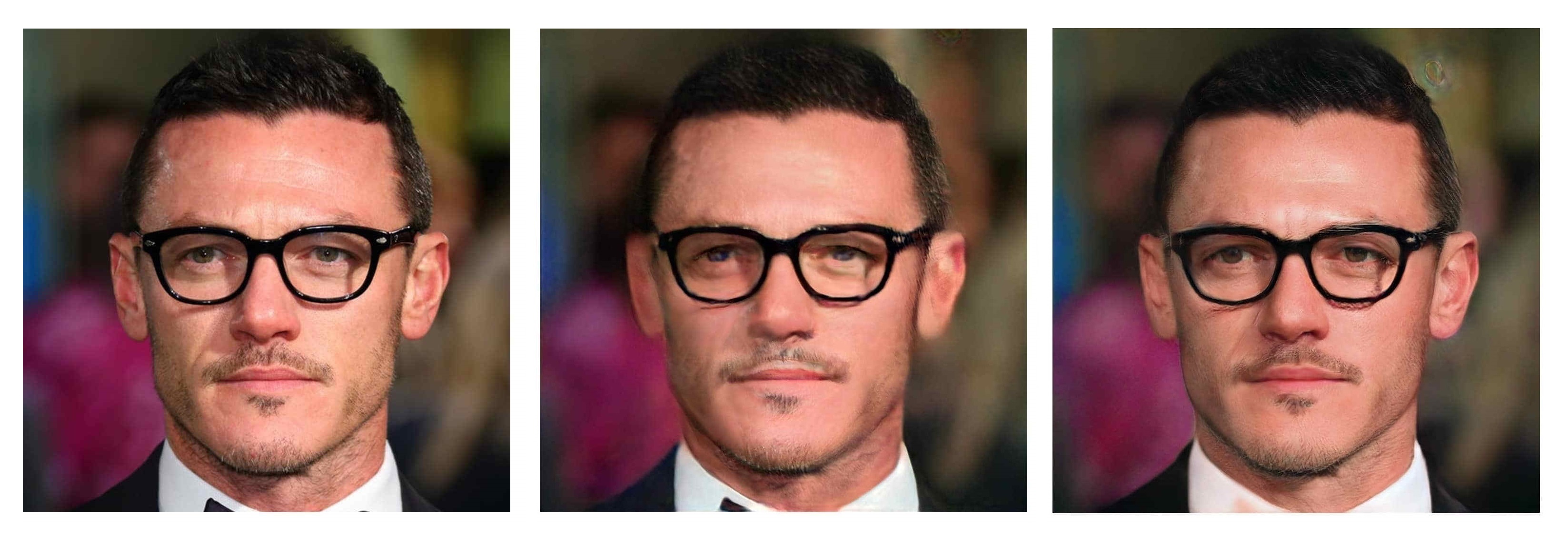}
     \includegraphics[width=\linewidth]{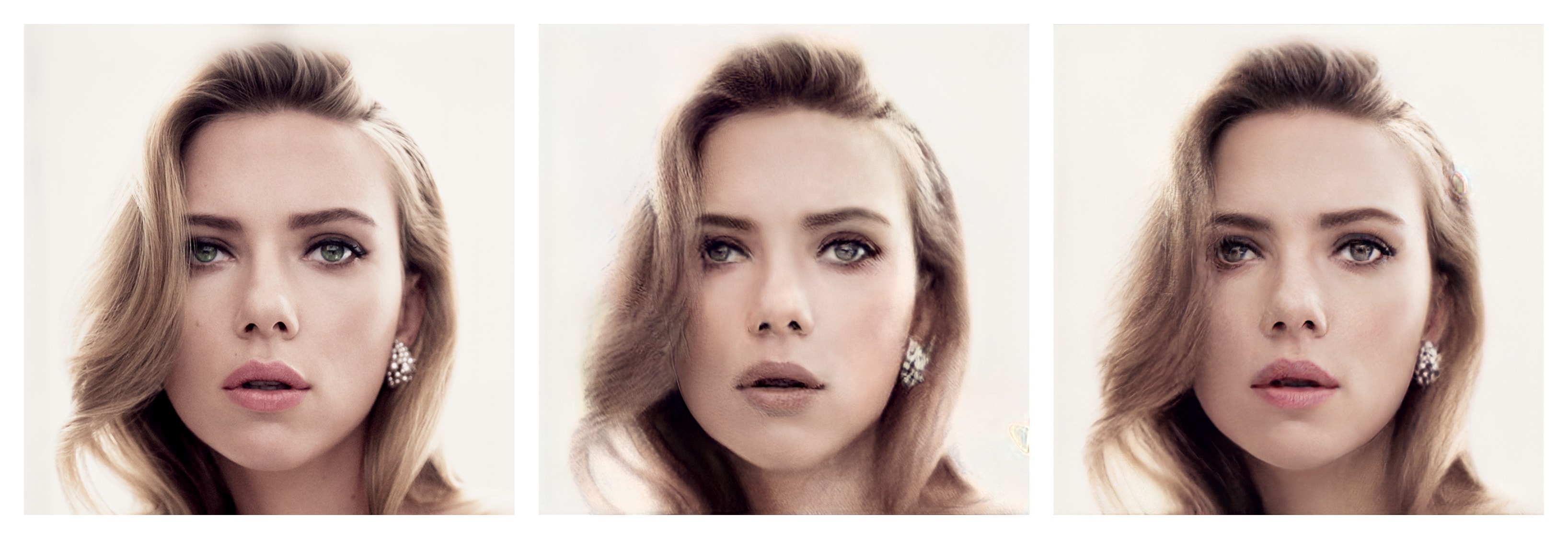}
    \includegraphics[width=\linewidth]{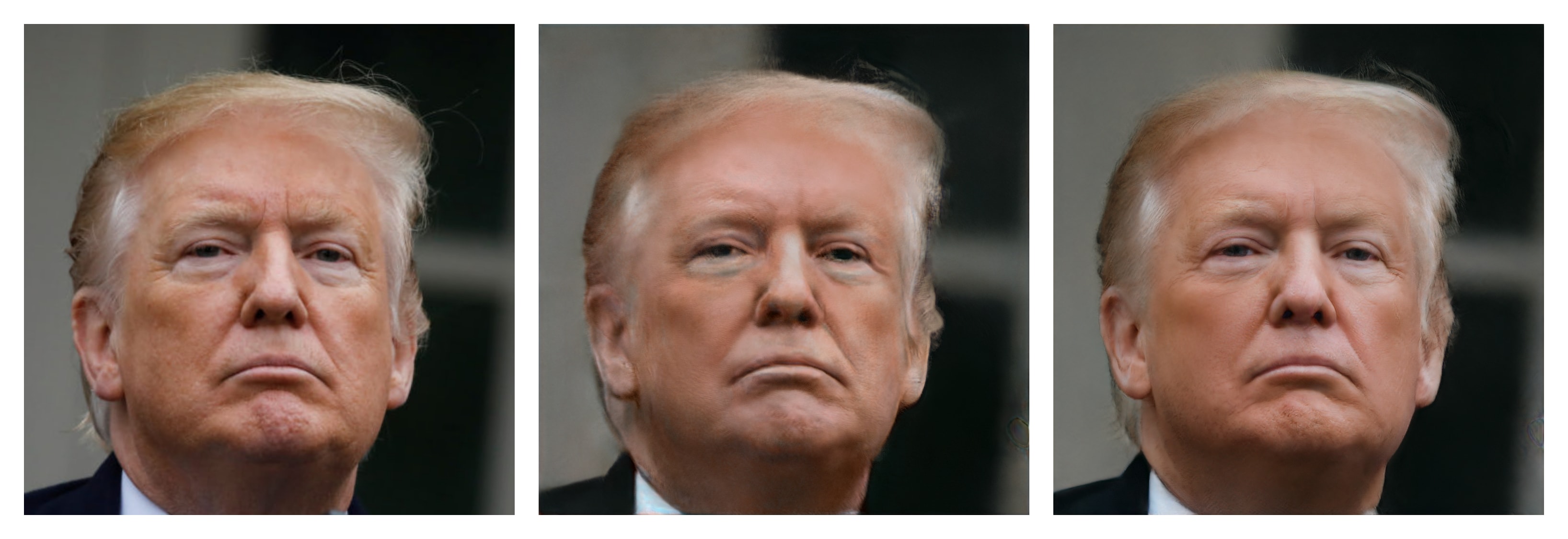}
     \includegraphics[width=\linewidth]{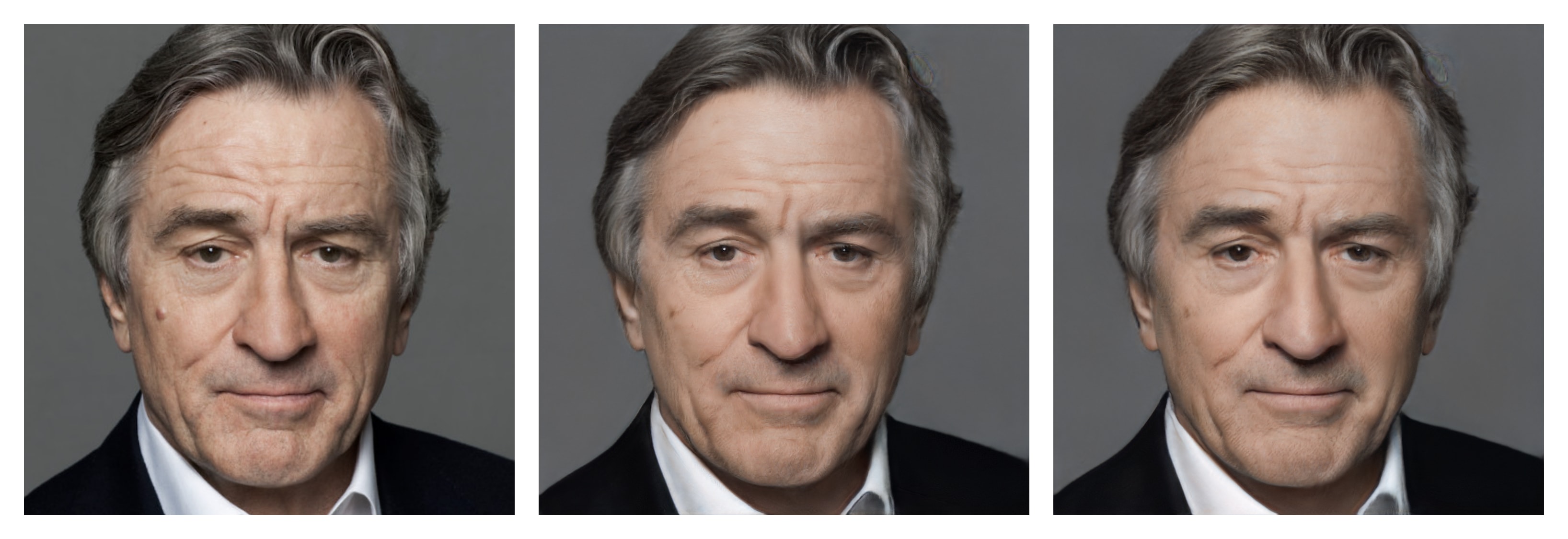}
 
\end{subfigure}
 \caption{Additional Embedding Results into $W+$ space. Left column: the original images. Middle column: the embedded images with random latent code initialization. Right column: the embedded images with $\bar{w}$ latent code initialization.} 
    \label{fig:comp22}
\end{figure*}
 \begin{table}[t]
    \centering
    \begin{tabular}{ l r r }
    \toprule
        Defect & $L$($\times 10^5$) & $\| w^* - \bar{\mathbf{w}} \|$ \\ \hline
        non-defective & 0.204 & 29.19 \\
        \hline
        Eyes  &   0.271 & 34.90 \\
        Nose &    0.311 & 39.20 \\ 
        Mouth &    0.301 & 37.04 \\
        Eyes and Mouth & 0.233 & 39.62 \\ 
        Eyes, Nose and Mouth & 0.285 & 37.59 \\

    \bottomrule
    \end{tabular}
    \caption{Quantitative results on defective image embedding (Figure 3 in the main paper). $L$ is the loss after optimization. $\| w^* - \bar{\mathbf{w}} \|$ is the distance between the latent codes $w^*$ and $\bar{\mathbf{w}}$ of the average face.}
    \label{tb:defect}
\end{table}

\begin{figure*}[t]
  
    \includegraphics[width = \linewidth]{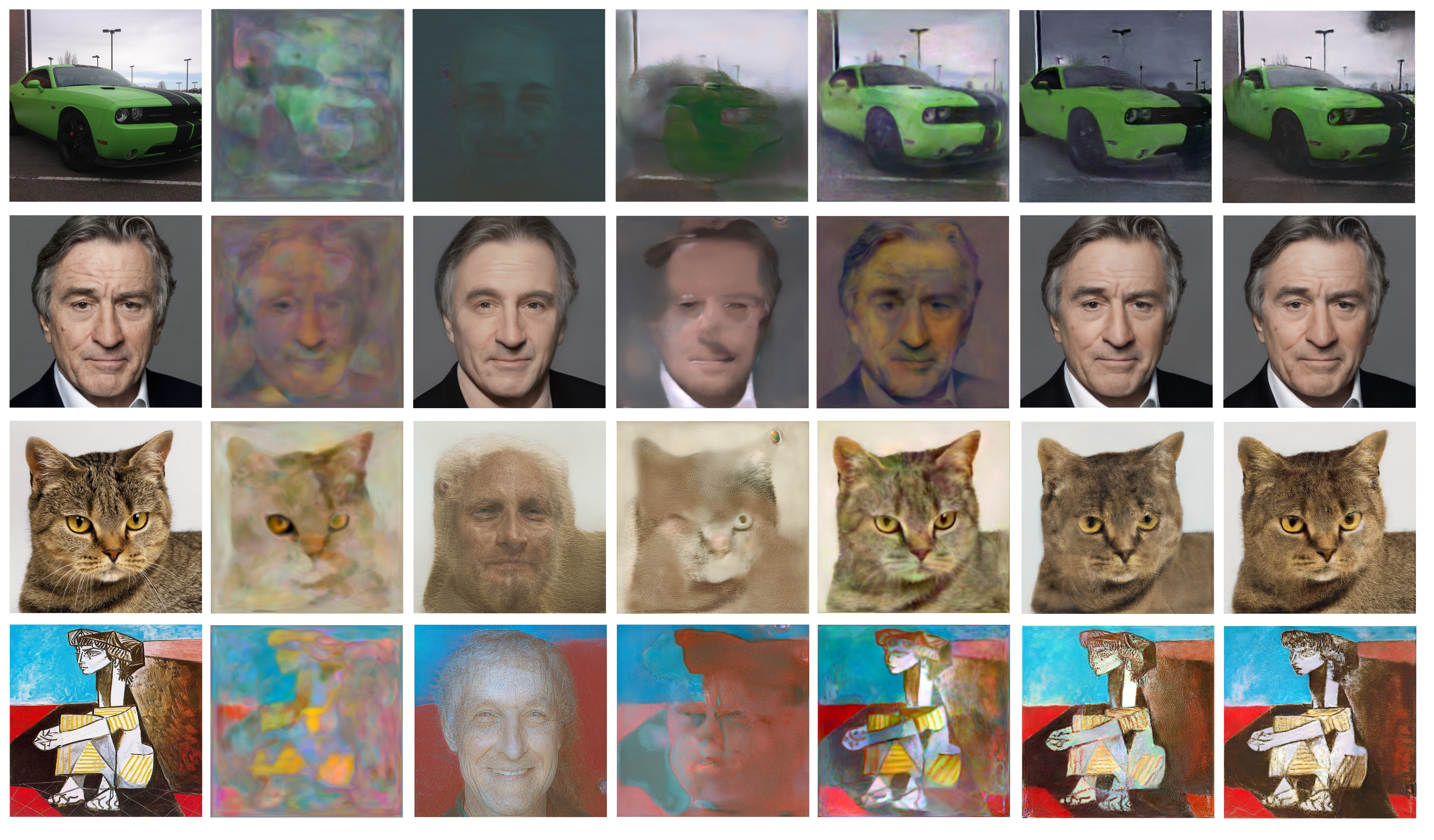}
    \hspace*{1.1cm}(a)\hspace{2.1cm}(b)\hspace{2.1cm}(c)\hspace{2.1cm}(d)\hspace{2.1cm}(e)\hspace{2.1cm}(f)\hspace{2.1cm}(g)
    \caption{Additional results on the justification of latent space
choice.(a) Original images. Embedding results into the original space $W$: (b) using random weights in the network layers; (c) with $\bar{\mathbf{w}}$ initialization; (d) with random initialization. Embedding results into the $W+$ space: (e) using random weights in the network layers; (f) with $\bar{\mathbf{w}}$ initialization; (g) with random initialization.} 
    \label{fig:random}
\end{figure*}

\paragraph{StyleGANs trained on Other Datasets}
To support our insights on the learned distribution, we further tested our embedding algorithm on the StyleGANs trained on three more datasets: the LSUN-Car ($512 \times 384$), LSUN-Cat ($256 \times 256$) and LSUN-Bedroom ($256 \times 256$) datasets.
The embedding results are shown in Figure \ref{fig:lsun}. 
It can be observed that the quality of the embedding is poor compared to that of the StyleGAN trained on the FFHQ dataset.
The linear interpolation (image morphing) results of LSUN-Cat, LSUN-Car, and LSUN-Bedroom StyleGANs are shown in Figure \ref{fig:morph_lsun} (a), (b) and (c) respectively.
Interestingly, we observed that linear interpolation fails on the LSUN-Cat and LSUN-Car StyleGANs. 
Recall that the FFHQ human face dataset is of very high quality in terms of scale, alignment, color, poses \etc, we believe that the low quality of the LSUN datasets is the source of such failure.
In other words, the quality of the data distribution is one of the key components to learn a meaningful model distribution.

\begin{figure*}

     \centering
    \includegraphics[width =0.9\linewidth]{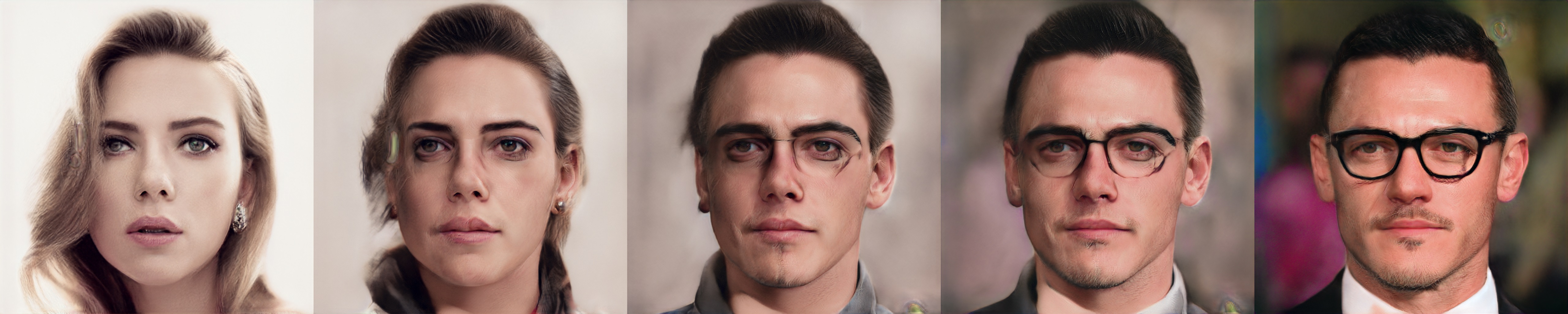}
     \includegraphics[width = 0.9\linewidth]{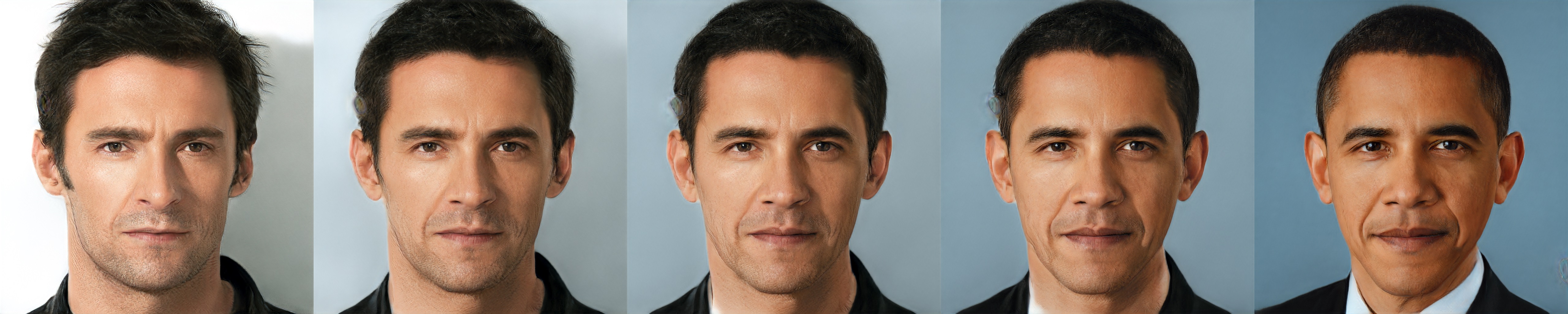}

    \includegraphics[width = 0.9\linewidth]{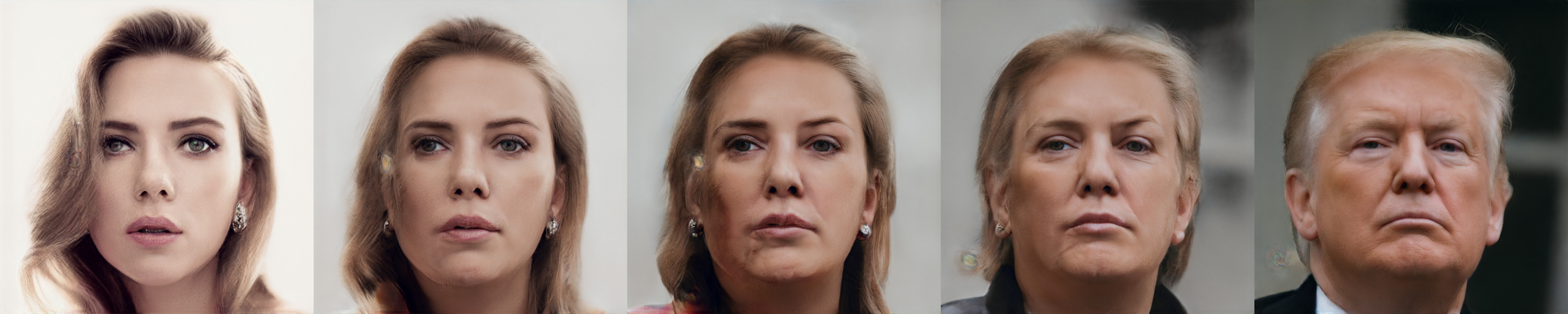}
    
    \includegraphics[width = 0.9\linewidth]{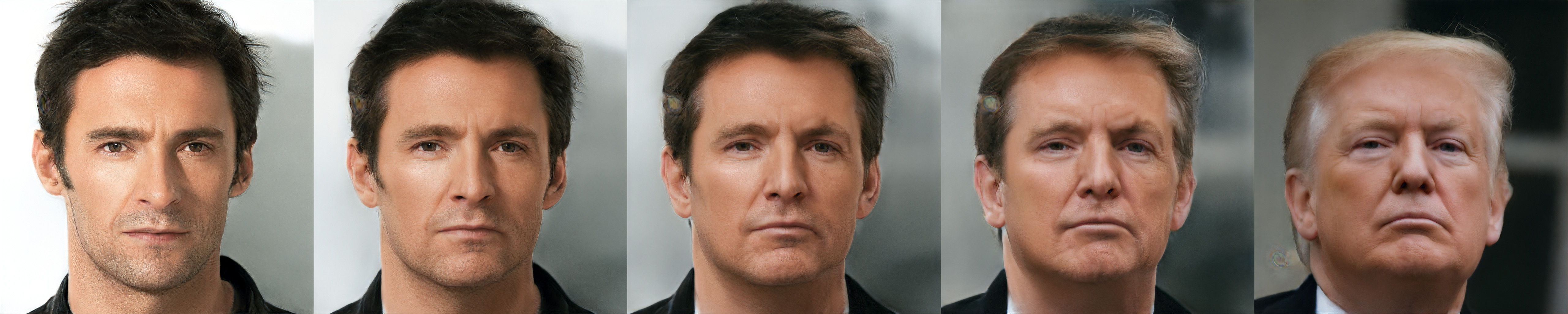}
    
    \includegraphics[width = 0.9\linewidth]{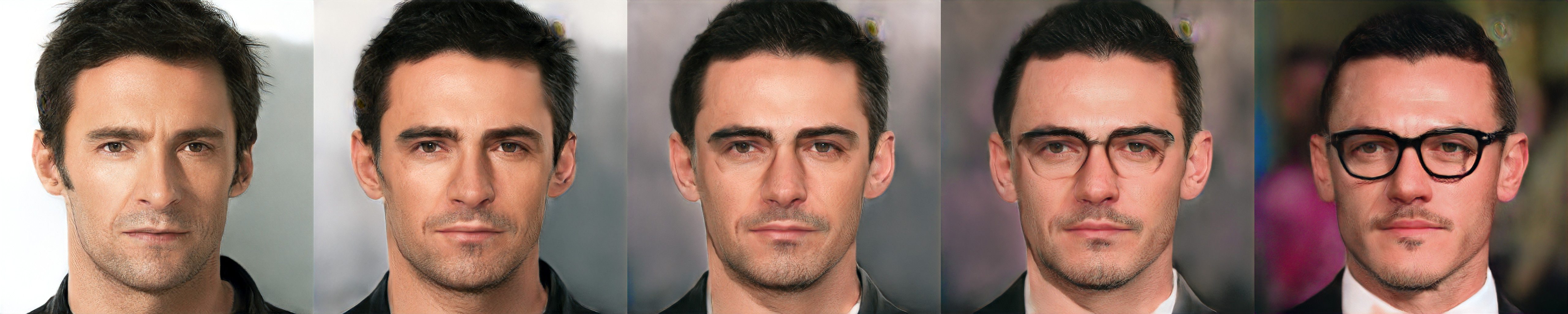}
    \includegraphics[width = 0.9\linewidth]{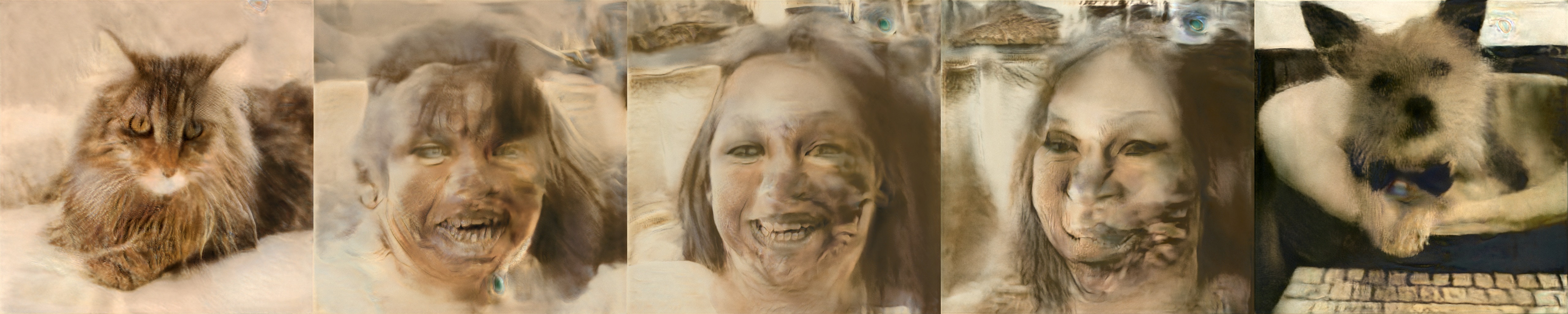}
    
    \includegraphics[width = 0.9\linewidth]{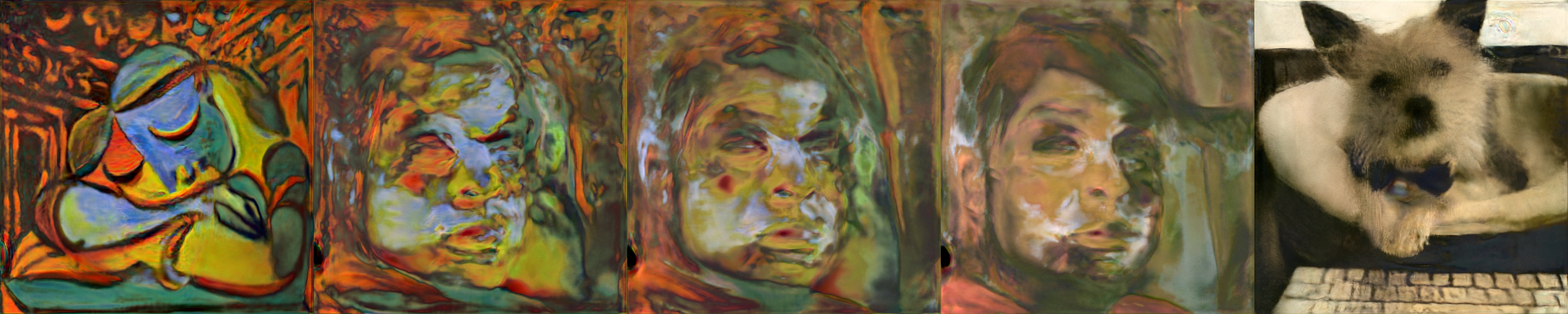}

    \caption{Additional morphing results between two embedded images (the
left-most and right-most ones).}
    \label{fig:mor}
\end{figure*}

\begin{figure}[t]
\centering

\includegraphics[width=\linewidth]{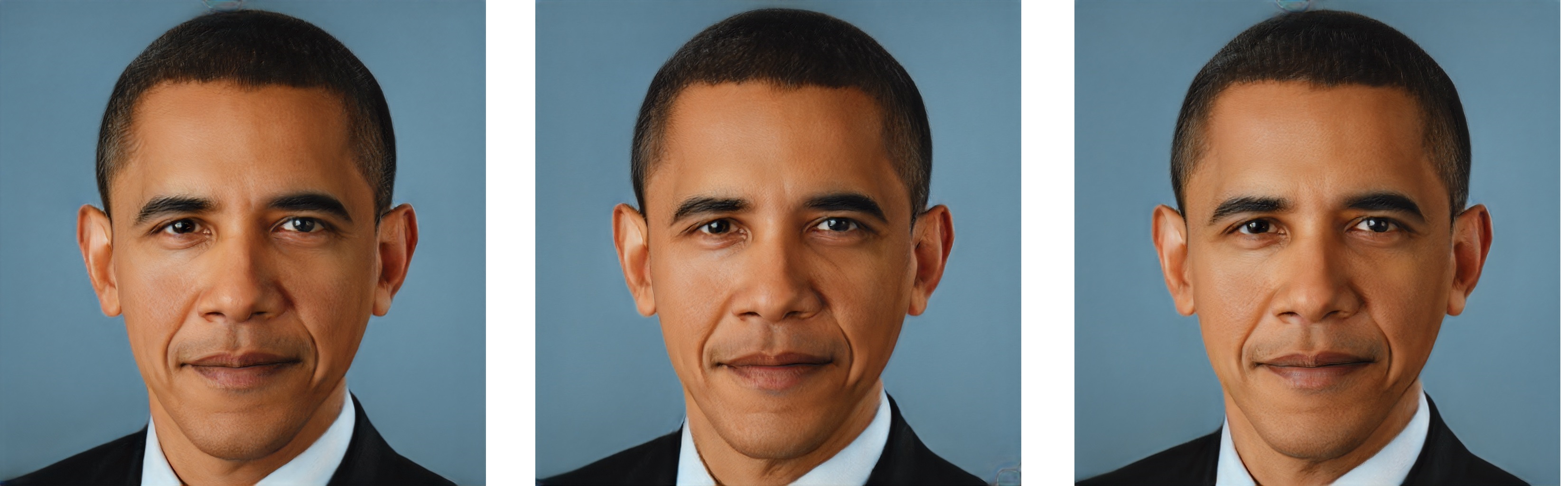}
 \caption{Image embedding using different constant noises.}
    \label{fig:emb}
\end{figure}
\paragraph{Additional Results on the Justification of Latent Space Choice}
Figure \ref{fig:random} shows additional results (cat, dog, car) on the justification of our choice of latent space $W^+$.
Similar to the main paper, we can observe that: (i) embedding into $W$ directly does not give reasonable results; (ii) the learned network weights is important to good embeddings.

\begin{figure*}[t]
  
    \includegraphics[width=\linewidth]{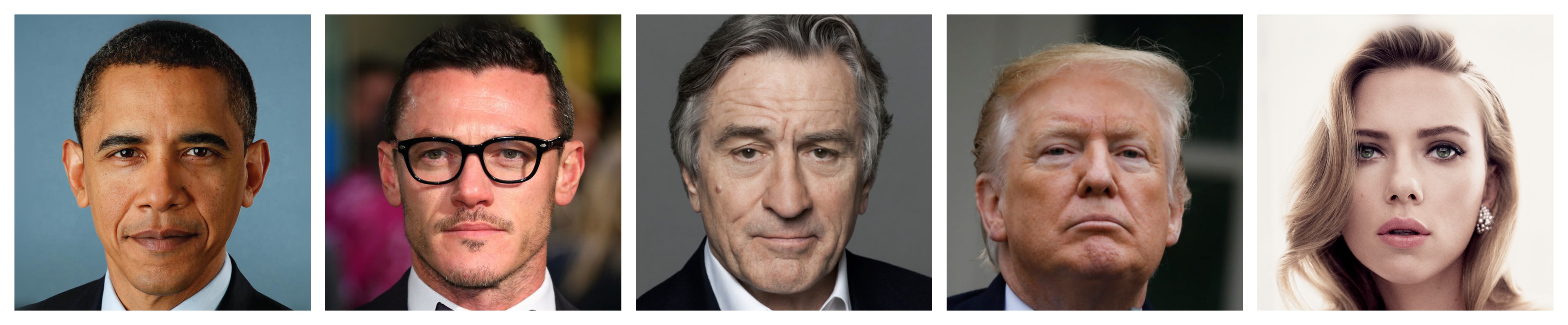}
    \includegraphics[width=\linewidth]{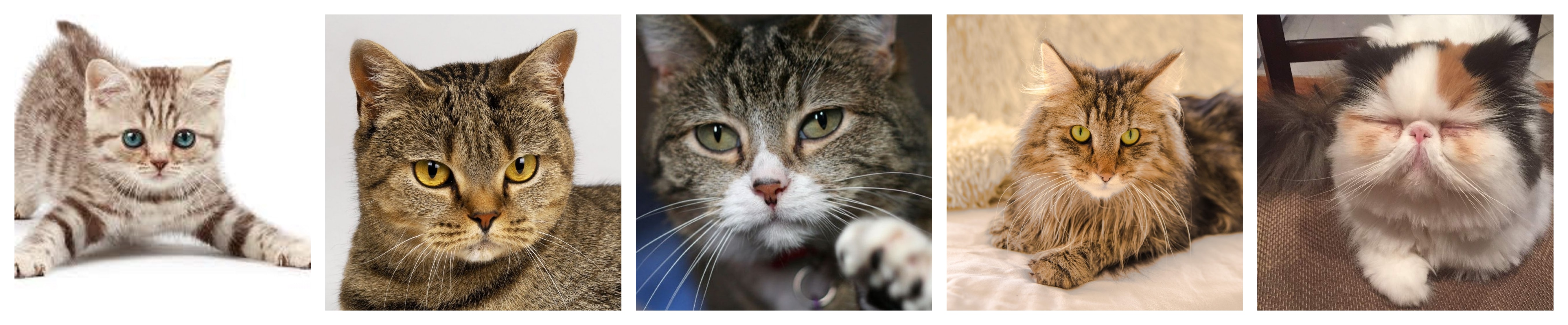}
    \includegraphics[width=\linewidth]{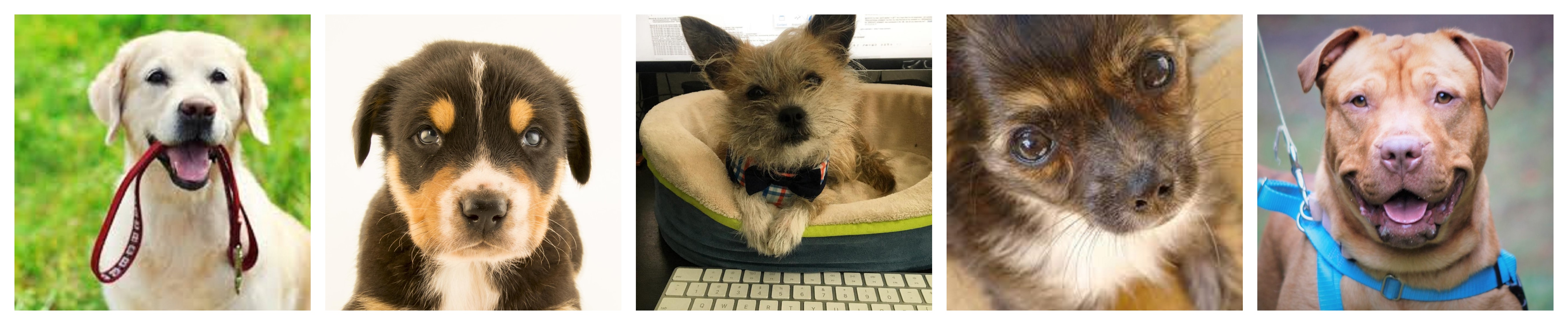}
     \includegraphics[width=\linewidth]{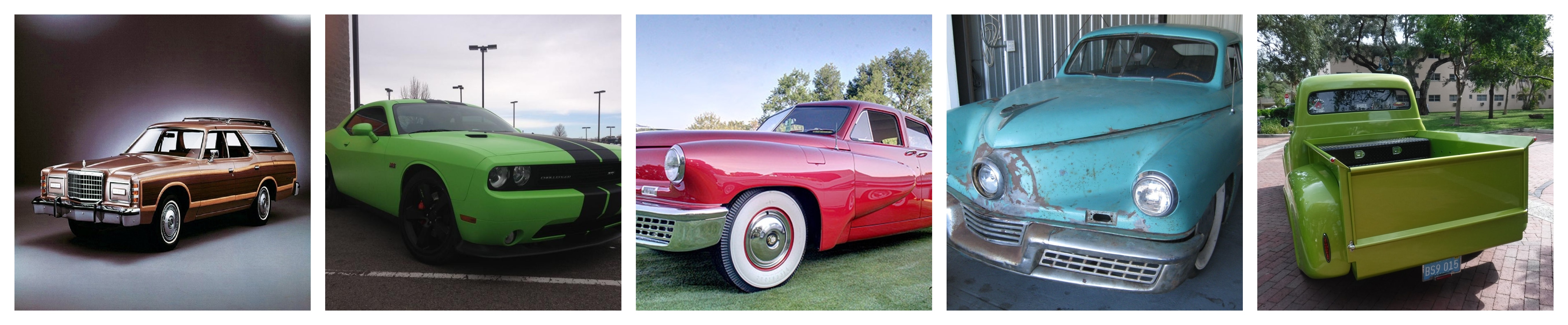}
    \includegraphics[width=\linewidth]{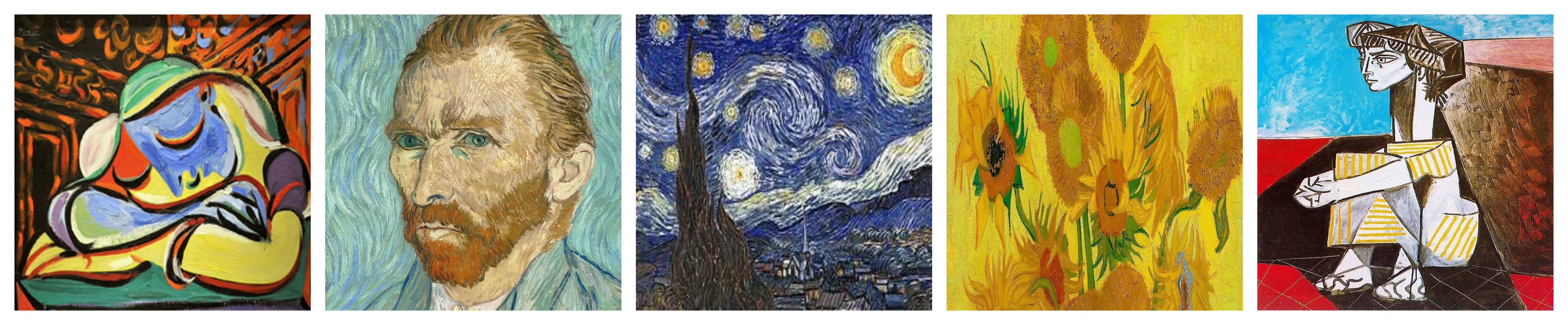}
   
    \caption{The collected 25 images of our dataset. First row: human faces. Second row: cats. Third row: dogs. Fourth row: cars. Fifth row: paintings.   }
    \label{fig:whole}
\end{figure*}

\begin{figure}
   
    \begin{subfigure}{\linewidth}
     \centering
    \includegraphics[width =0.99\textwidth]{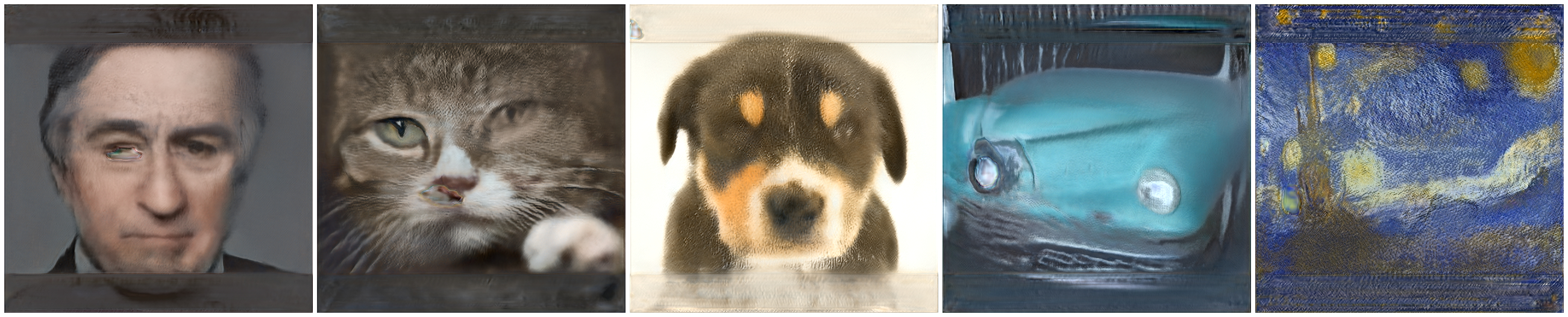}
     \includegraphics[width = 0.99\textwidth]{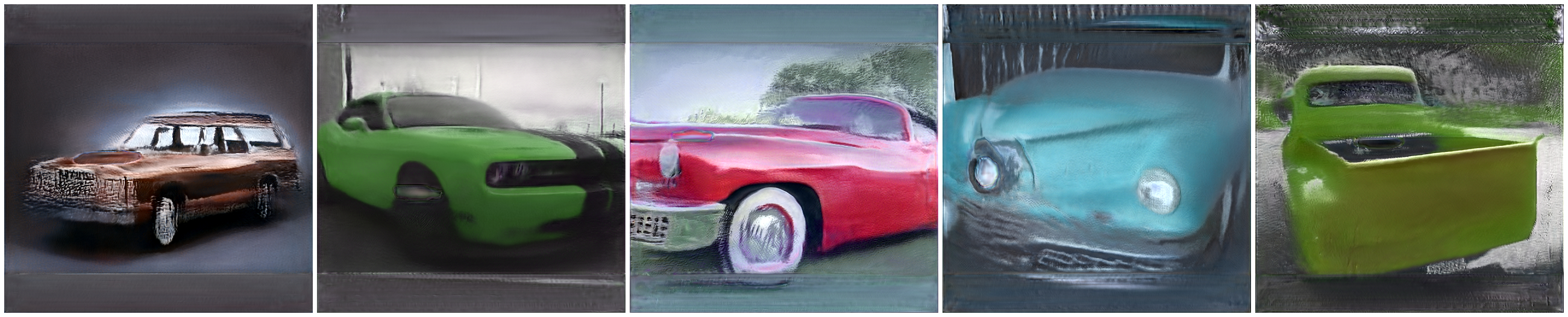}
     \subcaption{}
    \end{subfigure}
      \begin{subfigure}{\linewidth}
     \centering
    \includegraphics[width = 0.99\textwidth]{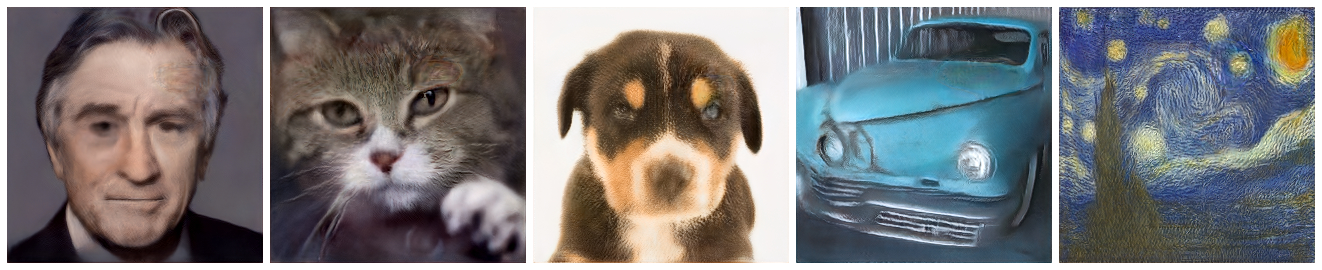}
     \includegraphics[width = 0.99\textwidth]{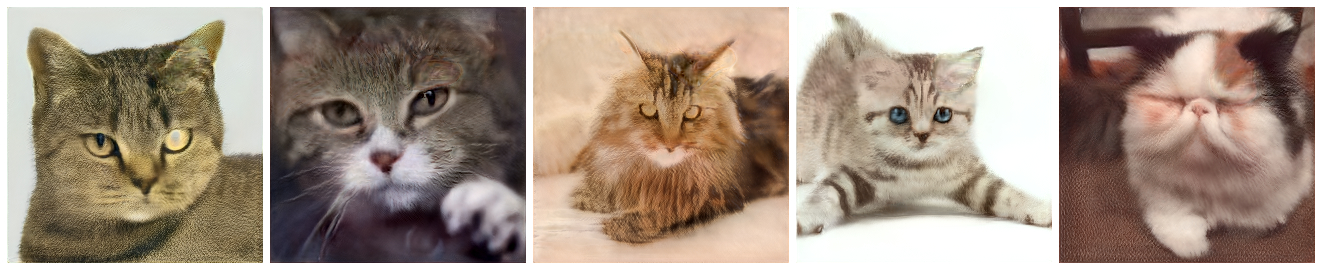}
     \subcaption{}
    \end{subfigure}
      \begin{subfigure}{\linewidth}
     \centering
    \includegraphics[width = 0.99\linewidth]{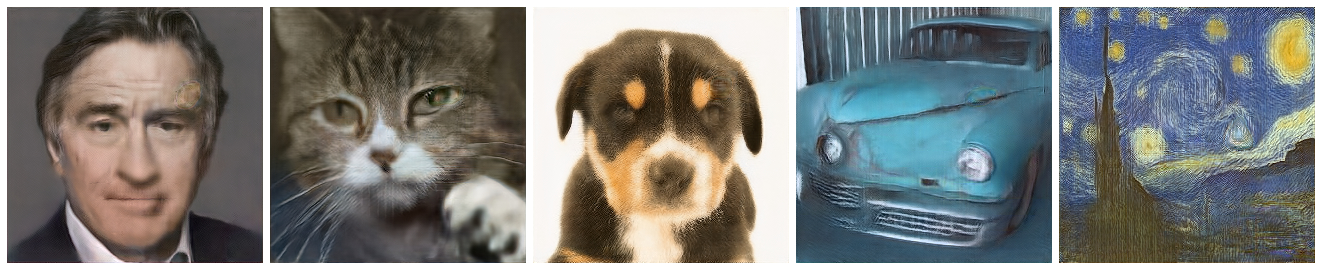}     
    \subcaption{}
    \end{subfigure}
    \caption{Embedding results of StyleGANs trained on (a) LSUN-Car, (b) LSUN-Cat and (c) LSUN-Bedroom datasets. For each subfigure, the first row shows the embedding results of the images in 5 different classes in our dataset. The second row shows the embedding results of the images of the corresponding class in our dataset (``cars'' in (a) and ``cats'' in (b)). Note that (c) has only one row because we did not collect bedroom images in our dataset.}
    
    \label{fig:lsun}
\end{figure}

\begin{figure}[t]
   
    \begin{subfigure}{\linewidth}
     \centering
    \includegraphics[width =0.99\textwidth]{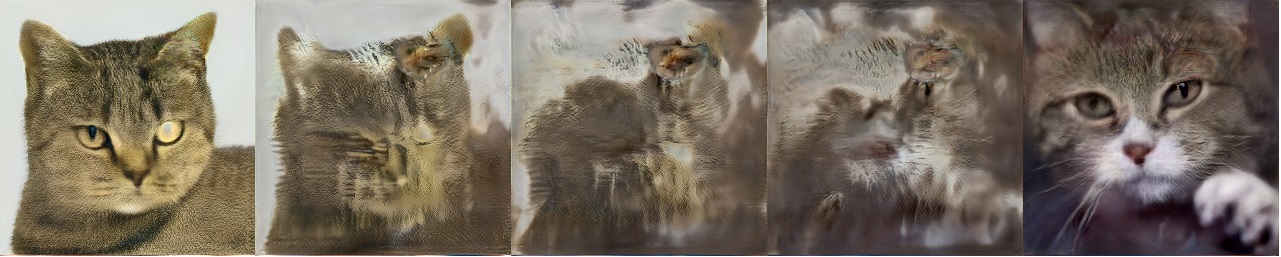}
     \includegraphics[width = 0.99\textwidth]{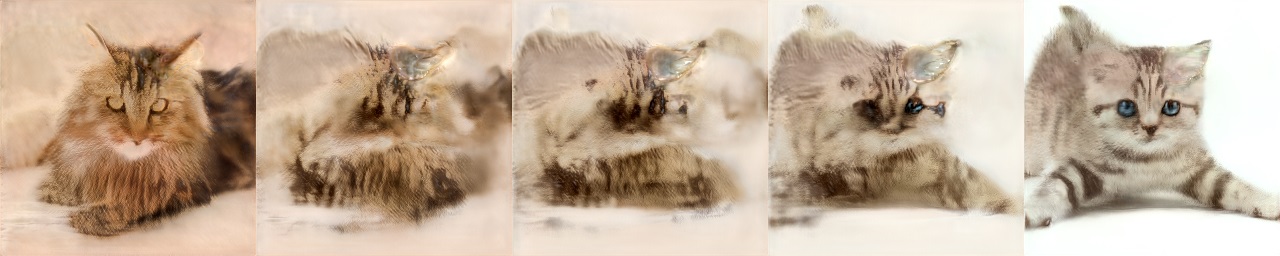}
     \subcaption{}
    \end{subfigure}
      \begin{subfigure}{\linewidth}
     \centering
    \includegraphics[width = 0.99\textwidth]{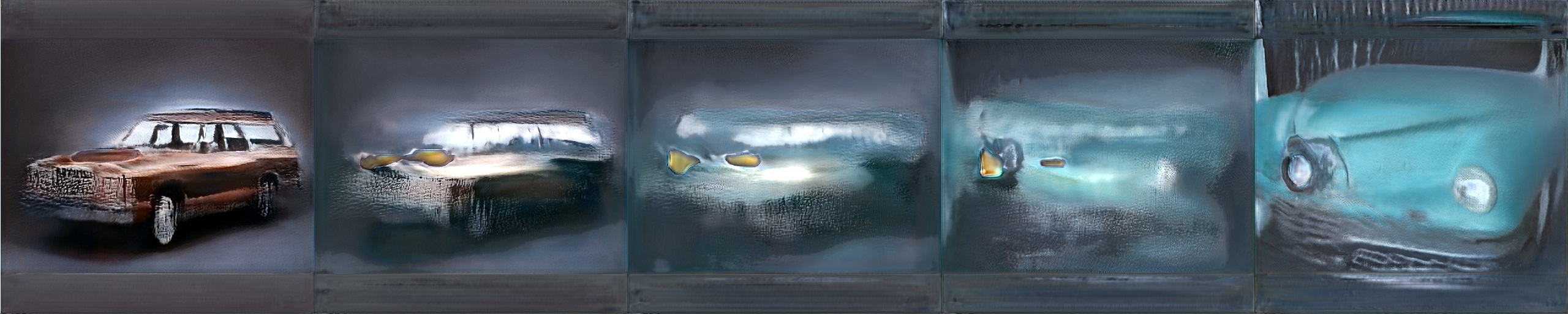}
     \includegraphics[width = 0.99\textwidth]{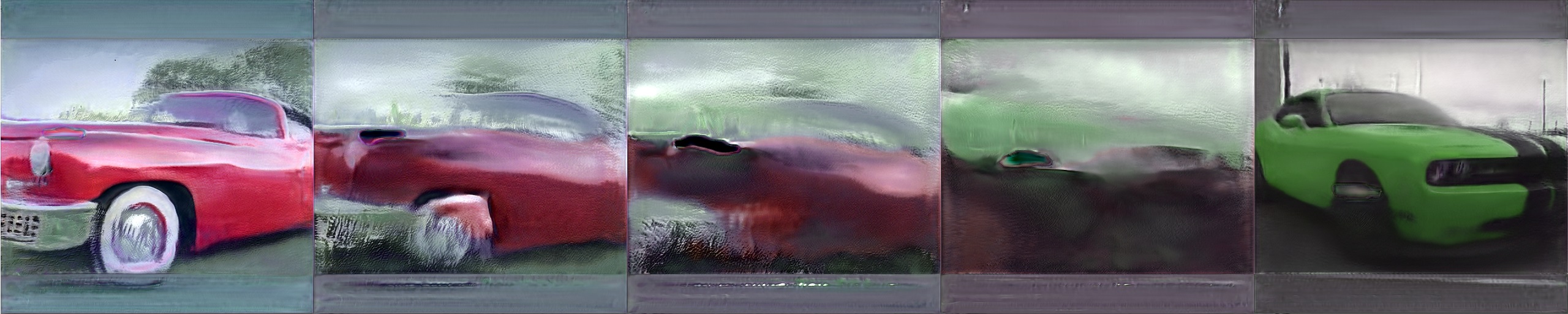}
     \subcaption{}
        \centering
    \includegraphics[width = 0.99\textwidth]{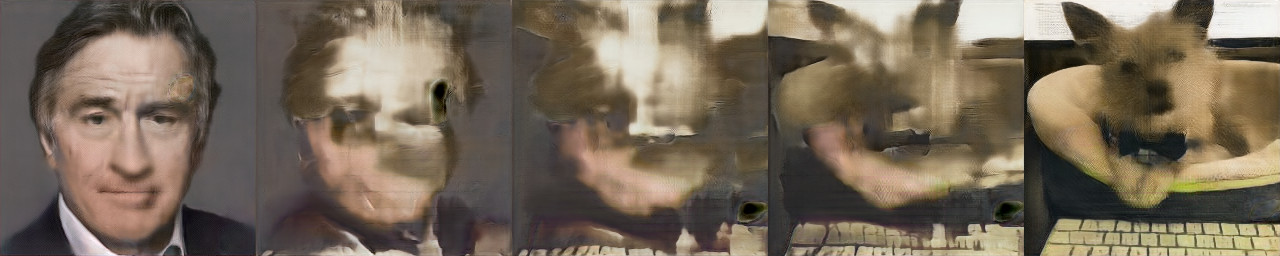}
     \includegraphics[width = 0.99\textwidth]{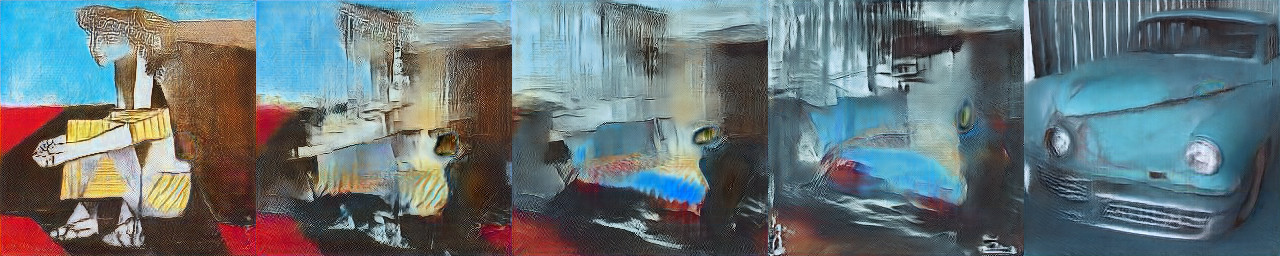}
     \subcaption{}
    \end{subfigure}
      
    \caption{Results on linear interpolations (image morphing) in the latent spaces of StyleGANs trained on (a) LSUN-Cat (b) LSUN-Car (c) LSUN-Bedroom datasets.}
    \label{fig:morph_lsun}
\end{figure}

\begin{figure*}[b]
    \centering
    \includegraphics[width=\linewidth]{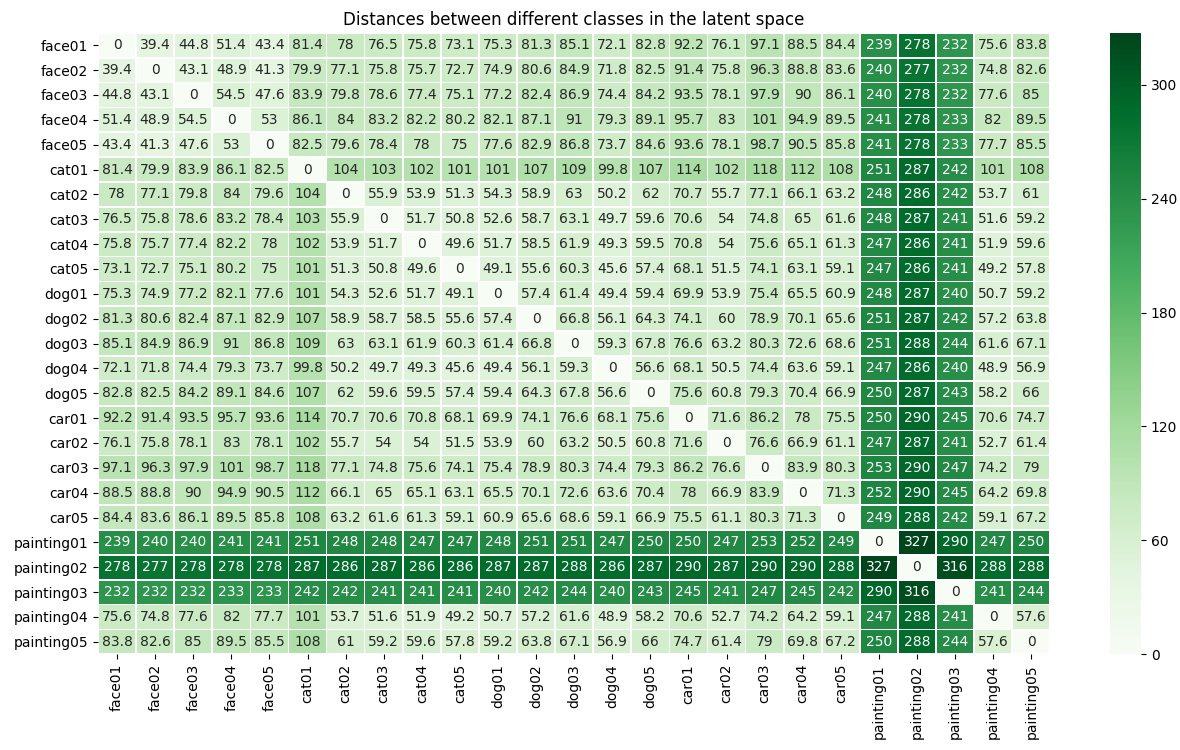}
    \caption{Heat map of the inter- and intra-class $L2$ distances between embedded images.}
    \label{fig:heat2}
\end{figure*}

\begin{figure*}[t]
    \centering
    \includegraphics[width=0.75\linewidth]{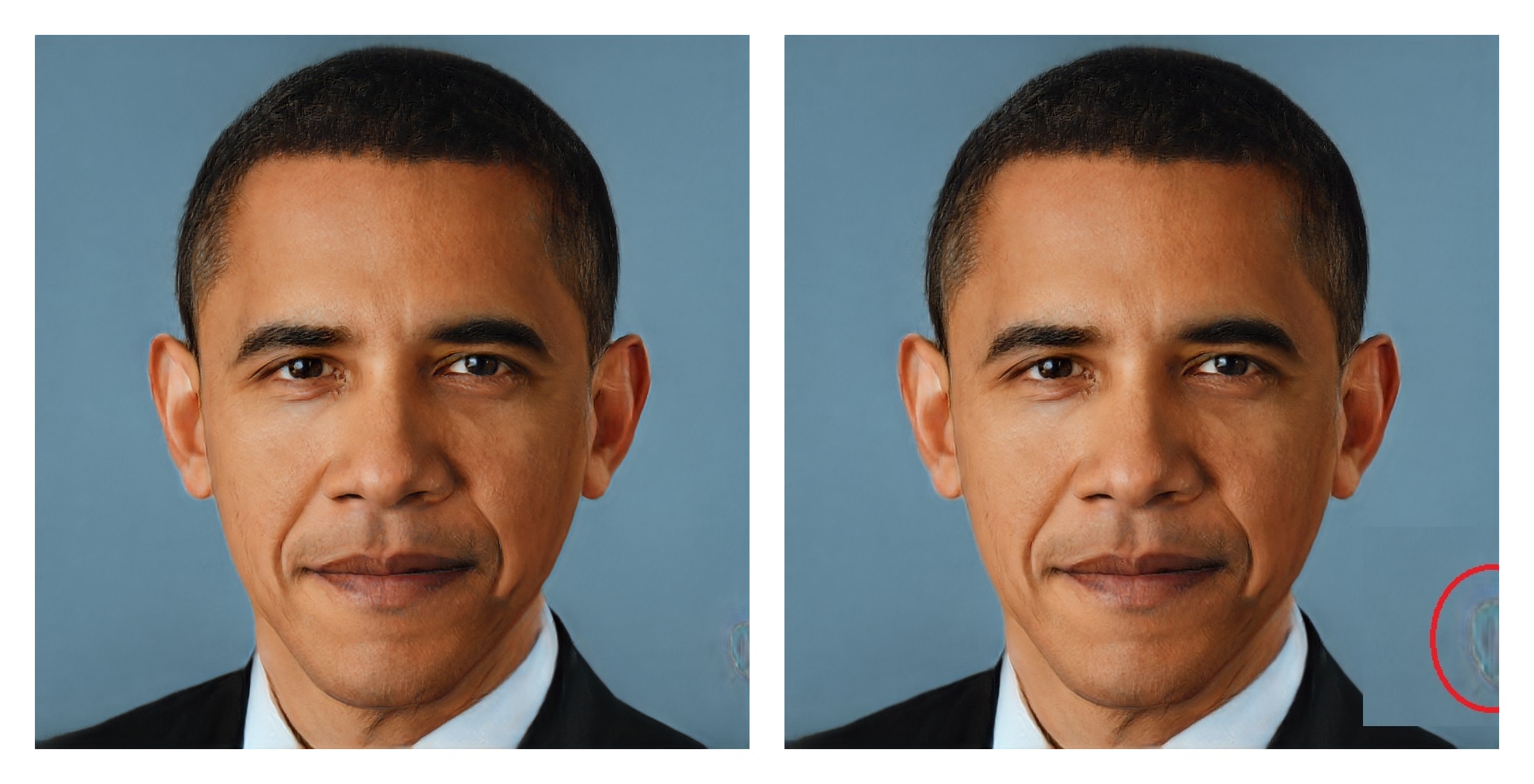}
      \includegraphics[width=0.75\linewidth]{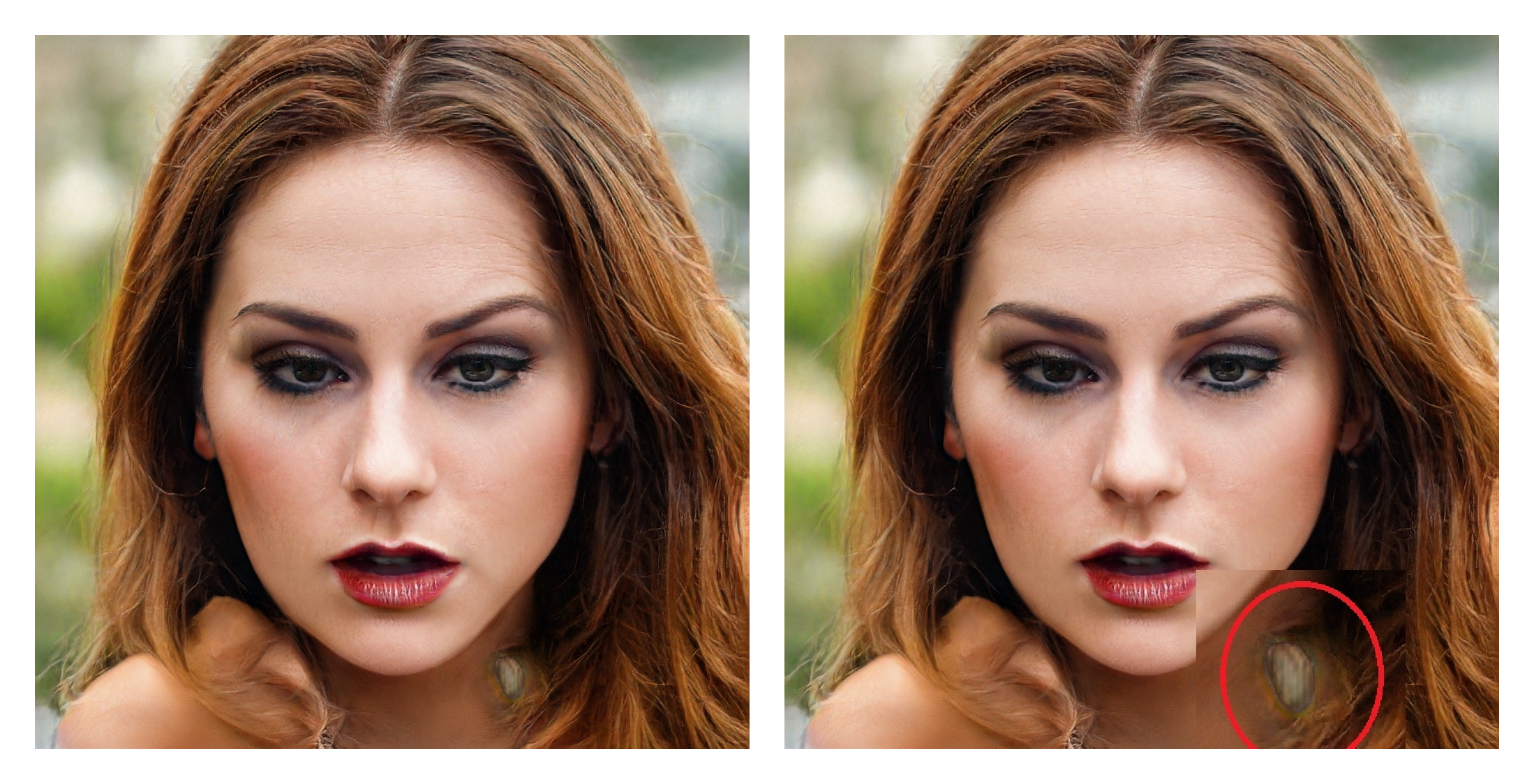}
       \includegraphics[width=0.75\linewidth]{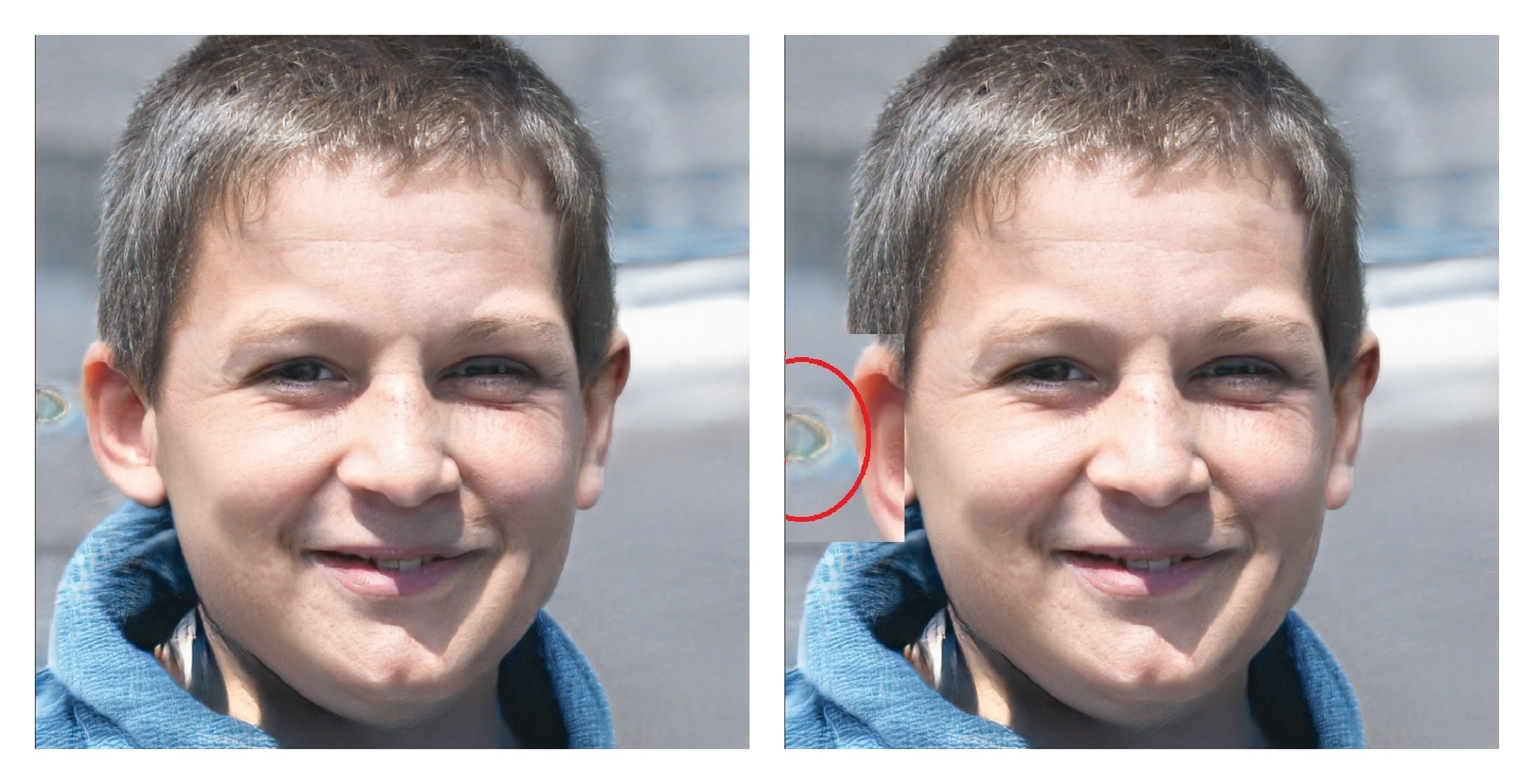}
    \caption{Inherent circular artifacts of StyleGAN. First row: circular artifacts in the embeded images. Second and third rows: randomly generated images. Left column: images with circular artifacts. Right column: highlighted artifacts by zooming in their local neighbourhood.}
    \label{fig:eye}
\end{figure*}

\begin{figure*}[t]
\centering
\includegraphics[width=0.99\textwidth]{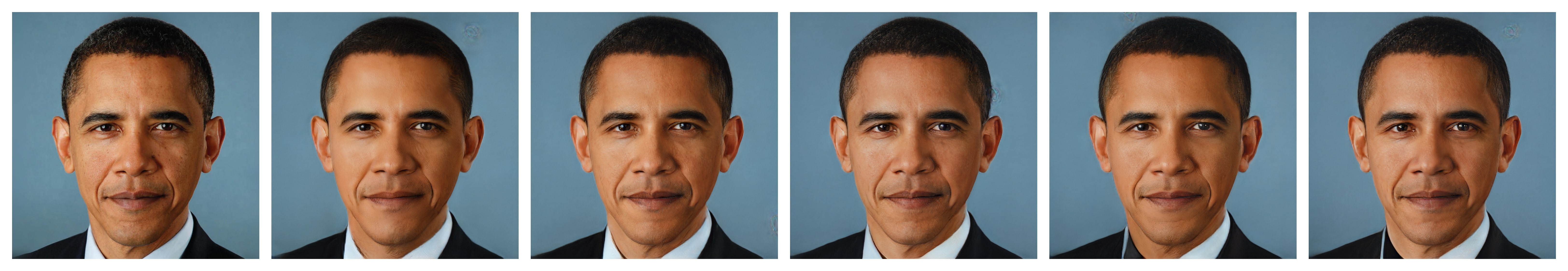}
\includegraphics[width=0.99\textwidth]{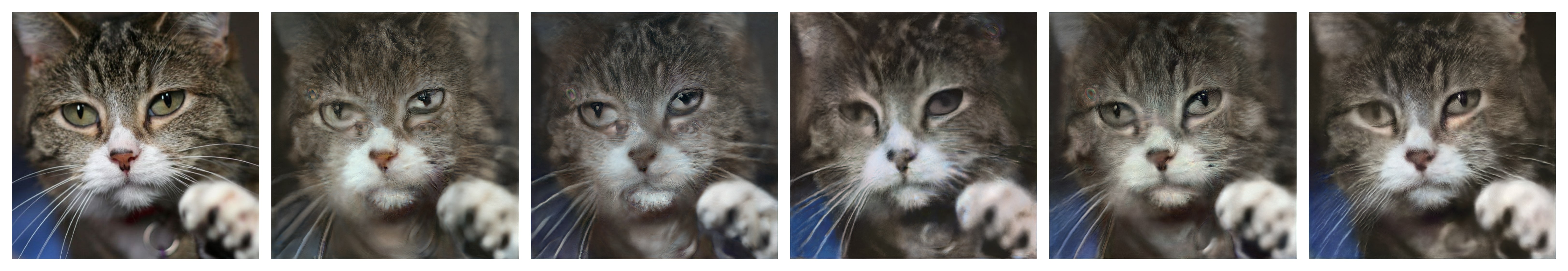}
\includegraphics[width=0.99\textwidth]{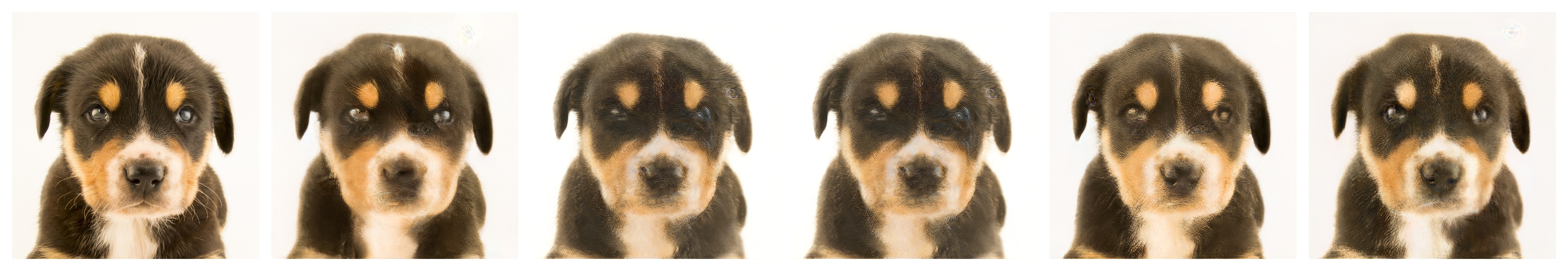}
\includegraphics[width=0.99\textwidth]{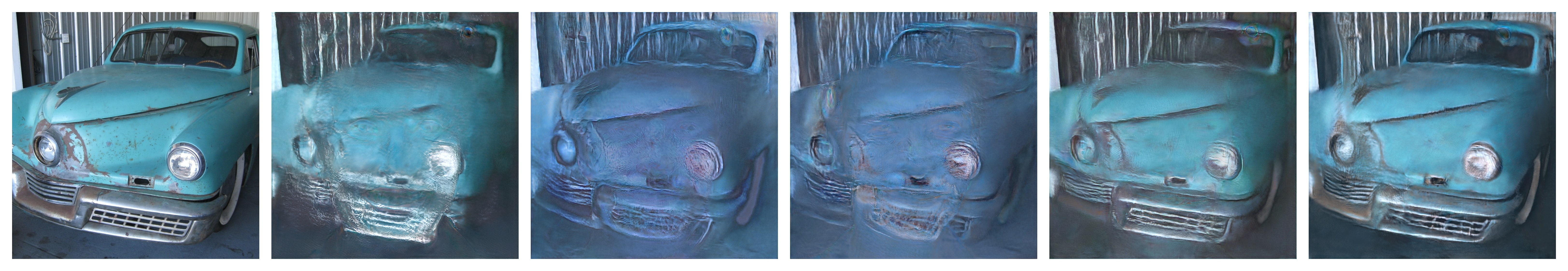}
\includegraphics[width=0.99\textwidth]{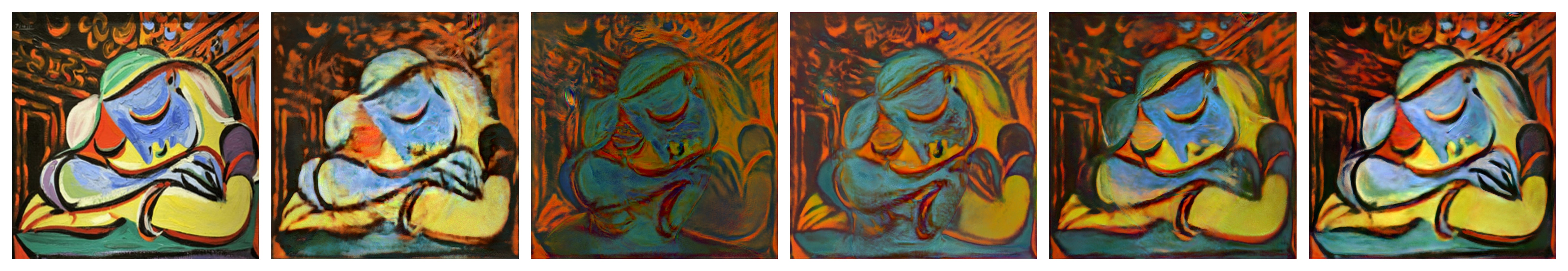}
 \caption{Additional results of the algorithmic choice justification on the loss function. Each row shows the results of an image from the five different classes in our test dataset respectively.
 From left to right, each column shows: (1) the original image; (2) pixel-wise MSE loss only; (3) perceptual loss on VGG-16 $conv3\_2$ layer only; (4) pixel-wise MSE loss and VGG-16 $conv3\_2$; (5) perceptual loss only; (6) our loss function . }
    \label{fig:algorithmic_choice_loss_function}
\end{figure*}
\begin{figure*}[t]
    \centering
    \includegraphics[width=\linewidth]{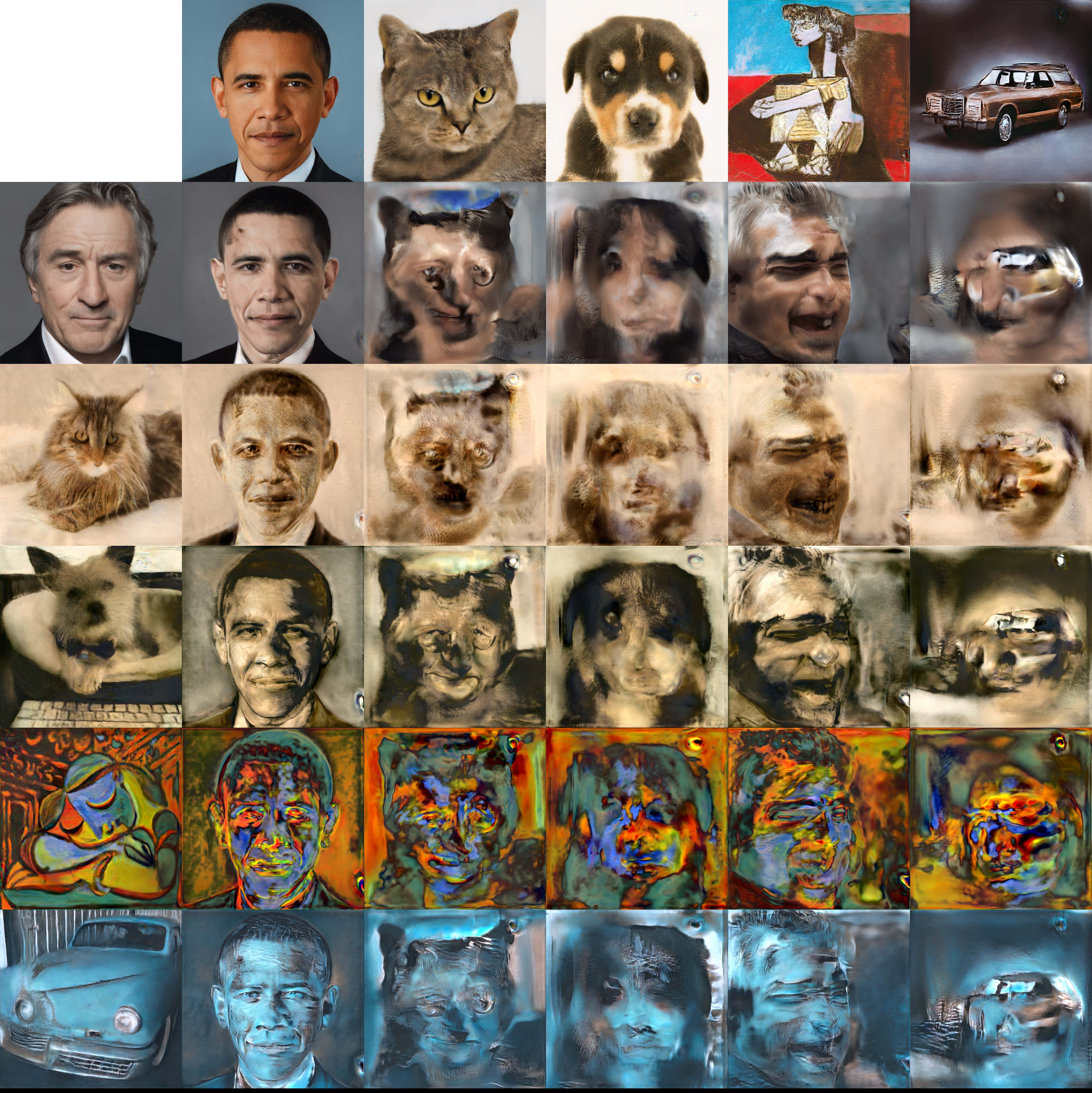}
    \caption{Complete table of the style transfer results. Left-most column: the embedded style image. First row: the embedded content images.}
    \label{fig:style}
\end{figure*}

\begin{figure*}[h]
\begin{subfigure}{0.99\linewidth}
    \centering
    \includegraphics[width=0.7\linewidth]{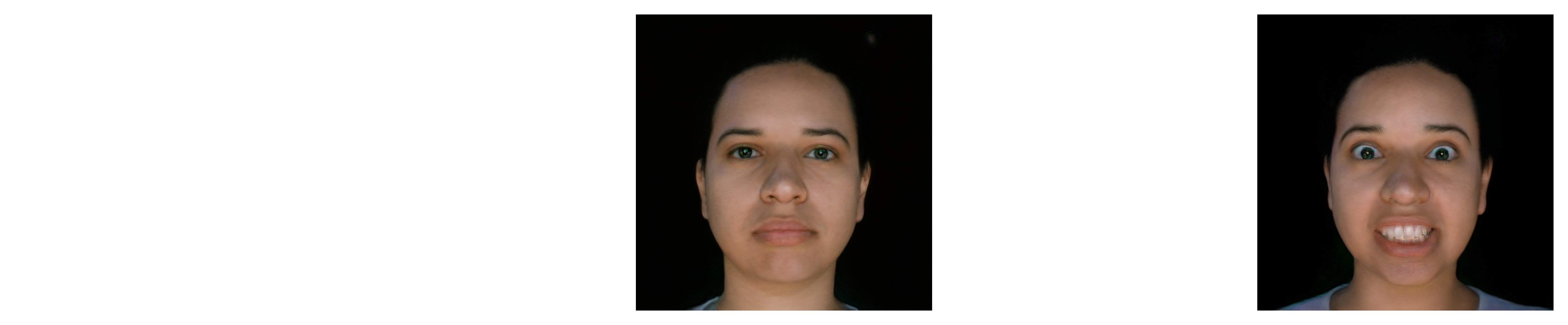}
    
    \includegraphics[width=0.7\linewidth]{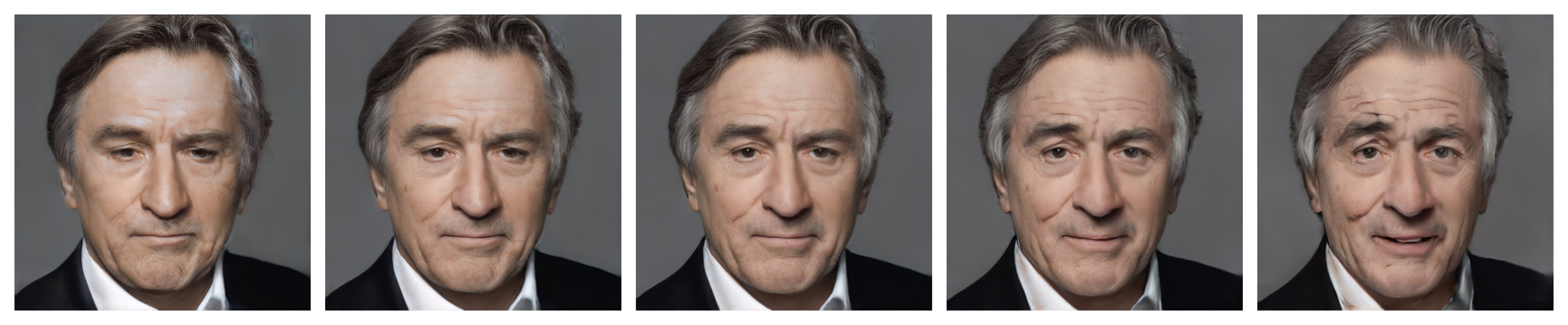}
    
    \includegraphics[width=0.7\linewidth]{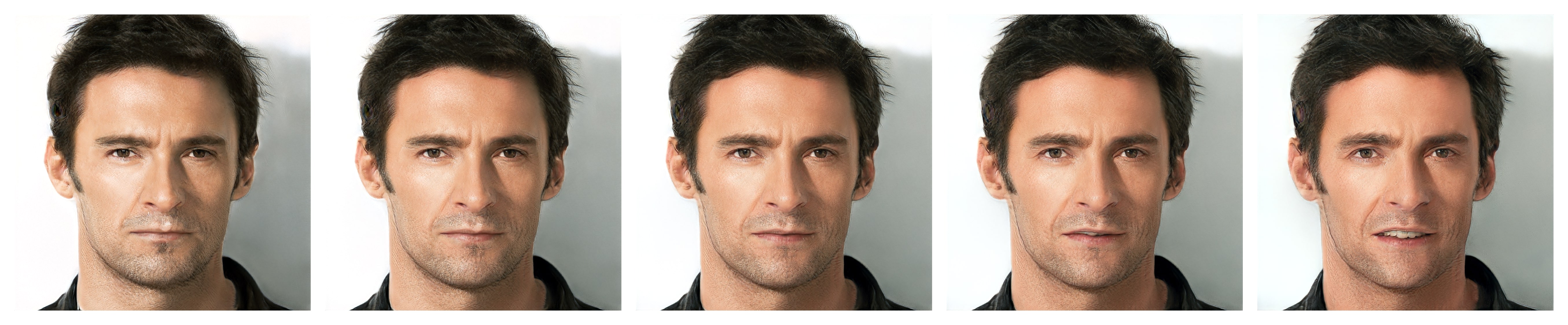}
    
    \includegraphics[width=0.7\linewidth]{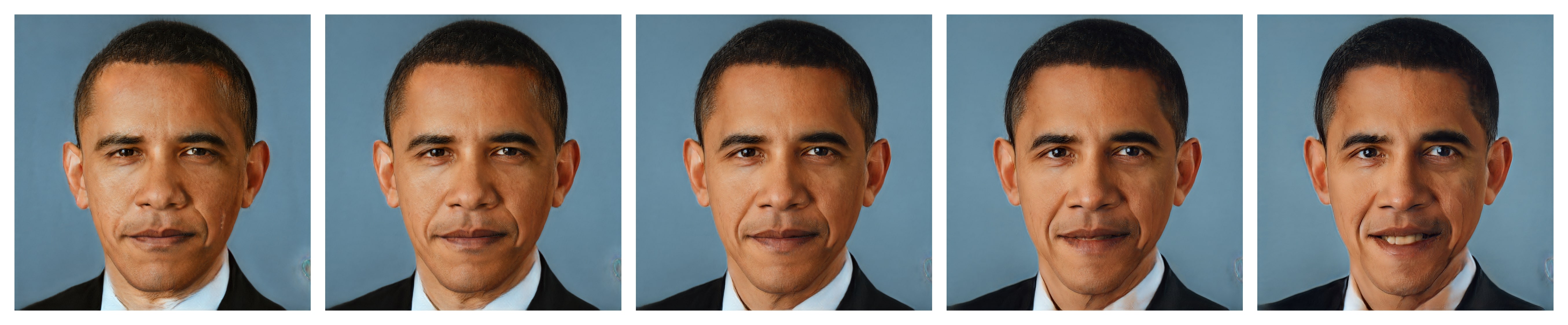}
    
    \includegraphics[width=0.7\linewidth]{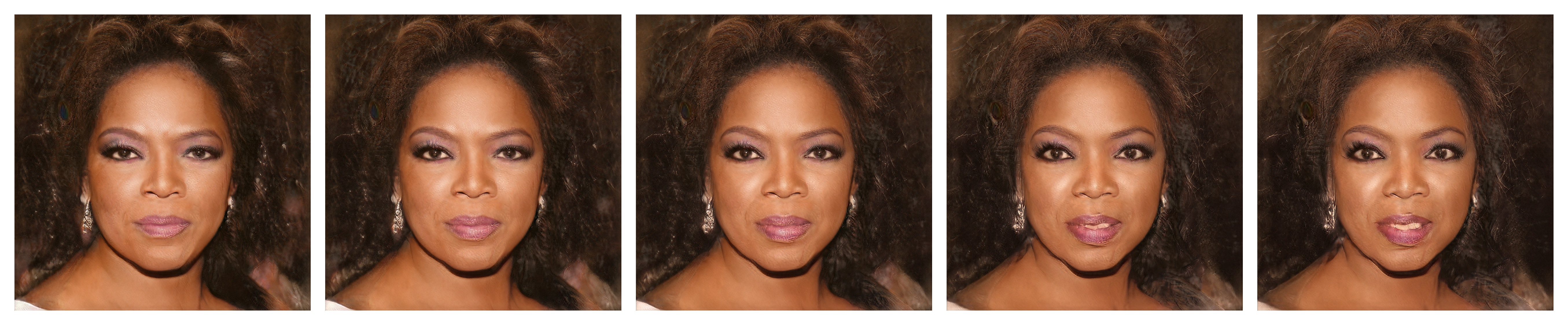}
    
    \includegraphics[width=0.7\linewidth]{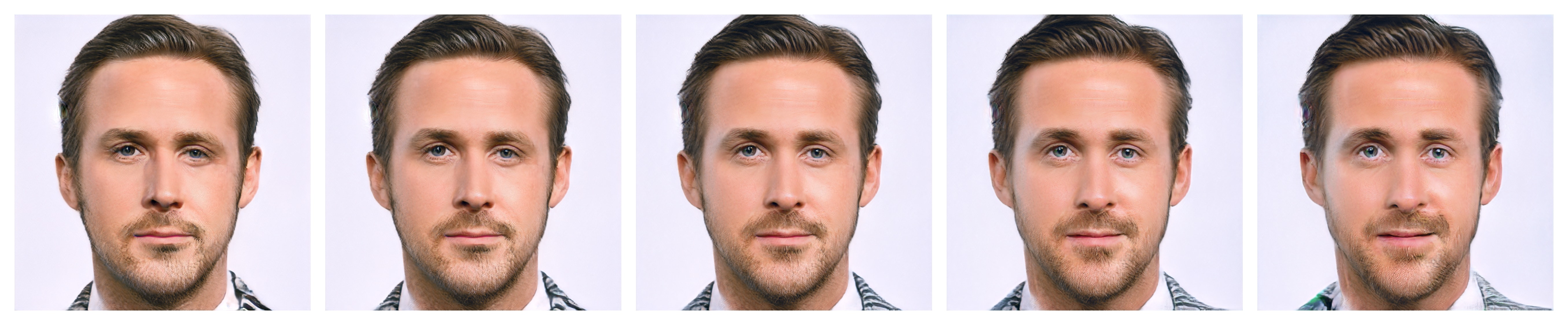}
    
    \includegraphics[width=0.7\linewidth]{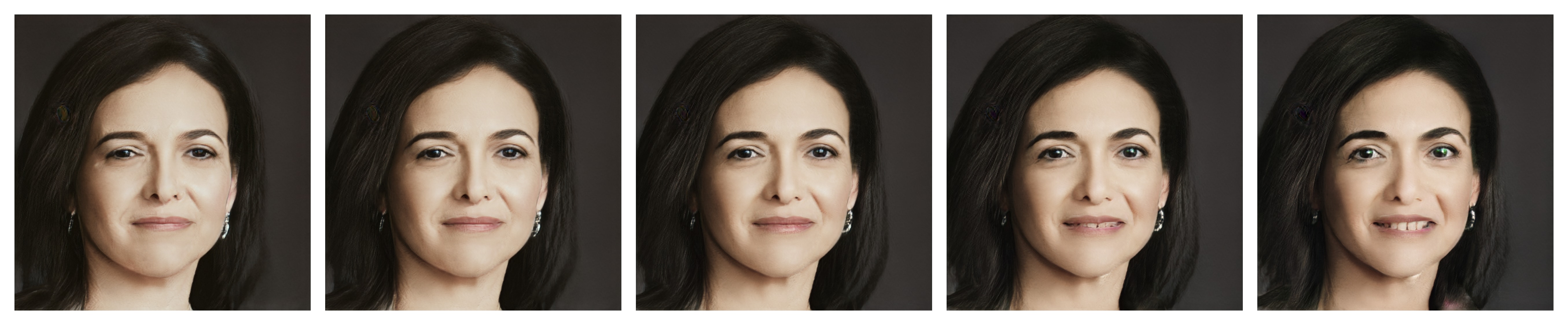}
    
    \includegraphics[width=0.7\linewidth]{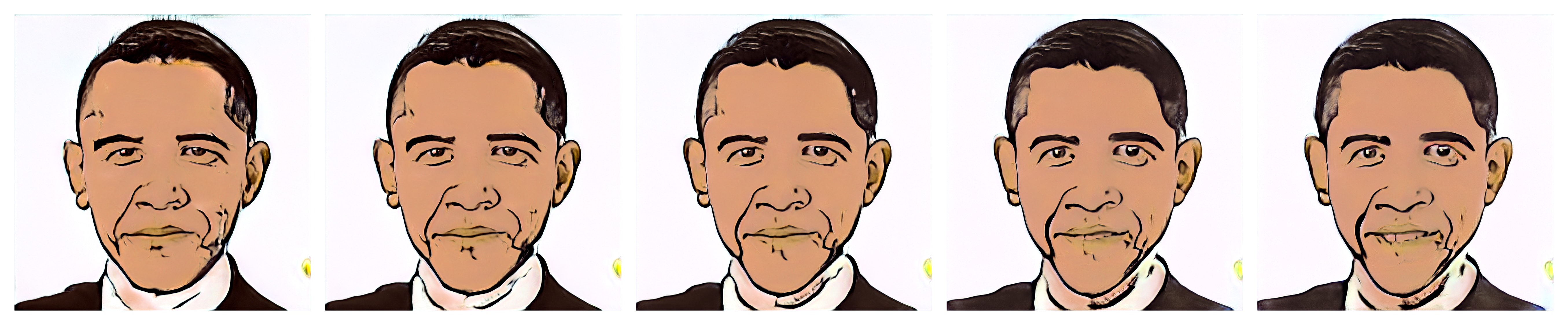}
    \subcaption*{(a)}
\end{subfigure}
\end{figure*}

\begin{figure*}[h]
\begin{subfigure}{0.99\linewidth}
    \centering

    \includegraphics[width=0.7\linewidth]{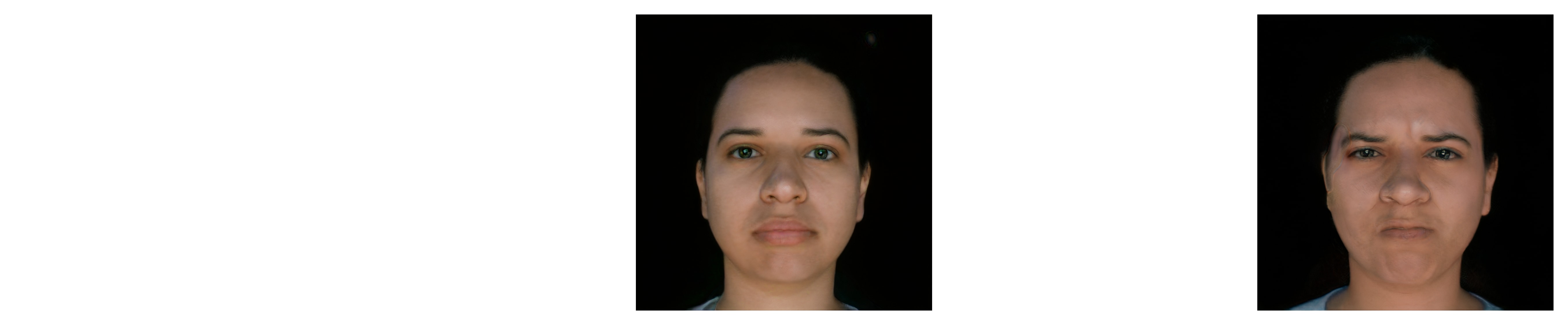}
    
    \includegraphics[width=0.7\linewidth]{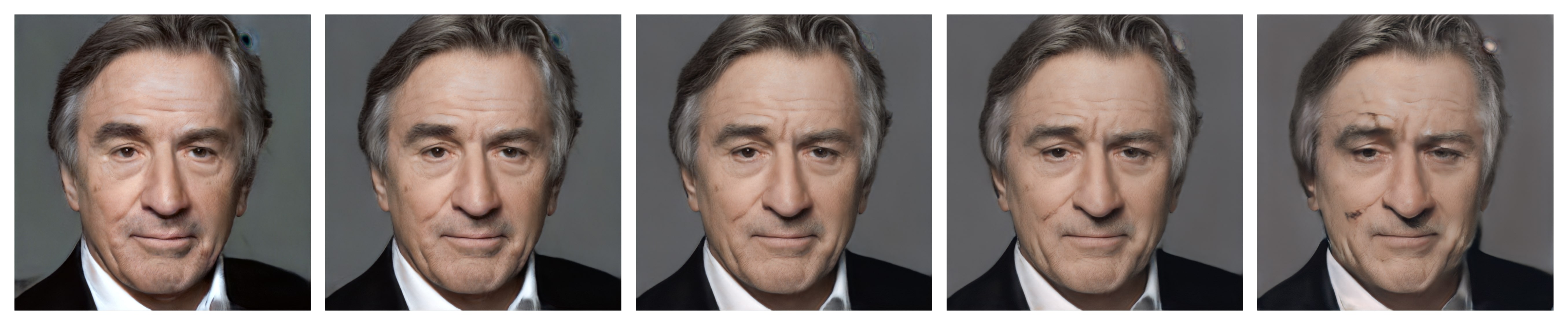}
    
    \includegraphics[width=0.7\linewidth]{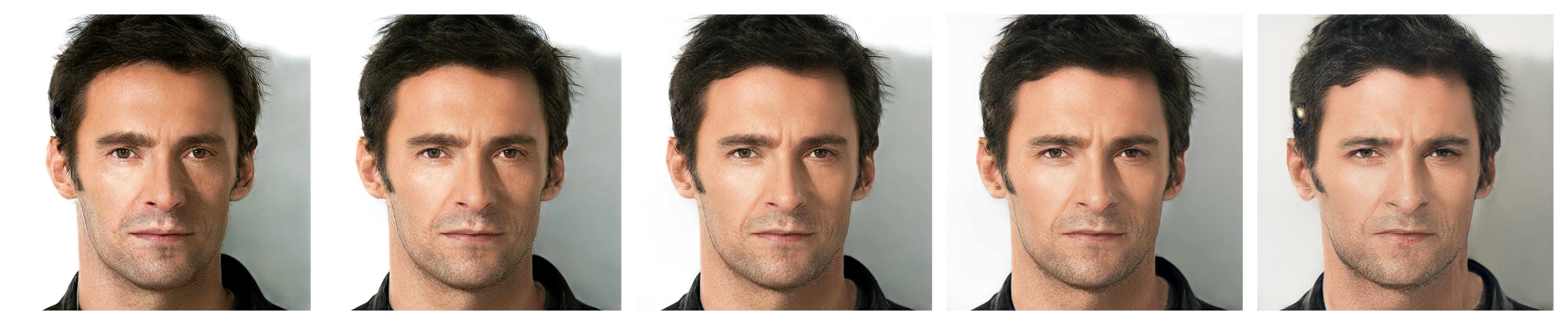}
    
    \includegraphics[width=0.7\linewidth]{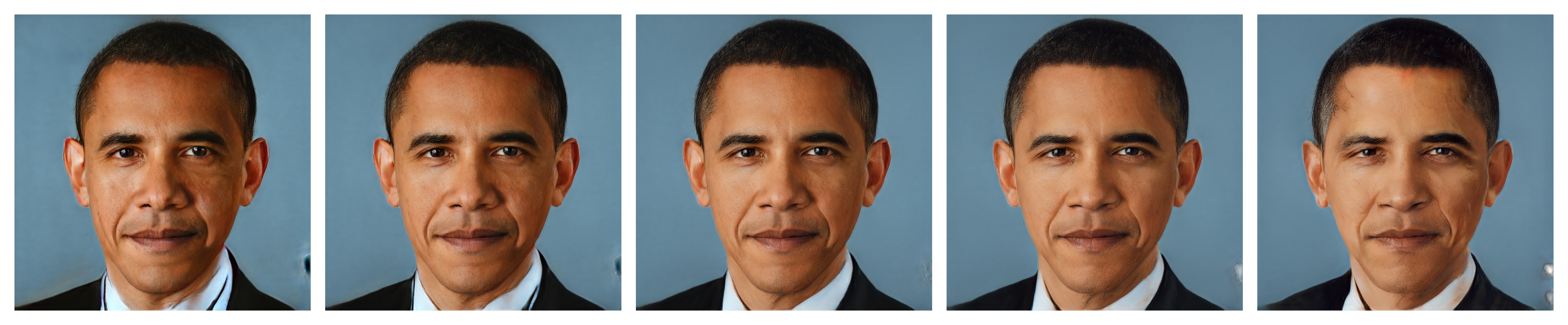}
    
    \includegraphics[width=0.7\linewidth]{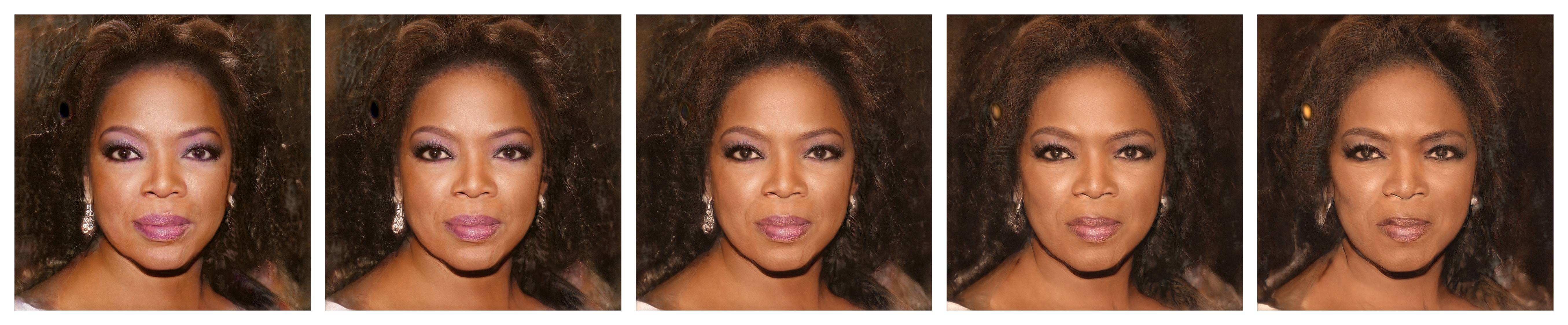}
    
    \includegraphics[width=0.7\linewidth]{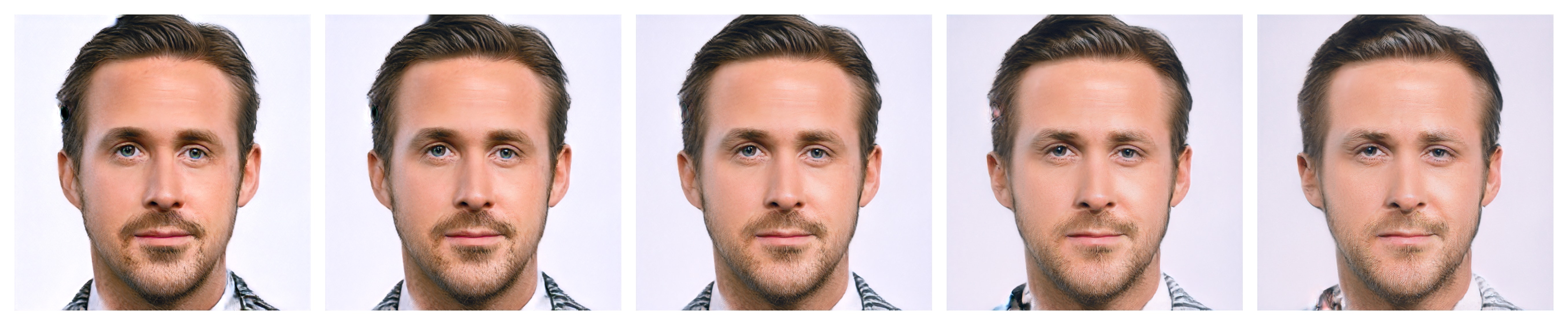}
    
    \includegraphics[width=0.7\linewidth]{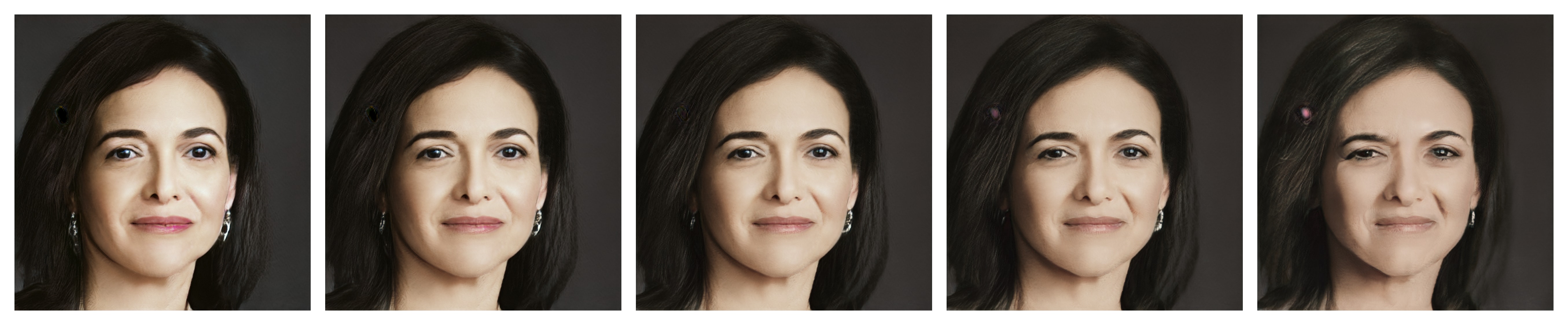}
    
    \includegraphics[width=0.7\linewidth]{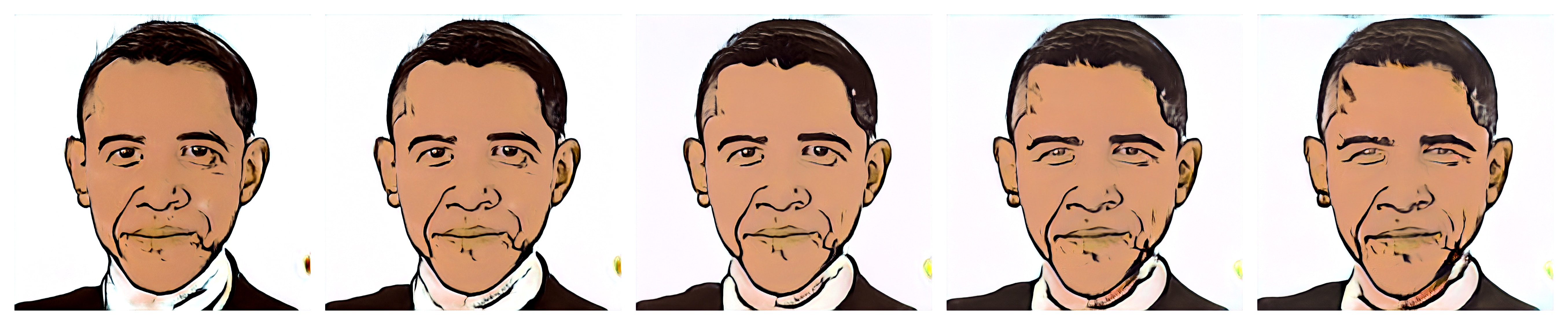}
    \subcaption*{(b)}
\end{subfigure}
\caption{Additional results on expression transfer.
In each subfigure, the first row shows the reference images from IMPA-FACES3D \cite{IMPA_FACES3D2011} dataset; in the following rows, the middle image in each of the examples is the embedded image, whose expression is gradually transferred to the reference expression (on the right) and the opposite direction (on the left) respectively.}
\label{fig:expression_transfer}
\end{figure*}

\paragraph{Clustering or Scattering?}
To support our insight that only face images form a cluster in the latent space, we compute the $L2$ distances between the embeddings of all pairs of test images (Figure \ref{fig:heat2}).
It can be observed that the distances between the faces are relatively smaller than those of other classes, which justifies that they are close to each other in the $W^+$ space and form a cluster. 
For images in other classes, especially the paintings, the pairwise distances are much higher. This implies that they are scattered in the latent space.

\paragraph{Justification of Loss Function Choice}
Figure \ref{fig:algorithmic_choice_loss_function} validates the algorithmic choice of the loss function used in the main paper.  
It can be observed that (i) matching the image features at multiple layers of the VGG-16 network works better than at a single layer; (ii) the combination of pixel-wise MSE loss and perceptual loss works the best.

\paragraph{Influence of Noise Channels}
Figure \ref{fig:emb} shows that restarting the embedding with a different noise leads to similar results.
In addition, we observed significantly worse quality when resampling the noise during the embedding (at each update step).
To this end, we kept the noise channel constant during the embedding for all our experiments.

\section{Additional Results on Applications}

Figure \ref{fig:mor} shows additional results of the image morphing. 
Figure \ref{fig:style} shows the complete table of the style transfer results between different classes. 
The results support our insight that the multi-class embedding works by using an underlying human face structure (encoded in the first couple of layers) and painting powerful styles onto it (encoded in the latter layers). Figure \ref{fig:expression_transfer} shows additional results on the expression transfer. We also include an accompanying video in the supplementary material to show it works with noisy images taken by a commodity camera in a typical office environment. 
The random walk results (of two classes `human faces' and `cars') from the embedded image towards the mean face image are also shown in videos.

\end{document}